\documentclass[11pt]{article}

\usepackage[final]{acl}

\usepackage{times}
\usepackage{latexsym}

\usepackage[T1]{fontenc}

\usepackage[utf8]{inputenc}

\usepackage[final,nopatch=footnote]{microtype}

\usepackage{inconsolata}

\usepackage{epsfig}
\usepackage{graphicx}
\usepackage{times}

\usepackage{comment}

\usepackage{pifont}

\usepackage{microtype}
\usepackage[font=small]{caption}
\usepackage{arydshln}
\usepackage{tabularx}
\usepackage{adjustbox}
\usepackage{array}
\usepackage{longtable}
\usepackage{xtab}
\usepackage{colortbl}
\usepackage{float,wrapfig}
\usepackage{hhline}
\usepackage{multirow}
\usepackage{subfigure} %
\usepackage[percent]{overpic}

\usepackage{fontawesome}

\usepackage{stmaryrd}
\usepackage{breqn}

\usepackage{amsmath,amsfonts,amssymb}
\usepackage{amsthm}
\usepackage{amsmath}
\usepackage{mathtools}
\usepackage{duckuments}

\newtheorem{theorem}{Theorem}
\newtheorem{proposition}{Proposition}
\newtheorem{lemma}{Lemma}

\usepackage{bm}
\usepackage{nicefrac}
\usepackage{microtype}
\usepackage{dsfont}
\usepackage{changepage}
\usepackage{extramarks}
\usepackage{fancyhdr}
\usepackage{setspace}
\usepackage{soul}
\usepackage{xspace}
\usepackage{hhline}
\usepackage{pifont}
\usepackage{multirow}

\usepackage{makecell}

\usepackage{enumitem}
\usepackage[title]{appendix}

\usepackage{placeins}
\usepackage[stable]{footmisc}

\usepackage[most]{tcolorbox}

\makeatletter
\newcommand{\smallsym}[2]{#1{\mathpalette\make@small@sym{#2}}}
\newcommand{\make@small@sym}[2]{%
  \vcenter{\hbox{$\m@th\downgrade@style#1#2$}}%
}
\newcommand{\downgrade@style}[1]{%
  \ifx#1\displaystyle\scriptstyle\else
    \ifx#1\textstyle\scriptstyle\else
      \scriptscriptstyle
  \fi\fi
}
\makeatother

\newcommand{\ignorethis}[1]{}

\newcommand{\myparagraph}[1]{\vspace{1pt} \noindent \textbf{#1}}

\def\1{\bm{1}}

\newcommand{\R}{\mathbb{R}}

\newcolumntype{L}[1]{>{\raggedright\let\newline\\\arraybackslash\hspace{0pt}}m{#1}}
\newcolumntype{C}[1]{>{\centering\let\newline\\\arraybackslash\hspace{0pt}}m{#1}}
\newcolumntype{R}[1]{>{\raggedleft\let\newline\\\arraybackslash\hspace{0pt}}m{#1}}

\newcommand{\ignore}[1]{}

\renewcommand*{\thefootnote}{\arabic{footnote}}

\makeatletter
\DeclareRobustCommand\onedot{\futurelet\@let@token\@onedot}
\def\@onedot{\ifx\@let@token.\else.\null\fi\xspace}
\def\eg{e.g\onedot,\xspace} 

\def\ie{i.e\onedot,\xspace}

\def\etc{\emph{etc}\onedot}

\makeatother

\usepackage{color}
\usepackage{xcolor}

\definecolor{R1}{rgb}{1,0.5,0}
\definecolor{R2}{rgb}{0.19, 0.55, 0.91}
\definecolor{R3}{rgb}{0.58, 0.34, 0.92}

\definecolor{pref}{rgb}{0.13, 0.55, 0.13}

\usepackage{float}
\usepackage{soul}
\usepackage{tikz}

\definecolor{celadon}{rgb}{0.67, 0.88, 0.69}
\definecolor{oldrose}{rgb}{0.75, 0.5, 0.51}

\definecolor{ggr}{gray}{0.92}
\definecolor{fgreen}{rgb}{0.24,0.50,0.19}
\definecolor{babyblueeyes}{rgb}{0.63, 0.79, 0.95}
\definecolor{ssrhl}{rgb}{0.88, 0.93, 1.0}

\usepackage[utf8]{inputenc} %
\usepackage[T1]{fontenc}    %
\usepackage{hyperref}       %
\usepackage{url}            %
\usepackage{booktabs}       %
\usepackage{stfloats}

\newcolumntype{a}{>{\columncolor{ggr}}c}

\title{An Embarrassingly Simple Data Selection Strategy for Million-Scale Multi-Modal Instruction-Tuning}

\author{
 \textbf{Mengyao Lyu\textsuperscript{1,2}},
 \textbf{Yan Li\textsuperscript{3}},
 \textbf{Huasong Zhong\textsuperscript{3}},
 \textbf{Hui Chen\textsuperscript{1,2}},
\\
 \textbf{Jungong Han\textsuperscript{1,2}},
 \textbf{Guiguang Ding\textsuperscript{1,2,\dag}},
 \textbf{Zhenheng Yang\textsuperscript{3}}
\\
 \textsuperscript{1}Tsinghua University,
 \textsuperscript{2}BNRist,
 \textsuperscript{3}Bytedance
\\
  \texttt{mengyao.lyu@outlook.com} \quad \texttt{dinggg@tsinghua.edu.cn} \\
}

\begin{document}
\maketitle
\begingroup
\renewcommand\thefootnote{\textdagger}
\footnotetext{Corresponding author.}
\endgroup
\begingroup
\renewcommand\thefootnote{}
\footnotetext{Project page: \urlstyle{same}\url{https://lyumengyao.github.io/projects/ssr}.}
\endgroup

\begin{abstract}
Recent research demonstrates that pretrained large language models (LLMs) require minimal but carefully curated data for instruction-tuning. Building on LLMs, multi-modal models further contend with far larger data volumes, heterogeneous sources, and cross-modal alignment, amplifying both the necessity and the complexity of data selection.
However, existing methods rely on \textbf{deeply extracted} embeddings, gradients or \textbf{narrow} quality scores that are time-consuming to compute and limited in expressiveness, frequently failing to surpass a simple random-sampling baseline.
To overcome these limitations, we step back from complex embedding spaces
and target the fundamental objective of instruction tuning: {how} and {what} knowledge should be elicited from the pretrained model and exposed to users. 
We leverage interaction \underline{S}tyles as a proxy for structural diversity, and decompose data quality into capability-specific \underline{S}cores, yielding an \textbf{explicit} yet \textbf{\underline{R}ich} (SSR) profile for each sample.
We further cast the selection as submodular coverage maximization over the capability--style grid, providing an approximation guarantee.
SSR consistently delivers superior performance across two distinct architecture stacks, LLaVA-1.5 (Vicuna + CLIP) and LLaVA-OneVision (Qwen2 + SigLIP), two large-scale data pools, six data budget settings, and 14 multi-modal benchmarks.
To our knowledge, SSR is the first data selection method validated at million-scale, achieving 99.1\% of full-data performance using only 30\% of a 2.6M pool.
Furthermore, the SSR profile models and selected subsets transfer training-free across data pools and model architectures
with capability extensibility.
\end{abstract}

\section{Introduction}
\vspace{-0.05in}
\label{sec:intro}

\begin{figure}[t]
    \centering
    \vspace{-0.1in}
    \includegraphics[width=1\columnwidth]{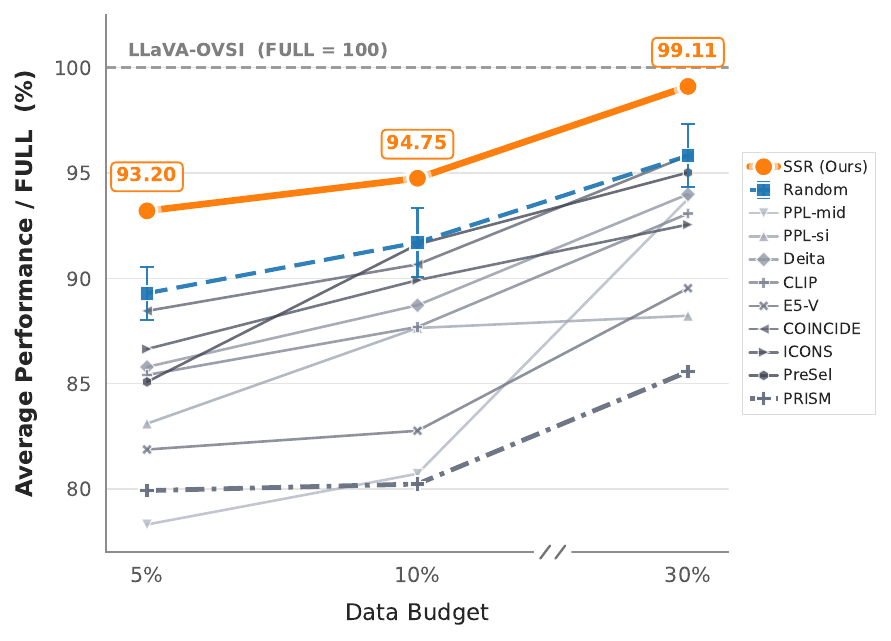}
    \vspace{-0.35in}
    \caption{\textbf{Million-scale selection exposes the failure of prior methods.} Failing to beat random sampling across 5\%, 10\%, and 30\% budgets on the 2.6M data pool, existing methods expose a scalability gap that SSR closes with consistent gains.}
    \label{fig:avg_lineplot}
    \vspace{-0.25in}
\end{figure}

Data matters for the successful scaling of large foundation models~\cite{li2024DataCompLMSearchNext,wettig2024QuRatingSelectingHighQuality,liu2024Deita,lu2024INSTAG,li2024LLaVAOneVisionEasyVisual}, particularly during supervised fine-tuning (SFT), which aligns pre-trained models to follow diverse instructions across general and specialized domains.
Early work on large language models (LLMs) focused on filtering million-scale SFT pools for deduplication~\cite{lee2022dedupllm,elazars2024whatsinmybigdata}, quality, and safety~\cite{joulin2016fasttext,penedo2023refinedweb,dubey2024Llama3Herd,team2024gemma,chung2024flant5}.
LIMA~\cite{zhou2024lima} then argued that ``less is more'' for alignment: with 1{,}000 carefully curated samples, it showed that LLM knowledge is acquired during pre-training, so instruction tuning needs little data provided it is of high quality and diverse. This sparked a wave of automated selection methods~\cite{lu2024INSTAG,xia2024LESSSelectingInfluential,liu2025SelectITSelectiveInstruction}.
However, subsequent surveys~\cite{diddee2024ChasingRandomInstruction,xia2024RandAllYouNeed} reveal that these methods are highly sensitive to data budgets, sources, and evaluation protocols, and frequently fail to beat random sampling. They typically depend on \textit{heavy extracted} representations such as high-dimensional embeddings or gradients~\cite{liu2024Deita,pang2024ds2,li2024QuantityQualityBoosting,xia2024LESSSelectingInfluential,presel2025}, which are costly to scale and may hurt transferability~\cite{liu2025SelectITSelectiveInstruction}. A complementary line rates data with proprietary LLMs (\eg GPT-4) on heuristic metrics~\cite{wettig2024QuRatingSelectingHighQuality,liu2024Deita,chen2024AlpaGasusTrainingBetter}, but the resulting criteria are typically too \textit{narrow}, limiting expressiveness~\cite{liu2024Deita,pang2024ds2} and generality~\cite{wettig2024QuRatingSelectingHighQuality}.

These failure modes are further exacerbated in Multi-modal LLMs (MLLMs): massive token volumes, heterogeneous sources, and
diverse downstream requirements turn \textit{million-scale} fine-tuning into a critical yet underexplored bottleneck for data selection (Fig.~\ref{fig:avg_lineplot}).
\textit{First}, \textit{model-centric} statistics fail to scale at million-level MLLM data volumes. High-dimensional intrinsic-feature clustering~\cite{coincide,presel2025,bi2025prism}, gradient-based distribution matching~\cite{coincide}, and loss re-weighting via score embeddings~\cite{chen2024self_filter} grow prohibitive in compute or brittle under distribution shift, leaving state-of-the-art (SOTA) automated selection research stuck at thousand-scale datasets and lagging orders of magnitude behind today's million-sample MLLM pools.
\textit{Second}, selection signals anchored in model-specific latents do not readily transfer across settings by construction. Each shift in target architecture 
poses new extraction costs and adaptation challenges, and such signals remain brittle under changes in budget, source, or evaluation protocol~\cite{diddee2024ChasingRandomInstruction,xia2024RandAllYouNeed}.
\textit{Third}, existing methods yield selections too narrow for the broad generality that MLLM demands. Beyond model-specific latents, scalar criteria collapse each sample into one or a few dimensions, leaving the diverse axes of downstream MLLM capabilities uncovered, while noise and variability in multi-modal curation~\cite{chen2024sharegpt4v,li2024LLaVAOneVisionEasyVisual,liu2023llava15} further make selection for target-model generality fragile.
Experiments in Sec.~\ref{sec:exp-main} validate these vulnerabilities and motivate a new paradigm for multi-modal selection.

Toward such a paradigm, we ground our method in the fundamental objective of SFT. Beyond ``less is more'', LIMA's Superficial Alignment Hypothesis~\cite{zhou2024lima} posits that a model's core knowledge is acquired during pre-training and fine-tuning mainly teaches how to articulate that knowledge. This motivates two axes for sample valuation: an interaction \textit{style} encoding \textit{how} the knowledge is delivered, and a capability \textit{score} encoding \textit{what} knowledge should be prioritized.
Interaction \textbf{S}tyles and capability \textbf{S}cores together form an explicit yet \textbf{R}ich profiling framework (\textbf{SSR}): each sample receives a style label (caption, yes/no) and a vector of transparent, human-interpretable capability scores, both judged per sample superficially without latent computation. We cast selection over this Cartesian-product coverage grid as a monotone submodular maximization under cardinality and partition-matroid constraints.
This design departs from previous methods and directly addresses the challenges raised above. 
\textit{First}, against model-centric statistics, SSR adopts a \textit{data-centric} profile: each sample is judged explicitly and superficially without latent computation, which scales linearly to millions of samples.
\textit{Second}, against model-centric latents, SSR's profile is intrinsic to each sample: style and capability are read superficially and explicitly from the sample itself, so different models and data pools converge on the same profile. The profile therefore transfers training-free across data domains and architectures, where latent-anchored methods drift.
\textit{Third}, against fragile generality, our capability--style coverage is rich, with $(1-1/e)$ coverage and EF1 fairness guarantees across cells (Thms.~\ref{thm:approx},~\ref{thm:ef1}).
SSR is therefore scalable, transferable, and generalizable by construction.

We validate SSR on three architecture families, \ie LLaVA-1.5 (Vicuna + CLIP)~\cite{liu2023llava15}, LLaVA-OneVision (Qwen2 + SigLIP)~\cite{li2024LLaVAOneVisionEasyVisual}, and the vision-centric Cambrian-1 (Llama-3 + a multi-encoder ensemble)~\cite{tong2024cambrian}, and three data pools at different scales: the 624K ShareGPT4V pool, the 2.6M LLaVA-OneVision pool, and the 7M Cambrian-7M pool.
Despite its simplicity, SSR consistently outperforms baselines across six data budgets and two model sizes, benchmarked over 14 diverse evaluations, and transfers training-free to out-of-domain data pools and unseen architectures.
Our main contributions are:
\begin{itemize}[itemsep=1pt, parsep=0pt, topsep=0pt, partopsep=0pt]
    \item We propose a SFT-aligned selection paradigm, SSR, that delivers key multi-modal capabilities through diverse interactional styles, yielding an explicit, model-agnostic profiling of multi-modal instruction data.
    \item We are the first to scale principled multi-modal SFT data selection to the million-scale regime, an urgent step as today's MLLM training pools have outgrown the thousand-scale convention of prior selection research.
    \item We achieve SOTA effectiveness, transferability and scalability across 3 architecture families, 6 data budgets, and 14 benchmarks.
\end{itemize}

\section{Related Work}
\vspace{-0.05in}
We draw on data efficiency studies in LLMs, MLLMs, VLMs, and active learning, broadly categorized into hand-crafted heuristics, model-based indicators, and judge-based scores, which can be interchangeably or complementarily applied.

\myparagraph{Hand-Crafted Heuristics} use expert-defined quality criteria, \eg relevance, clarity, diversity, safety, vocabulary, N-grams, length, for filtering~\cite{team2023gemini,touvron2023llama,dubey2024Llama3Herd,qin2024UnleashingPowerData,penedo2023refinedweb}. They are interpretable but labor-intensive, bias-prone, narrow, and potentially ill-suited to multi-modality.

\myparagraph{Model-Based Indicators} leverage target-model gradients, predictive distributions, or embeddings to assess uncertainty, learnability, and similarity~\cite{evans2024BadStudentsMake,liu2024LessMoreHighvalue,lyu2023box,lyu2025lftl,sener2018coreset,liu2024Deita,liu2025SelectITSelectiveInstruction,settles09AL}, with strong in-domain results but limited interpretability and transferability~\cite{diddee2024ChasingRandomInstruction,munjal2022RobustReproducibleAL}. Proxy-model variants train classifiers~\cite{joulin2016fasttext,brown2020gpt3,gao2025MetadataConditioningAccelerates,li2024DataCompLMSearchNext} or use proxy representations: COINCIDE~\cite{coincide} trains a 2B proxy on the target pool for coreset~\cite{sener2018coreset} selection via high-dimensional clustering, ICONS~\cite{wu2024icons} estimates sample influence via gradient similarity to a validation set, PreSel~\cite{presel2025} trains a reference model to compute modality relevance and task importance, and PRISM~\cite{bi2025prism} scores samples by their correlation with a corpus-mean-centered pool of MLLM features: all anchor sample value in one model's latents, and incur substantial overhead at scale. VLM-side multi-modal quality metrics target alignment~\cite{maini2024TMARSImprovingVisual,gadre2023DataCompSearchNext,goyal2024ScalingLawsData} but correlate weakly with instruction-following and lead to redundancy.

\myparagraph{Judge-Based Scoring}~\cite{liu2024TakeEssenceDiscard,liu2024Deita,chen2024AlpaGasusTrainingBetter,pang2024ds2} employs an external judge, often a proprietary API, as a cost-effective alternative to human annotation for candidate valuation. The resulting LLM-side quality dimensions are often too narrow to capture multi-modal variability, and additional sample-level or pairwise scoring with similarity thresholds~\cite{liu2024Deita} does not scale globally, leaving judge-based selection largely unexplored for MLLMs. SSR addresses both: a multi-dimensional capability axis broadens beyond singular scalars, and an interaction-style axis with submodular coverage replaces latent compute, making judge-based selection viable and effective at MLLM scale.

\myparagraph{Offline vs.\ Online Selection.} The methods above perform \emph{offline} selection: a subset is chosen once from a static pool before fine-tuning, so the output is a reusable, auditable artifact independent of any particular training run. A complementary line performs \emph{online} selection, where per-batch decisions track the state of the model being trained: OASIS~\cite{lee2025oasisonlinesampleselection} adaptively admits informative samples during continual visual instruction tuning, and PROGRESS~\cite{progress2025} clusters the pool offline into concepts, then prioritizes them online by the model's relative error-driven learning progress. Online selection reacts to what the model has learned, but couples selection to training, yields no transferable subset, and presumes a fixed target architecture. Offline selection instead amortizes profiling once and, in SSR's data-centric form, transfers across pools and architectures. 
The two paradigms are complementary rather than competing, as PROGRESS's illustrates; as an offline selector, SSR is compared against offline methods.

\section{Method}
\label{sec:method}
\vspace{-0.05in}

\begin{figure*}[htbp]
\begin{center}
\vspace{-0.3in}
\centerline{\includegraphics[width=1\linewidth]{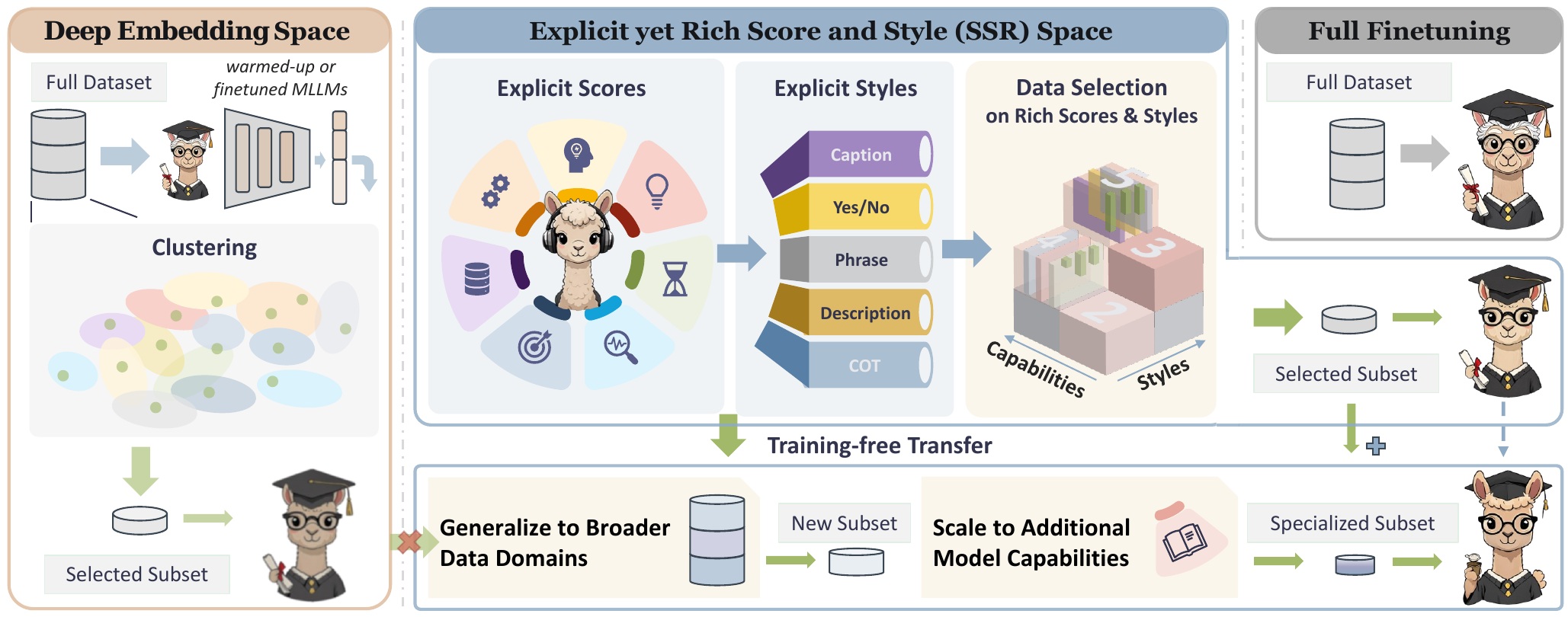}}
\caption{\textbf{Overview of the SSR framework compared to existing paradigms.}
Full fine-tuning indiscriminately processes all data (Top Right), while previous selection methods (Left) rely on \emph{model-centric} signals (\eg latent embeddings). In contrast, our proposed SSR (Middle) adopts a \emph{data-centric} profile: rich and explicit capability scores and an interaction style, which together drive submodular coverage maximization over the capability--style grid, yielding an efficient subset with superior performance, training-free transfer across data pools and target models, and seamless capability extensibility.
}
\label{fig:main}
\end{center}
\vskip -0.4in
\end{figure*}

\myparagraph{Problem setup.} Given a large-scale multi-modal instruction pool $D=\{X_1,\dots,X_N\}$ with $X_i=(I_i,Q_i,A_i)$ and a budget percentage $b\%\,$ ($b\in(0,100]$, e.g., $b=10$ or $30$ in our experiments), we seek a subset $\hat D\subseteq D$ with $|\hat D|\le b\%\cdot|D|$ that, when used to fine-tune a pretrained MLLM $\mathcal{M}_\theta$, maximizes downstream instruction-following ability. Inspired by the Superficial Alignment Hypothesis (SAH)~\cite{zhou2024lima}, we ground selection in two complementary axes of sample value: a capability score that encodes \textit{what} knowledge each sample conveys, and an interaction style that encodes \textit{how} that knowledge is delivered. Let $C=\{c_1,\dots,c_{|C|}\}$ be the set of capabilities and $\Sigma=\{\sigma_1,\dots,\sigma_{|\Sigma|}\}$ the set of interaction styles (taxonomy derivation in Supp.~\ref{sec:sup-expset}). The Cartesian product $C\times\Sigma$ defines $|C|\cdot|\Sigma|$ capability--style \emph{cells}; covering this grid is the formal target of SSR.

\myparagraph{Pipeline.} As illustrated in Fig.~\ref{fig:main}, SSR comprises two algorithmic stages, accompanied by inherent benefits. (i) \emph{Profiling}: assign every $X_i\in D$ a non-negative capability vector $r_i\in[0,5]^{|C|}$ and an interaction-style indicator $g_i\in\{0,1\}^{|\Sigma|}$ (Sec.~\ref{sec:method-profile}).
(ii) \emph{Selection}: maximize a monotone submodular coverage objective $F$ over the capability--style grid by stratified round-robin, which admits both a $(1-1/e)$ approximation bound and an envy-free-up-to-one-item (EF1) fairness guarantee~\cite{lipton2004approximately} (Sec.~\ref{sec:method-select},~\ref{sec:method-guarantees}). (iii) \emph{Extensibility}: by carrying semantic rather than model-specific signals, the profile transfers across data pools and target architectures (Sec.~\ref{sec:exp-transfer}), and the capability set $C$ accommodates new skills (Supp.~\ref{sec:sup-scale}).

\subsection{SSR Profiling}
\label{sec:method-profile}

The dominant paradigm for MLLM data selection is \emph{model-centric}: per-instance value is computed from the target MLLM itself via gradient norms, influence functions, latent embeddings or predicted loss~\cite{xia2024LESSSelectingInfluential,wu2024icons,coincide,presel2025}. These signals are coupled to one model checkpoint, demand gradient or hidden-state access, and yield representations that offer limited interpretability and tend to transfer less readily across target models or data pools. In contrast, a \emph{data-centric} metric assigns each sample a scalar quality score $s(X_i)\in\R$ by prompting a strong judge~\cite{pang2024ds2,wettig2024QuRatingSelectingHighQuality,chen2024AlpaGasusTrainingBetter,liu2024TakeEssenceDiscard}. Such scores are architecture-agnostic and semantically interpretable, but narrow in granularity.

Formally, for any summary $s$ of a sample $X$, the data-processing inequality bounds the mutual information with the downstream SFT utility $U$ as $I(s(X);U)\le I(X;U)$; how much of $I(X;U)$ is recovered by $s$ depends on its structure. One or a few broad criteria such as quality and complexity compress a sample's value into too few axes to track the heterogeneity of MLLM SFT utilities.
This gap is empirically witnessed by score-based methods~\cite{liu2024Deita,pang2024ds2,liu2024TakeEssenceDiscard} that re-inject diversity through an explicit post-hoc feature-dissimilarity-aware algorithm, a mitigation that re-introduces poor scalability and model-specific dependence. We instead read the axes of value off the structure of SFT
to factorize a sample's contribution into knowledge content and interaction style, two semantically distinct axes forming a \emph{rich} per-sample profile:
\begin{itemize}[itemsep=1pt,parsep=0pt,topsep=2pt,leftmargin=*]
\item\emph{Capability scores} $r_i\in[0,5]^{|C|}$ rate \emph{what} knowledge each instance contributes to each fine-grained capability (e.g., spatial reasoning, STEM knowledge, OCR). Scores are produced against a unified 0--5 rubric so they are commensurable across capabilities and samples.
\item\emph{Interaction styles} $g_i\in\{0,1\}^{|\Sigma|}$ record \emph{how} that knowledge is delivered, capturing the dominant interaction format (e.g., multiple choice, captioning) emphasized by LIMA's SAH.
\end{itemize}
$C$ and $\Sigma$ are derived from open-source instruction data through a one-time procedure of discovery, expansion, and iterative refinement (Supp.~\ref{sec:sup-expset}).
The resulting $|C|$ capabilities and $|\Sigma|$ interaction styles, being explicitly semantic rather than model-specific, are interpretable and transferable.

The two axes combine into a non-negative tensor $q\in\R_{\ge 0}^{N\times|C|\times|\Sigma|}$ whose per-sample slice $q_i\in\R_{\ge 0}^{|C|\times|\Sigma|}$ is the rank-one outer product
\vspace{-0.1in}
\begin{equation}
  q_{c\sigma}(i)\;=\;r_i^c\cdot g_i^\sigma,\qquad (c,\sigma)\in C\times\Sigma.
\label{eq:q}
\end{equation}
\vspace{-0.25in}

\noindent Equivalently, $q_i=r_i\otimes g_i$ replicates sample $i$'s capability scores along its active style channels ($g_i^\sigma>0$) and vanishes on inactive cells.

\subsection{Submodular Coverage Objective}
\label{sec:method-select}

Given target capabilities $\hat C\subseteq C$ and styles $\hat\Sigma\subseteq\Sigma$, we want $\hat D$ to cover the grid $\hat C\times\hat\Sigma$ broadly, not to greedily exhaust any single cell. We instantiate this intuition with the saturated coverage family of Lin and Bilmes~\cite{lin2011classsubmodular}: for each cell $(c,\sigma)$, the contribution saturates once a fraction of pool informativeness has been collected, encoding diminishing returns. Define
\vspace{-0.1in}
\begin{equation}
F(S)\;=\;\sum_{(c,\sigma)\in\hat C\times\hat\Sigma} w_{c\sigma}\,
\min\!\Bigl\{\tau_{c\sigma},\;\sum_{i\in S} q_{c\sigma}(i)\Bigr\},
\label{eq:ssr-submodular}
\end{equation}
with cell weights $w_{c\sigma}\ge 0$ (uniform by default) and saturation thresholds $\tau_{c\sigma}>0$ (defaulting to a fraction of pool capacity, $\tau_{c\sigma}=\alpha\!\cdot\!\sum_{i\in D} q_{c\sigma}(i)$ with $\alpha\in(0,1]$). The $\min$ operator turns per-cell sums from modular into submodular: once a cell has accumulated $\tau_{c\sigma}$ worth of evidence, additional items in it contribute zero marginal value. Note that $w_{c\sigma}$ and $\alpha$ are analysis variables, not hyperparameters.

\begin{proposition}[Submodularity]
\label{prop:submodular}
The set function $F\colon 2^D\to\R$ defined by Eq.~\ref{eq:ssr-submodular} is normalized, monotone non-decreasing, and submodular.
\end{proposition}
\noindent\emph{Proof sketch.} Each cell term composes a non-decreasing concave map $x\mapsto w_{c\sigma}\min\{\tau_{c\sigma},x\}$ with the modular sum $\sum_{i\in S} q_{c\sigma}(i)\ge 0$, which is monotone submodular~\cite{lin2011classsubmodular}; $F$ is their non-negative linear combination. Please refer to Supp.~\ref{sec:app-proof-prop1} for full proof. 

\myparagraph{Structural contrast with scalar top-$k$.}
Top-$k$ selection on any scalar quality score $s(X_i)$ maximizes a \emph{modular} set function $\sum_{i\in S} s(X_i)$ under a cardinality constraint: the optimum is exact, but the objective is additive in items and therefore does not penalize redundancy among them. Eq.~\ref{eq:ssr-submodular} is submodular by construction (Prop.~\ref{prop:submodular}): the saturation operator encodes diminishing returns within each cell, so once a cell is well covered, additional items in it contribute less to $F$ than items in under-covered cells. This change in problem class, from modular to submodular under a partition matroid, is what enables the approximation and fairness guarantees we exploit in Sec.~\ref{sec:method-guarantees}; the same guarantees do not arise for scalar scoring~\cite{liu2024Deita,chen2024AlpaGasusTrainingBetter} or single-axis tagging~\cite{lu2024INSTAG}, which lack a coverage structure to optimize.

\subsection{Approximation and Guarantees}
\label{sec:method-guarantees}

Exact maximization of Eq.~\ref{eq:ssr-submodular} under a cardinality or per-cell budget is NP-hard in general, subsuming maximum coverage. We approximate it with a \emph{stratified round-robin selector} that admits both a $(1-1/e)$ approximation bound and an envy-free-up-to-one-item (EF1) allocation guarantee across capability--style cells.

\myparagraph{Algorithm.} For each $(c,\sigma)\in\hat C\times\hat\Sigma$, form the candidate group $G_{c\sigma}=\{X_i\in D:\, q_{c\sigma}(i)>0\}$ and sort it in descending order of $q_{c\sigma}(\cdot)$, ties broken by a fixed rule. Set the per-cell budget $k_{c\sigma}=\lfloor b\%\!\cdot\!|D|/(|\hat C|\!\cdot\!|\hat\Sigma|)\rfloor$, fix any cyclic ordering of cells, and initialize per-cell claim lists $A_{c\sigma}\leftarrow\emptyset$ and the selected pool $\hat D\leftarrow\emptyset$. For each round $t=1,2,\dots$ and each cell $(c,\sigma)$ in the cyclic order: if $|A_{c\sigma}|<k_{c\sigma}$ and $G_{c\sigma}\setminus\hat D\ne\emptyset$, let $i^\star$ be the highest-$q_{c\sigma}$ element of $G_{c\sigma}\setminus\hat D$ and set $A_{c\sigma}\leftarrow A_{c\sigma}\cup\{i^\star\}$ and $\hat D\leftarrow\hat D\cup\{i^\star\}$. Terminate once every cell has $|A_{c\sigma}|=k_{c\sigma}$ or $G_{c\sigma}\subseteq\hat D$. A sample may appear in multiple candidate groups, but once selected it leaves all remaining queues, so each sample is claimed by exactly one cell and $|\hat D|=\sum_{(c,\sigma)}|A_{c\sigma}|\le\sum_{(c,\sigma)} k_{c\sigma}\le b\%\!\cdot\!|D|$, with strict inequality only when a queue exhausts early.

\myparagraph{Guarantees.} Two matching theoretical bounds hold for the round-robin output; formal statements, the lifted ground-set construction they rest on, and full proofs are deferred to Supp.~\ref{sec:app-proofs}.
\begin{itemize}[itemsep=1pt,parsep=0pt,topsep=2pt,leftmargin=*]
\item \emph{$(1-1/e)$ coverage approximation} (Thm.~\ref{thm:approx}). Lifting the per-cell budgets to a partition matroid yields a coverage objective $\widetilde F$ that decomposes additively across cells; an idealized non-competing round-robin coincides with discrete greedy run independently inside each block, so the per-block Nemhauser--Wolsey--Fisher bound~\cite{nemhauser1978analysis} delivers a $(1-1/e)$-factor approximation to the per-cell-budgeted optimum, which the implemented selection inherits by containment (Lem.~\ref{lem:containment}).
\item \emph{EF1 fairness across cells} (Thm.~\ref{thm:ef1}). Treating each cell $(c,\sigma)$ as an additive agent with valuation $v_{c\sigma}(S)=\sum_{i\in S} q_{c\sigma}(i)$, the allocation $\{A_{c\sigma}\}$ produced by round-robin is envy-free up to one item~\cite{lipton2004approximately}: no cell can be starved by more than a single item's value relative to any other. This rules out scalar top-$k$'s failure mode of pouring the entire budget into one cell.
\end{itemize}

\subsection{Profile Sources}
\label{sec:method-sources}

The selection machinery in Sec.~\ref{sec:method-select}--\ref{sec:method-guarantees} consumes only the cell-conditioned value tensor $q$. Any procedure that produces $r_i$ and $g_i$ for every $X_i\in D$ is a valid profile source. We outline two complementary instantiations; both are validated in Sec.~\ref{sec:exp-application}.

\begin{table*}[t]
\footnotesize
\setlength{\tabcolsep}{2pt}
    \centering
    \makebox[\linewidth][c]{%
    \resizebox{1\linewidth}{!}{
    \begin{tabular}{l|ccccccccccaa}
    \toprule
    &MMBench\textsubscript{en-v1.1}& MMStar& MMMU& MMVet& BLINK& MMT-Bench& MME& AI2D& ScienceQA& MathVista\textsubscript{mini} & \textbf{$>$Rand}& \textbf{/FULL}\\

    \midrule
     \multicolumn{13}{c}{\textbf{Budget: 5\%}} \\
    \midrule
Random &   73.74 &   47.98 & \textbf{43.70} &   42.34 &   50.61 &   58.87 & \textbf{2004.50} &   73.07 &   81.52 &   45.47 & - &   89.29\\
PPL-mid &   67.34 &   45.27 &   38.98 &   30.18 &   45.27 &   54.33 &   1887.71 &   66.74 &   74.76 &   31.40 & 0/10 &   78.31\\
PPL-si &   71.98 &   44.67 &   38.48 &   35.14 &  \underline{\textbf{54.10}} &   57.98 &   1856.79 &   67.84 &   78.24 &   36.50 & 1/10 &   83.10\\
Deita\textsubscript{top-$k$} &   72.91 &   47.47 &   41.28 &   40.23 &  \underline{52.59} &   56.57 &   1956.50 &   70.76 &   79.57 &   36.10 & 1/10 &   85.79\\
CLIP & \underline{74.23} &   47.27 &   40.08 &   35.73 & \underline{52.96} &   56.73 &   1902.65 & \underline{73.61} &   78.63 &   39.80 & 3/10 &   85.41\\
E5-V &   70.90 &   43.00 &   38.78 &   38.44 &   49.94 &   54.65 &   1810.47 &   66.58 &   77.54 &   37.40 & 0/10 &   81.87\\
COINCIDE &   72.76 &  \underline{48.33} &   43.17 &  \underline{\textbf{45.60}} &   49.43 &   57.53 &   1852.66 &  \underline{73.15} &   79.62 &   45.40 & 3/10 &   88.45\\
ICONS &   66.72 &  \underline{52.20} &   41.18 &   38.03 &   47.92 &   55.96 &   1811.13 &  \underline{76.20} &  \underline{83.64} &  \underline{46.90} & 4/10 &   86.64\\
PreSel &   72.68 & \underline{48.33} &   40.96 &   35.18 & \underline{52.21} &   55.68 &   1954.35 &   69.01 &   79.38 &   40.70 & 2/10 &   85.08\\
PRISM &   65.56 &   44.07 &   41.07 &   33.49 &   45.90 &   51.84 &   1807.73 &   65.54 &   76.60 &   40.30 & 0/10 &   79.92\\
\rowcolor{ssrhl}\textbf{SSR (Ours)} &  \underline{\textbf{77.79}} &  \underline{\textbf{53.33}} &   43.27 &  \underline{43.53} &  \underline{51.83} &  \underline{\textbf{59.16}} &   1938.68 &  \underline{\textbf{77.66}} &  \underline{\textbf{88.45}} &  \underline{\textbf{52.00}} & \textbf{8/10} &  \underline{\textbf{93.20}}\\

     \midrule 
     \multicolumn{13}{c}{\textbf{Budget: 10\%}} \\
     \midrule  

Random &   74.57 &   51.57 &   44.72 &   42.91 &   52.59 &   58.99 &   2033.28 &   74.42 &   84.33 &   47.80 & - &   91.70\\
PPL-mid &   63.54 &   46.87 &   39.08 &   36.93 &   45.90 &   54.30 &   1831.03 &   67.23 &   73.87 &   39.50 & 0/10 &   80.72\\
PPL-si & \underline{74.69} &   49.80 &   41.28 &   40.60 & \underline{53.09} &   57.95 &   1841.11 & \underline{75.16} &   80.71 &   40.40 & 3/10 &   87.64\\
Deita\textsubscript{top-$k$} & \underline{75.39} &   48.80 &   43.77 &   42.25 & \underline{54.48} &   57.40 &   1996.34 &   71.60 &   78.33 &   40.80 & 2/10 &   88.72\\
CLIP & \underline{75.23} &   49.87 &   40.38 &   37.16 & \underline{53.59} & \underline{59.35} &   1921.04 & \underline{76.62} &   80.07 &   41.00 & 4/10 &   87.69\\
E5-V &   70.51 &   45.13 &   38.78 &   39.59 &   50.57 &   55.10 &   1787.94 &   68.94 &   77.54 &   37.20 & 0/10 &   82.76\\
COINCIDE & \underline{75.23} &   49.73 & \underline{44.77} &   42.52 &   50.69 &   58.71 &   2027.58 & \underline{74.77} &   82.05 &   47.00 & 3/10 &   90.66\\
ICONS &   71.67 & \underline{53.33} &   44.17 &   40.46 &   49.18 &   57.40 &   1789.60 & \underline{76.65} & \underline{85.23} & \underline{51.10} & 4/10 &   89.91\\
PreSel & \underline{76.93} & \underline{\textbf{53.87}} &   42.82 &   33.94 & \underline{\textbf{55.23}} &   57.28 &   1904.98 & \underline{\textbf{79.44}} & \underline{\textbf{92.86}} & \underline{50.40} & 6/10 &   91.61\\
PRISM &   64.63 &   45.47 &   41.31 &   30.78 &   45.65 &   53.21 &   1849.71 &   66.13 &   76.20 &   42.10 & 0/10 &   80.23\\
\rowcolor{ssrhl}\textbf{SSR (Ours)} &  \underline{\textbf{77.32}} &  \underline{53.27} &  \underline{\textbf{45.06}} &  \underline{\textbf{42.98}} &  \underline{54.10} &  \underline{\textbf{59.61}} &  \underline{\textbf{2045.00}} &  \underline{78.76} &  \underline{89.94} &  \underline{\textbf{52.40}} & \textbf{10/10} &  \underline{\textbf{94.75}}\\

     \midrule 
     \multicolumn{13}{c}{\textbf{Budget: 30\%}} \\
     \midrule  

Random &   78.25 &   54.60 &   44.40 &   46.10 &   55.23 &   59.61 &   2092.60 &   78.28 &   88.32 &   52.57 & - &   95.82\\
PPL-mid &   73.99 &  \underline{54.93} &   43.97 &   41.01 &   53.09 &   58.78 &   2036.54 &   77.20 &   87.01 &  \underline{56.40} & 2/10 &   93.77\\
PPL-si &   72.52 &   48.33 &   42.57 &   43.62 &   51.83 &   55.07 &   1976.46 &   76.55 &   78.48 &   42.20 & 0/10 &   88.22\\
Deita\textsubscript{top-$k$} &   76.93 &   54.13 &   43.67 &   44.04 &   55.11 & \underline{59.66} &   2042.63 & \underline{79.50} &   83.54 &   50.30 & 2/10 &   93.99\\
CLIP &   74.30 &   53.80 &   43.07 &   45.87 &   51.95 &   59.16 &   2039.14 & \underline{80.02} &   83.99 &   48.80 & 1/10 &   93.07\\
E5-V &   74.30 &   46.07 &   43.27 & \underline{47.80} &   50.32 &   57.85 &   1955.13 &   74.45 &   81.61 &   43.70 & 1/10 &   89.53\\
COINCIDE &   78.02 & \underline{55.47} & \underline{\textbf{45.66}} & \underline{46.24} &   52.84 & \underline{59.80} &   2047.37 & \underline{79.73} &   84.33 & \underline{55.10} & 6/10 &   95.82\\
ICONS &   71.90 &   53.40 &   43.87 &   42.25 &   50.32 &   59.23 &   1985.64 &   78.21 &   86.76 & \underline{54.10} & 1/10 &   92.55\\
PreSel &   76.86 & \underline{57.20} &   41.89 &   36.93 & \underline{\textbf{59.14}} &   58.91 &   1971.60 & \underline{\textbf{81.61}} & \underline{\textbf{93.75}} & \underline{55.20} & 5/10 &   95.02\\
PRISM &   64.32 &   49.60 &   41.77 &   40.60 &   46.66 &   54.01 &   1898.22 &   70.43 &   81.26 &   47.40 & 0/10 &   85.57\\
\rowcolor{ssrhl}\textbf{SSR (Ours)} &  \underline{\textbf{79.57}} &  \underline{\textbf{57.53}} &  \underline{44.87} &  \underline{\textbf{48.49}} &  \underline{56.24} &  \underline{\textbf{59.83}} &  \underline{\textbf{2132.93}} &  \underline{81.25} &  \underline{92.46} &  \underline{\textbf{57.40}} & \textbf{10/10} &  \underline{\textbf{99.11}}\\

     \midrule 
     \multicolumn{13}{c}{\textbf{FULL}} \\
     \midrule  

LLaVA-OVSI & 80.57 & 59.40 & 45.16 & 47.16 & 56.87 & 60.73 & 2117.56 & 81.87 & 92.76 & 59.60 & -& 100\\

    \bottomrule
    \end{tabular}
    }}
    \vspace{-0.1in}
    \caption{\textbf{Million-scale scalability and benchmark-wide generality.} Performance on multi-modal benchmarks across 5\%, 10\% and 30\% budgets of LLaVA-OVSI; best result in \textbf{boldface}, \underline{underline} if a method beats the random baseline. The {$>$Rand} column reports the number of benchmarks where the method outperforms random sampling; {/FULL} compares to the FULL dataset. Among all evaluated methods, SSR is the only one that leads the $>$Rand metric at every budget, attaining a clean \textbf{10/10} sweep at the 10\% and 30\% budgets and recovering \textbf{99.1\%} of full-data performance with 30\% data.}
    \label{tab:main}
    \vspace{-0.3cm}
\end{table*}

\myparagraph{Direct judge query} (no finetuning). A capable judge (a proprietary API such as GPT-4o, an open-weight MLLM such as Qwen2-VL, or an in-house fine-tuned MLLM checkpoint) is prompted with a fixed rubric to emit $r_i$ and $g_i$ for every $X_i\in D$. This path requires no training; its cost scales linearly in $|D|$, with per-call price or local inference compute as the constant factor.

\myparagraph{Distilled judge } (optional, separately trained). When the judge cost on the full pool $D$ exceeds the available annotation budget, we amortize annotations by fine-tuning two \emph{independent} student MLLMs on a small judge-labeled subset $D_J\subset D$, 
following the standard visual-instruction-tuning recipe~\cite{li2024LLaVAOneVisionEasyVisual} under the next-token-prediction objective
\vspace{-0.08in}
\begin{equation}
\mathcal{L}_{i}\;=\;-\sum_t\log p_\theta\,\bigl(y_{i,t}^J\,\big|\,y_{i,<t}^J,\,X_i\bigr),
\vspace{-0.12in}
\label{eq:distill}
\end{equation}
where $y_i^J$ is the score string and the style string. At inference, $\hat r_i$ and $\hat g_i$ are 
combined via Eq.~\ref{eq:q} into $q$. 
We query GPT-4o on $|D_J|/|D|\,\approx\!15\%$ data for distillation, and validate this choice on Supp.~\ref{sec:sup-score}.

\section{Experiments}
\vspace{-0.05in}

Sec.~\ref{sec:exp-main} reports million-scale selection on the 2.6M multi-modal data pool, where, among all evaluated methods, SSR is the only one that scales and delivers consistent benchmark-wide gains over prior baselines.
Sec.~\ref{sec:exp-ablate} dissects the richness mechanism behind these gains and probes robustness to annotator choices.
Sec.~\ref{sec:exp-transfer} demonstrates training-free transfer of SSR across architectures and data pools, including zero-shot transfer to a third architecture family on a 7M pool. 
Additional details (Supp.~\ref{sec:sup-exp-details}), SSR model fidelity (Supp.~\ref{sec:sup-score}), capability scaling (Supp.~\ref{sec:sup-scale}), results in 6 budget settings (Supp.~\ref{sec:sup-scale}), and efficiency (Supp.~\ref{sec:sup-efficiency})
are deferred to appendix.

\begin{table*}[t]
\footnotesize
\setlength{\tabcolsep}{1pt}
    \centering
    \resizebox{\linewidth}{!}{
    \begin{tabular}{l@{\hspace{3pt}}|cc|ccccccc@{\hspace{4pt}}cccaa}
    \toprule
    & R\textsubscript{scr} & R\textsubscript{sty} &MMBench\textsubscript{en-v1.1}& MMStar& MMMU& MMVet& BLINK& MMT-Bench& MME& AI2D& ScienceQA& MathVista\textsubscript{mini} & \textbf{$>$Rand} & \textbf{/FULL}\\

     \midrule 
     \multicolumn{15}{c}{\textbf{Budget: 5\%}} \\
     \midrule  
     Random & &  &  73.74  &  47.98  &  43.70  &  42.34  &  50.61  &  58.87  &  \textbf{2004.50}  &  73.07  &  81.52  &  45.47  & - &  89.29\\
SP\textsubscript{GPT-4o} &   &     &  \underline{75.85}  &  \underline{51.27}  &  42.97  &  \underline{\textbf{44.27}}  &  \underline{51.95}  &  58.14  &  1940.27  &  \underline{73.61}  &  81.46  &  45.00  & 5/10 &  \underline{90.14}\\
ScrR\textsubscript{GPT-4o} &  \checkmark  &     &  \underline{74.23}  &  \underline{49.07}  &  43.07  &  41.56  &  \underline{52.08}  &  57.56  &  1996.27  &  \underline{74.77}  &  80.12  &  44.80  & 4/10 &  89.17\\
StyR\textsubscript{GPT-4o} &   &  \checkmark   &  \underline{76.48}  &  \underline{52.52}  &  \underline{43.85}  &  41.58  &  \underline{52.46}  &  \underline{58.89}  &  1945.68  &  \underline{76.64}  &  \underline{86.88}  &  \underline{49.87}  & \textbf{8/10} &  \underline{92.07}\\
SSR\textsubscript{Qwen2-VL}  &  \checkmark  &  \checkmark   &  \underline{75.08}  &  \underline{51.00}  &  \underline{\textbf{45.16}}  &  \underline{42.57}  &  \underline{52.71}  &  57.37  &  1955.78  &  \underline{74.74}  &  \underline{84.88}  &  \underline{48.90}  & \textbf{8/10} &  \underline{91.37}\\
SSR\textsubscript{LLaVA-OVSI}  &  \checkmark  &  \checkmark   &  \underline{77.40}  &  \underline{50.60}  &  \underline{44.77}  &  41.10  &  \underline{\textbf{54.35}}  &  58.62  &  1952.97  &  \underline{75.81}  &  \underline{87.75}  &  40.40  & 6/10 &  \underline{90.68}\\
\rowcolor{ssrhl}SSR\textsubscript{GPT-4o}  &  \checkmark  &  \checkmark   &  \underline{\textbf{77.79}}  &  \underline{\textbf{53.33}}  &  43.27  &  \underline{43.53}  &  \underline{51.83}  &  \underline{\textbf{59.16}}  &  1938.68  &  \underline{\textbf{77.66}}  &  \underline{\textbf{88.45}}  &  \underline{\textbf{52.00}}  & \textbf{8/10} &  \underline{\textbf{93.20}}\\

    \midrule
     \multicolumn{15}{c}{\textbf{Budget: 10\%}} \\
    \midrule
Random & &  &  74.57  &  51.57  &  44.72  &  42.91  &  52.59  &  58.99  &  2033.28  &  74.42  &  84.33  &  47.80  & - &  91.70\\
SP\textsubscript{GPT-4o} &   &     &  \underline{77.24}  &  50.40  &  44.27  &  42.52  &  \underline{53.47}  &  \underline{59.48}  &  \underline{\textbf{2084.39}}  &  \underline{76.07}  &  81.36  &  46.10  & 5/10 &  \underline{91.73}\\
ScrR\textsubscript{GPT-4o} &  \checkmark  &     &  73.76  &  49.40  &  \underline{44.77}  &  42.80  &  46.91  &  57.24  &  2000.17  &  \underline{75.39}  &  83.79  &  44.40  & 2/10 &  89.27\\
StyR\textsubscript{GPT-4o} &   &  \checkmark   &  \underline{77.72}  &  \underline{54.36}  &  44.17  &  \underline{44.62}  &  \underline{54.05}  &  \underline{59.60}  &  1928.81  &  \underline{78.66}  &  \underline{89.75}  &  \underline{\textbf{52.80}}  & 8/10 &  \underline{94.61}\\
SSR\textsubscript{Qwen2-VL}  &  \checkmark  &  \checkmark   &  \underline{76.24}  &  \underline{53.33}  &  \underline{44.87}  &  \underline{\textbf{45.60}}  &  \underline{\textbf{55.11}}  &  \underline{59.16}  &  2012.94  &  \underline{76.75}  &  \underline{87.11}  &  \underline{52.70}  & 9/10 &  \underline{94.59}\\
SSR\textsubscript{LLaVA-OVSI}  &  \checkmark  &  \checkmark   &  \underline{\textbf{77.79}}  &  \underline{\textbf{54.40}}  &  44.67  &  42.02  &  \underline{54.98}  &  58.23  &  2013.74  &  \underline{\textbf{78.85}}  &  \underline{89.59}  &  42.00  & 5/10 &  \underline{92.72}\\
\rowcolor{ssrhl}SSR\textsubscript{GPT-4o}  &  \checkmark  &  \checkmark   &  \underline{77.32}  &  \underline{53.27}  &  \underline{\textbf{45.06}}  &  \underline{42.98}  &  \underline{54.10}  &  \underline{\textbf{59.61}}  &  \underline{2045.00}  &  \underline{78.76}  &  \underline{\textbf{89.94}}  &  \underline{52.40}  & \textbf{10/10} &  \underline{\textbf{94.75}}\\

     \midrule 
     \multicolumn{15}{c}{\textbf{Budget: 30\%}} \\
     \midrule  

Random & &  &  78.25  &  54.60  &  44.40  &  46.10  &  55.23  &  59.61  &  2092.60  &  78.28  &  88.32  &  52.57  & - &  95.82\\
SP\textsubscript{GPT-4o} &   &     &  77.86  &  53.13  &  \underline{\textbf{45.76}}  &  \underline{48.03}  &  54.85  &  58.78  &  2050.69  &  \underline{78.92}  &  86.91  &  \underline{55.80}  & 4/10 &  \underline{96.31}\\
ScrR\textsubscript{GPT-4o} &  \checkmark  &     &  77.09  &  52.67  &  43.47  &  44.31  &  53.59  &  58.23  &  2024.57  &  \underline{79.11}  &  87.90  &  52.20  & 1/10 &  93.93\\
StyR\textsubscript{GPT-4o} &   &  \checkmark   &  \underline{78.27}  &  \underline{55.84}  &  42.87  &  43.11  &  54.43  &  59.44  &  2079.25  &  \underline{80.42}  &  \underline{92.15}  &  \underline{55.96}  & 5/10 &  \underline{96.07}\\
SSR\textsubscript{Qwen2-VL}  &  \checkmark  &  \checkmark   &  78.02  &  \underline{57.13}  &  43.07  &  \underline{47.39}  &  \underline{55.49}  &  \underline{\textbf{60.89}}  &  \underline{2096.60}  &  \underline{\textbf{81.64}}  &  \underline{90.28}  &  \underline{\textbf{57.40}}  & 8/10 &  \underline{97.91}\\
SSR\textsubscript{LLaVA-OVSI}  &  \checkmark  &  \checkmark   &  77.55  &  54.53  &  43.37  &  44.72  &  55.23  &  58.59  &  1980.48  &  \underline{81.02}  &  \underline{91.87}  &  49.60  & 2/10 &  94.73\\
\rowcolor{ssrhl}SSR\textsubscript{GPT-4o}  &  \checkmark  &  \checkmark   &  \underline{\textbf{79.57}}  &  \underline{\textbf{57.53}}  &  \underline{44.87}  &  \underline{\textbf{48.49}}  &  \underline{\textbf{56.24}}  &  \underline{59.83}  &  \underline{\textbf{2132.93}}  &  \underline{81.25}  &  \underline{\textbf{92.46}}  &  \underline{\textbf{57.40}}  & \textbf{10/10} &  \underline{\textbf{99.11}}\\

    \bottomrule
    \end{tabular}
    }
    \vspace{-0.1in}
    \caption{\textbf{Richness drives generality, robustly across annotator sources.} Both rich scores (ScrR) and styles (StyR) are necessary for accuracy and generality: dropping either degrades coverage, and the single-criterion, \ie \textit{quality} SP(oor) baseline fails to cover the diverse axes of evaluation. \{method\}\textsubscript{\{source\}} indicates querying scores and styles from \{source\} for \{method\}. With either distilled proxies (SSR\textsubscript{GPT-4o}) or direct  open-weight model prompting (SSR\textsubscript{Qwen2-VL}, SSR\textsubscript{LLaVA-OVSI}), SSR consistently beats prior baselines, confirming the efficacy stems from the explicit, rich SSR formulation that is robust to annotators.}
    \label{tab:analysis}
    \vspace{-0.25cm}
\end{table*}

\subsection{Experimental Setup}
\label{sec:setup}
\vspace{-0.05in}
\myparagraph{Data pool.} We base our main experiments on the single-image SFT stage of LLaVA-OneVision (LLaVA-OVSI)~\cite{li2024LLaVAOneVisionEasyVisual}, the open-source, open-data SOTA MLLM series. Within its 3.2M high-quality instances\footnote{https://huggingface.co/datasets/lmms-lab/LLaVA-OneVision-Data}, 2.6M multi-modal data are openly available, which we consider as the full data pool and perform sample selection on it.
This well-curated dataset covers over 90 sources, encompassing natural images, math and reasoning questions, documents, charts, screenshots \etc.
For our transfer experiments (Sec.~\ref{sec:exp-transfer}), the earlier ShareGPT4V~\cite{chen2024sharegpt4v} and the Cambrian-7M pool~\cite{tong2024cambrian} are adopted.

\myparagraph{Target model.} SSR is evaluated in three distinct architectures, LLaVA-OneVision (Qwen2 + SigLIP), LLaVA-1.5 (Vicuna + CLIP), and Cambrian-1 (Llama-3 + a four-encoder ensemble).

\myparagraph{Training setup.} We utilize the stage-1.5-7B checkpoint
of LLaVA-OneVision as the pretrained model for both SSR finetuning and single-image instruction tuning. 
To mitigate computational overhead during our extensive comparative experiments, we limit the maximum token length to 12k, which allows all training runs to be executed on 64 Nvidia H100 GPUs with a global batch size of 128. Beyond this modification, we strictly adhere to the hyperparameter settings and training protocols established in the official implementation.
For our transfer experiments (Sec.~\ref{sec:exp-transfer}), we adopt the LLaVA-1.5-7B architecture (Vicuna + CLIP)~\cite{liu2023llava15} as the base model. Likewise, consistency is maintained by ensuring the fine-tuning procedure adheres strictly to the original protocols.

\myparagraph{Our configuration.} 
We consider the full spectrum of capabilities in our sampling, excluding OCR, which is reserved to validate capability scalability in Supp.~\ref{sec:sup-scale}. While SSR supports diverse judge sources, our main experiments utilize specialized models fine-tuned on 15\% GPT-4o annotations to infer on the entire pool. We further explore and analyze alternative deployment strategies across different resource settings in Sec.~\ref{sec:exp-application}.

\myparagraph{Baselines.} We compare SSR with 10 methods spanning random sampling, perplexity (PPL-mid, PPL-si), the score- and embedding-based LLM selector Deita~\cite{liu2024Deita}\footnote{Deita's $O(N^2)$ similarity-threshold diversity filter does not scale to 2.6M instances; we run its score-only variant Deita\textsubscript{top-$k$}, ranking by its scores (Supp.~\ref{sec:sup-baselines}).}, VL alignment metrics CLIP~\cite{radford2021clip} and E5-V~\cite{jiang2024E5VUniversalEmbeddings}, and MLLM SOTAs COINCIDE~\cite{coincide}, ICONS~\cite{wu2024icons}, PreSel~\cite{presel2025}, and PRISM~\cite{bi2025prism}, all under their official configurations (Supp.~\ref{sec:sup-baselines}).

\myparagraph{Evaluation benchmarks.} We evaluate on 14 multi-modal benchmarks with VLMEvalKit~\cite{duan2024vlmevalkit}; the full list and protocol are in Supp.~\ref{sec:sup-eval-bench}.

\subsection{Selection at Million-Scale}
\label{sec:exp-main}
\vspace{-0.05in}
\myparagraph{Million-scale scalability and benchmark-wide generality.} As shown in Tab.~\ref{tab:main}, among all evaluated methods SSR is the only one that scales to the 2.6M LLaVA-OVSI pool \textit{and} delivers generality across all benchmarks. It recovers 99.1\% of full-data performance at 30\% and remains robust from cold-start (5\%) to warm-start (30\%) regimes, regardless of the specific evaluation focus. The $>$Rand column makes the generality point sharp: at the 10\% and 30\% budgets SSR beats the best competitor, \ie random sampling on \textbf{10/10} benchmarks, while every competitor wins on at most 6 of the 10, validating that latents and narrow criteria fail to cover the diverse axes of MLLM evaluation. 

\begin{figure*}[t]
    \centering
    \vspace{-0.1in}
    \begin{minipage}[t]{0.32\textwidth}
        \centering
        \includegraphics[width=.93\linewidth]{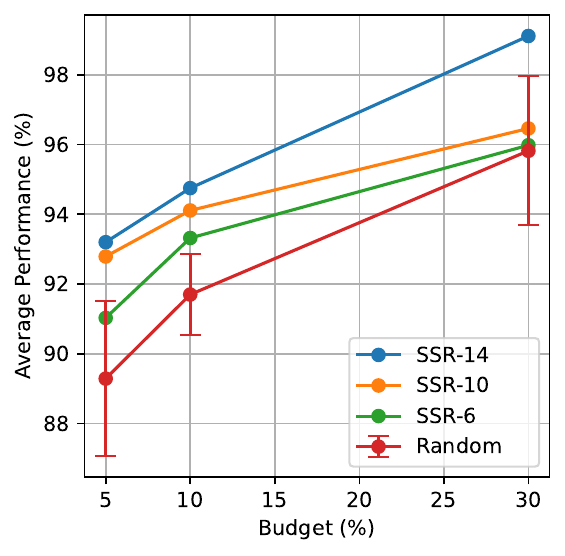}
        \captionof{figure}{\textbf{Richness governs informativeness.} Randomly cutting the capabilities set from full to 10 to 6 monotonically degrades performance, yet they still significantly outperform random sampling. Dimensional richness drives the gain, demonstrating robustness to composition.}
        \label{fig:cap-down}
    \end{minipage}
    \hfill
    \begin{minipage}[t]{0.32\textwidth}
        \centering
        \includegraphics[width=\linewidth, trim={2.2cm 2.8cm 2cm 2.1cm}, clip]{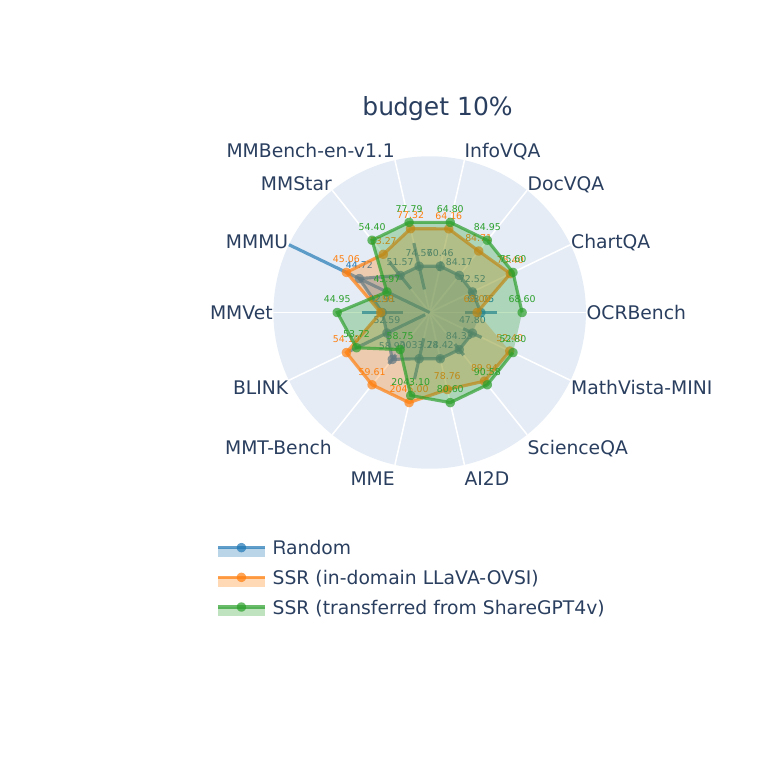}
        \captionof{figure}{\textbf{Training-free cross-domain SSR model transfer.} SSR models finetuned on 10\% of ShareGPT4V (12 sources, {\scriptsize Vicuna-CLIP}) directly sample from the LLaVA-OVSI pool (91 sources, {\scriptsize Qwen2-SigLIP}), identifying informative data under unseen distributions and target model.}
        \label{fig:radar-transfer}
    \end{minipage}
    \hfill
    \begin{minipage}[t]{0.31\textwidth}
        \centering
        \includegraphics[width=\linewidth, trim={2.5cm 2.8cm 2cm 2.1cm}, clip]{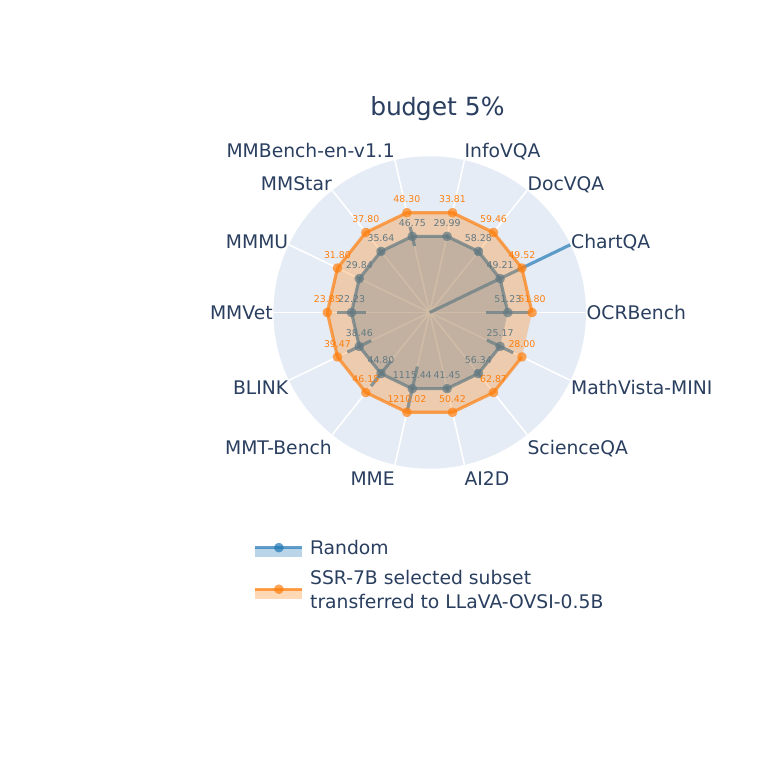}
        \captionof{figure}{\textbf{Cross-model selected subset transfer.} The 5\% subset selected by 7B distilled SSR models remains superior when used to train a 0.5B LLaVA-OVSI variant, confirming the explicit profile captures intrinsic data value robust to target-model scale and architecture.}
        \label{fig:radar-transfer-0.5B}
    \end{minipage}
    \vspace{-0.25cm}
\end{figure*}

\myparagraph{Competitor Analysis.} LLaVA-OV mid-stage model lacks instruction tuning, rendering its perplexity scores uninformative; the fully fine-tuned SFT checkpoint (LLaVA-OVSI) helps marginally but defeats the purpose of selection, which is to reduce training cost before full fine-tuning.
Deita's textual scorers show that narrow, text-centric metrics such as \textit{quality} and \textit{complexity} do not transfer to the multi-modal domain.
VL embedding-based methods (CLIP, E5-V), while capable of encoding both modalities, optimize for contrastive image-text alignment that is misaligned with the generative instruction-following objective of MLLMs.
Finally, methods validated on the thousand-scale LLaVA pool all prove brittle to the scale and distribution shifts of a 4$\times$-larger, heterogeneous pool, confirming that model-anchored cluster or influence signals do not scale and generalize. For example, recent PRISM's intrinsic-feature scores are computed against a single corpus-level reference and thresholded per sample, so on a pool spanning 90+ sources the ranking reflects global distribution shift more than local redundancy.

\subsection{Richness Drives Generality Robustly}
\label{sec:exp-ablate}
\vspace{-0.05in}
\myparagraph{Generality stems from richness}, not from a particular criterion set or annotator.
Why does SSR scale successfully to million-scale and generalize robustly across benchmarks? We attribute it to the \textit{richness} of the capability--style profile, then verify this attribution against two practical perturbations: changes in the criterion set, and changes in the annotator that produced the profile.

\noindent\textit{Are both score- and style-richness necessary?} The SSR profile combines rich capability scores with rich interaction styles. Tab.~\ref{tab:analysis} ablates the two: Score-rich (ScrR) and Style-rich (StyR) each lift performance over random sampling, but only their combination delivers benchmark-wide gains: scores select high-value knowledge while styles ensure that knowledge is delivered through diverse forms. As an ablation, the SP(oor) uses a single  \textit{quality} metric; its narrow and inconsistent benchmark coverage confirms that a one-dim criterion cannot cover the diverse axes of MLLM evaluation. Rich scores and rich styles are jointly necessary.

\noindent\phantomsection\label{sec:exp-richness}\textit{Does richness require an optimal criterion composition?} If SSR depended on finding an NP-hard-optimal subset of capabilities, the method would be brittle. Fig.~\ref{fig:cap-down} rules this out: randomly cutting the criteria set from full to 10 to 6 capabilities monotonically degrades performance, yet every random subset still significantly outperforms random sampling. What matters is the \textit{richness} of data dimensions, not a singular optimal composition.

\noindent\phantomsection\label{sec:exp-application}\textit{Does richness require a high-end annotator?} Since the SSR profile is read explicitly from each sample, the framework should decouple from the annotator that produced it. We test two configurations sharing the same selection method and identical query prompts: SSR\textsubscript{GPT-4o} fine-tunes proxy models on 15\% GPT-4o annotations to infer over the full pool, while SSR\textsubscript{\{source\}} in Tab.~\ref{tab:analysis} directly leverages the instruction-following capability of \{source\} (Qwen2-VL-7B or LLaVA-OVSI-7B). Results show that specialized proxies remain optimal; yet direct querying with open-weight annotators already consistently outperforms the best competing baselines. SSR is thus robust to judge sources, with specialized proxies for peak quality and open-weight prompting for API-cost-free use.

\subsection{Cross-Datapool and -Model Transfer}
\label{sec:exp-transfer}
\vspace{-0.05in}
\myparagraph{Training-free cross-domain SSR model transfer.} SSR distilled scorer and styler models finetuned on one architecture with one data pool transfer zero-shot to another, with no retraining and no gradient update. To simulate practical curation under evolving data distributions, we finetune scorer and styler (Vicuna + CLIP) on a 10\% random subset of ShareGPT4V~\cite{chen2024sharegpt4v} containing only \textit{12 sources}, then apply them \emph{directly} to infer and sample from the 4$\times$-larger LLaVA-OVSI dataset~\cite{li2024LLaVAOneVisionEasyVisual} spanning \textit{91 distinct sources}. Fig.~\ref{fig:radar-transfer} shows SSR still identifies high-value data despite the unseen knowledge distribution and target model, because the SSR profile describes sample-intrinsic semantics that remain objective and meaningful when the settings shift.

\myparagraph{Cross-model selected subset transfer.} High-quality data should be universally beneficial rather than coupled to a particular architecture or training setup. To test this, we feed the subset selected by our 7B distilled SSR models into training a 14$\times$-smaller model. As shown in Fig.~\ref{fig:radar-transfer-0.5B}, SSR's superiority persists, confirming that the explicit profile captures intrinsic data value that is robust to the target model's scale and architecture.

\begin{table}[t]
\centering
\small
\setlength{\tabcolsep}{6pt}
\begin{tabular}{l|ccc}
\toprule
Cambrian-7M & 5\% & 10\% & 30\% \\
\midrule
Random / FULL & 90.54 & 93.57 & 97.85 \\
\rowcolor{ssrhl}\textbf{SSR (Ours)} / FULL & \textbf{92.40} & \textbf{94.35} & \textbf{98.19} \\
\bottomrule
\end{tabular}
\vspace{-0.1in}
\caption{\textbf{Zero-shot transfer to Cambrian-1-8B on the Cambrian-7M pool.} LLaVA-OVSI-distilled SSR models profile the unseen pool without adaptation; /FULL (\%) on the benchmark suite of Tab.~\ref{tab:main}.}
\label{tab:cambrian}
\vspace{-0.3cm}
\end{table}

\myparagraph{Zero-shot transfer to a third architecture family at 7M scale.} Stressing both transfers at once, we apply the SSR models distilled in Sec.~\ref{sec:exp-main} \emph{zero-shot} to the unseen 7M Cambrian-7M pool and train Cambrian-1-8B~\cite{tong2024cambrian} on the selected subsets under its official recipe (Supp.~\ref{sec:sup-cambrian}). Cambrian-1 is fully open across all training stages, permitting a controlled full-data reference, and is disjoint from both stacks above on the vision side (a four-encoder ensemble) and the language side (Llama-3-8B). Tab.~\ref{tab:cambrian} shows that SSR outperforms random sampling at every budget (+1.86, +0.78 and +0.34 points at 5\%, 10\% and 30\%), with the largest gain in the cold-start regime where selection matters most, and recovers 98.2\% of full-data performance with 30\% of the pool. The transfer is demanding along three axes at once, with an unseen pool 2.7$\times$ larger than the distillation pool, a new source mixture, and a new architecture family, yet needs no re-annotation, fine-tuning, or change to the taxonomy: nothing in the profile depends on the target backbone, so it carries over unchanged. 

Prior MLLM selectors are typically validated on a single LLaVA-1.5-era stack with LoRA fine-tuning. SSR's models and taxonomy transfer unchanged across architecture families, data pools of up to 7M samples, and standard full-parameter SFT: distill a selector once, and it can be reused directly as backbones and pools evolve, precisely the setting where curation decisions are made.

\section{Conclusions}
\label{sec:conclusion_limitation}
\vspace{-0.05in}

We introduced SSR, shifting multi-modal data selection from narrow model-centric signals to rich, sample-intrinsic criteria: each sample is profiled by an interaction style (\textit{how}) and a capability score vector (\textit{what}), and selection is cast as submodular maximization with theoretical coverage and EF1 guarantees. As the first principled multi-modal SFT selector at million-sample scale, SSR delivers benchmark-wide gains across three MLLM architecture families, data pools of up to 7M samples, 6 budgets, and 14 benchmarks, and transfers training-free across data pools and architectures. Its gains stem from the richness of our framework, which is robust to exact composition or annotator, positioning data-centric profiling as a practical lever as MLLM pools outgrow today's curation pipelines.

\section*{Limitations}
\label{sec:limitations}

SSR is validated across three architecture families (LLaVA-1.5, LLaVA-OneVision, Cambrian-1), three data pools of up to 7M samples, 6 budgets, 14 benchmarks, and three complementary judge sources, with formal $(1-1/e)$ and EF1 guarantees on the coverage surrogate; we note the boundaries of this evidence.
Our target backbones span 0.5B--8B parameters, the range in which controlled full-data references remain affordable and in which prior MLLM selection work is validated. Since the profile carries no model-intrinsic signal, we expect its gains to carry over to larger backbones, but a controlled study at the 72B scale is left to future work.
Our profiles depend on an MLLM judge. Gains are consistent across proprietary, open-weight, and distilled judges, and because every score is explicit and human-readable, judge bias is auditable and correctable rather than hidden in a latent space; still, the absolute quality of a selected subset on rare or highly specialized capabilities is bounded by the judge, and we did not sweep judge choices under matched annotation budgets.
Likewise, our capability--style taxonomy is broad and open to extension without retraining the selector, but we do not claim it is uniquely optimal, and characterizing its returns on emerging domains (\eg agentic tool use, long-horizon dialogue) and modalities (\eg video, audio) is a promising direction we leave to future exploration.

\section*{Ethics Statement}
\label{sec:ethics}

This work studies data selection for multi-modal instruction tuning and introduces no new human-subject data collection. All datasets (ShareGPT4V, LLaVA-OneVision, Cambrian) are publicly released by their authors and used here in accordance with their licenses for non-commercial academic research; we did not attempt to re-identify depicted individuals or augment the pools with private data.
Because SSR \emph{selects from} pre-existing pools rather than generating new content, any social, demographic, or linguistic biases in the source pool will persist in, and may be reweighted by, the curated subset. The capability--style coverage objective broadens representational diversity along instruction-relevant axes but does not target demographic fairness, and we make no fairness claim along protected-attribute dimensions; downstream users should audit content-level risks separately.
We query GPT-4o for distillation labels under OpenAI's research terms; its outputs are used only as supervisory signals for our distilled SSR models and are not redistributed. SSR is intended as a research tool to make MLLM instruction tuning more efficient and transparent; the interpretability of its capability and style is designed to make misuse easier, not harder, to detect.

\section*{Acknowledgements}
This work was partly supported by National Natural Science Foundation of China No. 62525103 and Beijing Natural Science Foundation No. L223023.

\bibliography{bib}

\appendix

\clearpage

\section{Proofs}
\label{sec:app-proofs}

This section supplies the formal statements and proofs underlying the guarantees claimed in Sec.~\ref{sec:method-guarantees}.

\subsection{Lifted ground set and partition matroid}
\label{sec:app-lift}

To analyze the stratified round-robin algorithm of Sec.~\ref{sec:method-guarantees}, we lift the pool to the instance--cell pairs
\begin{equation*}
  \widetilde V\;=\;\{(i,c,\sigma):\,i\in G_{c\sigma}\},
\end{equation*}
and define the lifted coverage objective
\begin{equation}
  \widetilde F(\widetilde S)\;=\;\sum_{(c,\sigma)} w_{c\sigma}\min\!\Bigl\{\tau_{c\sigma},\sum_{(i,c,\sigma)\in\widetilde S} q_{c\sigma}(i)\Bigr\}.
\label{eq:app-Ftilde}
\end{equation}
Let $\widetilde V_{c\sigma}$ denote the block of pairs in cell $(c,\sigma)$. The per-cell budgets $\{k_{c\sigma}\}$ induce the partition matroid
\begin{equation}
\widetilde{\mathcal{I}}\;=\;\bigl\{\widetilde S\subseteq\widetilde V:\;|\widetilde S\cap\widetilde V_{c\sigma}|\le k_{c\sigma}\ \forall(c,\sigma)\bigr\},
\label{eq:app-matroid}
\end{equation}
which is clean regardless of overlap among the original $G_{c\sigma}$. Cells with $G_{c\sigma}=\emptyset$ contribute no pairs and are dropped from the grid, so $\tau_{c\sigma}>0$ for every remaining cell.

We write $\widetilde S_{\mathrm{RR}}\subseteq\widetilde V$ for the \emph{idealized} round-robin assignment in which every cell independently claims the top-$k_{c\sigma}$ pairs of its own block $\widetilde V_{c\sigma}$ in descending order of $q_{c\sigma}$ (all of them if $|\widetilde V_{c\sigma}|<k_{c\sigma}$), under the same fixed tie-breaking rule as the selector; its projection onto the pool is $\hat D_{\mathrm{ideal}}=\{i:\exists(c,\sigma),\,(i,c,\sigma)\in\widetilde S_{\mathrm{RR}}\}$. The implemented selector of Sec.~\ref{sec:method-guarantees} differs in one respect: a sample selected by one cell is removed from every other queue, so each sample is claimed exactly once. The next lemma transfers guarantees proven for $\widetilde S_{\mathrm{RR}}$ to the implemented output $\hat D$.

\begin{lemma}[Containment]
\label{lem:containment}
Let $\hat D$ be the pool selected by the implemented stratified round-robin. Then $\hat D\supseteq\hat D_{\mathrm{ideal}}$, and consequently
\begin{equation*}
F(\hat D)\;\ge\;F(\hat D_{\mathrm{ideal}})\;\ge\;\widetilde F(\widetilde S_{\mathrm{RR}}).
\end{equation*}
\end{lemma}

\noindent\emph{Proof.} Fix $(i,c,\sigma)\in\widetilde S_{\mathrm{RR}}$, i.e., $i$ ranks within the top $k_{c\sigma}$ of $G_{c\sigma}$. If $i$ is selected by any cell during the implemented run, then $i\in\hat D$. Otherwise $i$ stays in cell $(c,\sigma)$'s queue throughout; removals by other cells only promote its rank, which starts at $\le k_{c\sigma}$, and the cell keeps drawing from its non-empty queue until it holds $k_{c\sigma}$ samples, so it would claim $i$ within its first $k_{c\sigma}$ picks: a contradiction. Hence $\hat D\supseteq\hat D_{\mathrm{ideal}}$. The first inequality then follows from the monotonicity of $F$ (Prop.~\ref{prop:submodular}); the second holds because, cell by cell, $\sum_{i\in\hat D_{\mathrm{ideal}}}q_{c\sigma}(i)\ge\sum_{(i,c,\sigma)\in\widetilde S_{\mathrm{RR}}}q_{c\sigma}(i)$ with $q\ge0$ and $x\mapsto\min\{\tau_{c\sigma},x\}$ non-decreasing.

\subsection{Submodularity of $F$ (Prop.~\ref{prop:submodular})}
\label{sec:app-proof-prop1}

\noindent\emph{Proof.} Each cell term $F_{c\sigma}(S)=w_{c\sigma}\min\{\tau_{c\sigma},\sum_{i\in S} q_{c\sigma}(i)\}$ is the composition $F_{c\sigma}=\varphi\circ m$ of the non-decreasing concave function $\varphi(x)=w_{c\sigma}\min\{\tau_{c\sigma},x\}$ with the modular (additive) sum $m(S)=\sum_{i\in S} q_{c\sigma}(i)$; such a composition is monotone submodular~\cite{lin2011classsubmodular}. Explicitly, for $S\subseteq T\subseteq D$ and $i\notin T$, we have $m(S)\le m(T)$ since $q_{c\sigma}\ge0$, and concavity of $\varphi$ gives
\begin{align*}
&\varphi\bigl(m(S)+q_{c\sigma}(i)\bigr)-\varphi\bigl(m(S)\bigr)\\
&\qquad\ge\;\varphi\bigl(m(T)+q_{c\sigma}(i)\bigr)-\varphi\bigl(m(T)\bigr),
\end{align*}
which is the submodularity inequality for $F_{c\sigma}$; monotonicity follows since $\varphi$ is non-decreasing and $q_{c\sigma}(i)\ge0$. Non-negative linear combinations of monotone submodular functions remain monotone submodular. Normalization holds since $F(\emptyset)=\sum_{(c,\sigma)}w_{c\sigma}\min\{\tau_{c\sigma},0\}=0$ with $\tau_{c\sigma}>0$.

\subsection{$(1-1/e)$ coverage approximation}
\label{sec:app-proof-thm-approx}

\begin{theorem}[Round-robin is greedy on a partition matroid; $(1-1/e)$]
\label{thm:approx}
Let $\widetilde S_{\mathrm{RR}}$ be the idealized round-robin assignment of Supp.~\ref{sec:app-lift} and let $\widetilde S^\star\in\arg\max_{\widetilde S\in\widetilde{\mathcal{I}}}\widetilde F(\widetilde S)$. Then
\begin{equation*}
\widetilde F(\widetilde S_{\mathrm{RR}})\;\ge\;\bigl(1-1/e\bigr)\,\widetilde F(\widetilde S^\star),
\end{equation*}
and hence, by Lem.~\ref{lem:containment}, the pool subset $\hat D$ selected by the implemented stratified round-robin satisfies $F(\hat D)\ge(1-1/e)\,\widetilde F(\widetilde S^\star)$.
\end{theorem}

\begin{table*}[t]
\centering
\fontsize{8}{9.6}\selectfont
\setlength{\tabcolsep}{8pt}
\begin{tabular}{l|ccc}
\toprule
Method & Vision-Flan & LLaVA-1.5 & LLaVA-OVSI \\
\midrule
COINCIDE~\cite{coincide} & $K{=}5{,}000,\ \tau{=}0.1$ & $K{=}10{,}000,\ \tau{=}0.1$ & $K{=}20{,}000,\ \tau{=}0.1$ \\
ICONS~\cite{wu2024icons} & warm-up $5\%$, $p{=}0.2$ & warm-up $5\%$, $p{=}0.2$ & warm-up $5\%$, $p{=}0.2$ \\
PreSel~\cite{presel2025} & $k{=}5$ & $k{=}5$ & $k{=}5$ \\
PRISM~\cite{bi2025prism} & Layer 1 & Layer 1 & Layer 1 \\
\bottomrule
\end{tabular}
\caption{\textbf{Baseline configurations.} Settings reported in the original papers (Vision-Flan, LLaVA-1.5) and those used on LLaVA-OVSI; symbols follow each method's original notation.}
\label{tab:baseline-hparams}
\end{table*}

\noindent\emph{Proof.} $\widetilde F$ has the same saturated-coverage form as $F$, so the argument of Prop.~\ref{prop:submodular} (composition of a non-decreasing concave function with a modular sum, followed by a non-negative linear combination) applies verbatim on $2^{\widetilde V}$; hence $\widetilde F$ is monotone submodular. The key structural property we exploit is that $\widetilde F$ decomposes additively across the partition blocks, $\widetilde F(\widetilde S)=\sum_{(c,\sigma)}\widetilde F_{c\sigma}\bigl(\widetilde S\cap\widetilde V_{c\sigma}\bigr)$ with $\widetilde F_{c\sigma}$ the $(c,\sigma)$ summand of Eq.~\ref{eq:app-Ftilde}, each depending only on pairs inside its own block. The partition-matroid constraint $|\widetilde S\cap\widetilde V_{c\sigma}|\le k_{c\sigma}$ decomposes block-wise as well, so maximizing $\widetilde F$ over $\widetilde{\mathcal{I}}$ splits into $|\hat C|\!\cdot\!|\hat\Sigma|$ independent cardinality-constrained monotone submodular subproblems, one per block. Within each block, the idealized assignment selects pairs in descending order of $q_{c\sigma}$; since the marginal gain of pair $(i,c,\sigma)$ when its block has accumulated mass $x$ is $w_{c\sigma}\min\{q_{c\sigma}(i),\max\{\tau_{c\sigma}-x,0\}\}$, non-decreasing in $q_{c\sigma}(i)$ for every $\tau_{c\sigma}>0$ and $w_{c\sigma}>0$, this order is a valid trajectory of the discrete greedy algorithm for the cardinality-constrained subproblem (ties, including the vanishing gains once a block saturates, broken by a fixed rule; the cyclic order across cells is irrelevant for the idealized assignment since its blocks do not interact). The Nemhauser--Wolsey--Fisher $(1-1/e)$ guarantee for cardinality-constrained monotone submodular maximization~\cite{nemhauser1978analysis} therefore applies in each block, and summation preserves the bound globally. The consequence for $\hat D$ is immediate from Lem.~\ref{lem:containment}. Since the argument only uses that the marginal gain is non-decreasing in $q_{c\sigma}(i)$, the bound holds verbatim for every $\alpha\in(0,1]$ and $w_{c\sigma}>0$. The benchmark $\widetilde S^\star$ is the per-cell-budgeted optimum, not the global $b\%$-cardinality optimum of $F$; the per-cell budgets are the coverage mechanism whose fairness Thm.~\ref{thm:ef1} certifies.

\subsection{Envy-freeness up to one item}
\label{sec:app-proof-thm-ef1}

\begin{theorem}[Envy-free up to one item across cells]
\label{thm:ef1}
Treat each cell $(c,\sigma)$ as an agent with additive valuation $v_{c\sigma}(S)=\sum_{i\in S} q_{c\sigma}(i)$. Let $A_{c\sigma}\subseteq\hat D$ be the set of instances claimed by cell $(c,\sigma)$ during stratified round-robin. Then the allocation $\{A_{c\sigma}\}$ is envy-free up to one item: for every pair of cells $(c,\sigma)$ and $(c',\sigma')$,
\begin{equation*}
  v_{c\sigma}(A_{c\sigma})\;\ge\;v_{c\sigma}(A_{c'\sigma'})-\max_{i\in A_{c'\sigma'}} v_{c\sigma}(\{i\}).
\end{equation*}
\end{theorem}

\noindent\emph{Proof.} Each cell-agent's valuation $v_{c\sigma}(S)=\sum_{i\in S}q_{c\sigma}(i)$ is additive with $q_{c\sigma}(i)\ge 0$. The implemented selector is the classical round-robin picking protocol over a common pool: agents pick once per round in a fixed cyclic order, each claiming the remaining sample it values most (samples outside $G_{c\sigma}$ have value exactly zero to agent $(c,\sigma)$, and a claimed sample is removed from every queue), with equal budgets $k_{c\sigma}$ keeping the sequence balanced. For additive valuations this protocol yields an envy-free-up-to-one-item allocation (the EF1 notion is due to Lipton, Markakis, Mossel and Saberi~\cite{lipton2004approximately}; the round-robin guarantee is standard, see, e.g., Caragiannis et al.~\cite{caragiannis2016unreasonable}); an approximate EF1 analogue holds for the saturated valuation $\min\{\tau_{c\sigma},v_{c\sigma}(\cdot)\}$~\cite{amanatidis2023roundrobin}. For self-containedness, fix agents $a$ and $b$: if $a$ precedes $b$ in the cyclic order, $a$'s round-$t$ pick was made while $b$'s round-$t$ pick was still available, so $v_a(A_a)\ge v_a(A_b)$; if $b$ precedes $a$, dropping $b$'s first pick and pairing $a$'s round-$t$ pick with $b$'s round-$(t{+}1)$ pick (still available at $a$'s turn in round $t$) gives the stated inequality. An agent whose queue empties early values all remaining samples at zero, so later picks add no envy.

\section{Additional Experimental Details}
\label{sec:sup-exp-details}

\subsection{Evaluation Benchmarks}
\label{sec:sup-eval-bench}

Under the VLMEvalKit~\cite{duan2024vlmevalkit}, we comprehensively evaluate our method on 14 multi-modal benchmarks, including MME~\cite{fu2024mmecomprehensiveevaluationbenchmark}, MMBench$_{en-v1.1}$~\cite{liu2023mmbench}, MMStar~\cite{chen2024we}, MMMU~\cite{yue2023mmmu}, MMVet~\cite{yu2023mmvetevaluatinglargemultimodal}, BLINK~\cite{fu2024blink}, MMT-Bench~\cite{mmtbench}, AI2D~\cite{kembhavi2016ai2d}, ScienceQA~\cite{lu2022scienceqa}, MathVista$_{MINI}$~\cite{lu2023mathvista}. For the experiment in Supp.~\ref{sec:sup-scale} that scales up in the OCR capability, we additionally evaluate SSR on OCRBench~\cite{liu2023ocrbench}, ChartQA~\cite{masry2022chartqa}, DocVQA~\cite{mathew2021docvqa} and InfoVQA~\cite{mathew2022infographicvqa}.
Since our setup focuses on the single-image SFT phase, the model does not possess the multi-image understanding ability. Thus, for MMMU and BLINK, we report results on the single-image QA split.

\subsection{Baselines}
\label{sec:sup-baselines}

We compare SSR with 10 methods, including
a) \textit{random sampling}: the strong diversity-prioritized baseline, evaluated based on the average results from three trials of different random splits; b) \textit{perplexity}, including its two variants before (PPL-mid) and after (PPL-si) the single-image SFT on the entire data pool; c) \textit{Deita}~\cite{liu2024Deita}, the score- and embedding-based selection method for LLMs, run in its score-only top-$k$ form (Deita\textsubscript{top-$k$}): Deita controls diversity by thresholding pairwise embedding similarity, whose $O(N^2)$ complexity and dataset-specific threshold do not scale to our 2.6M instances, and any threshold we chose on its behalf would be equally contestable, so we rank by its quality and complexity scores and label the variant explicitly; d) \textit{CLIP similarity}~\cite{radford2021clip} (ViT-L) that evaluates the image-text alignment; e) \textit{E5-V similarity}~\cite{jiang2024E5VUniversalEmbeddings}, the SOTA MLLM-based universal embedding model built on LLaMA-3-8B~\cite{dubey2024Llama3Herd}; f) COINCIDE~\cite{coincide}, ICONS~\cite{wu2024icons}, and PreSel~\cite{presel2025}, the SOTA selection strategies for MLLMs; and g) \textit{PRISM}~\cite{bi2025prism}, the most recent selector, which scores each sample by its correlation with the corpus-mean-centered intrinsic features of the target MLLM and prunes by a global threshold.

All baseline configurations follow settings endorsed by their original authors; Tab.~\ref{tab:baseline-hparams} lists the complete choices. For COINCIDE, the two hyperparameters are the number of clusters $K$ and the temperature $\tau$ of its cluster-level sampling distribution, with the official $\tau{=}0.1$. For ICONS, the warm-up ratio fixes the fraction of the pool used to train the warm-up model underlying its influence scores, and $p$ sets the proportion of top samples per task: a sample passes the consensus vote for a task when its influence reaches the task-specific threshold $\tau_k$, the $((1{-}p)\!\times\!100)$-th percentile of that task's influence distribution. We adopt its best-performing $5\%$ warm-up ratio (small warm-up sets consistently outperform the full pool in their ablation) and the main-experiment $p{=}0.2$, \ie 80th-percentile thresholds. For PreSel, $k$ is the number of nearest neighbors in feature space for computing the NC score, reported robust across $k\in\{5,10,20\}$ and set to $5$ in all its experiments. For PRISM, intrinsic visual features are extracted at the shallow Layer~1, which its authors find most effective for redundancy detection. The competitors were designed for thousand-scale pools; since the settings above are either scale-free or reported robust, we retained each method's official configuration. COINCIDE alone provides scale-adaptation guidance ($K$ grows from 5{,}000 on the 186K Vision-Flan pool to 10{,}000 on the 665K LLaVA pool, a $\approx$3.5$\times$ increase), which we extrapolated to $K{=}20{,}000$ on the 2.6M pool. We emphasize that the competitors' difficulty at million scale stems from their model-centric design and narrow criteria (Sec.~\ref{sec:exp-main}) rather than from hyperparameter sensitivity: SSR itself exposes no tunable selection hyperparameter (Sec.~\ref{sec:method-guarantees}), so no tuning asymmetry favors our method.

\subsection{Cambrian-1 Transfer Protocol}
\label{sec:sup-cambrian}

For the third-architecture transfer in Sec.~\ref{sec:exp-transfer} (Tab.~\ref{tab:cambrian}), we take the scorer and styler distilled on 15\% of LLaVA-OVSI (Sec.~\ref{sec:exp-main}) and apply them \emph{zero-shot} to profile all 7M samples of the Cambrian-7M pool~\cite{tong2024cambrian}, with no re-annotation, no fine-tuning, and no change to the taxonomy. The stratified round-robin selector then draws 5\%, 10\% and 30\% subsets, on which Cambrian-1-8B is fine-tuned from its official pre-trained checkpoint with the official recipe\footnote{\urlstyle{same}\url{https://github.com/cambrian-mllm/cambrian/blob/main/scripts/cambrian/finetune_cambrian_8b.sh}}; random sampling and the full-data reference use the identical recipe. Evaluation follows Supp.~\ref{sec:sup-eval-bench}.

{\renewcommand{\arraystretch}{1.2} %
\begin{table*}[hb]
\small
\centering
\begin{tabularx}{\textwidth}{>{\centering\arraybackslash}m{3.6cm}|X|>{\centering\arraybackslash}m{1.3cm}}
\toprule
{multi-modal Capabilities} & {Definitions} & {Examples} \\

\midrule
activity recognition            & actions or behaviors of humans, animals, or objects & Fig.~\ref{fig:supp-examples_activity}\\
causal reasoning                & cause-and-effect relationships between events or variables to predict outcomes and explain phenomena  & Fig.~\ref{fig:supp-examples_causal}\\
humanities              & history, literature, philosophy, art, and culture to understand human experiences and societal developments  & Fig.~\ref{fig:supp-examples_humanities}\\
STEM knowledge          & science, technology, engineering, and mathematics, chemistry, economics etc & Fig.~\ref{fig:supp-examples_stem} \\
comparative analysis            & compare multiple entities, concepts, or datasets to identify similarities, differences, and relationships  & Fig.~\ref{fig:supp-examples_comparative}\\

data understanding              & documents, tables, charts, graphics, infographics & Fig.~\ref{fig:supp-examples_data}\\
{object spatial understanding}            & the positions, orientations, countings and relationships of objects & Fig.~\ref{fig:supp-examples_spatial}\\
attribute identification                & various characteristics and properties of objects, such as identity, color, size, shape, material, emotion, and other distinguishing features & Fig.~\ref{fig:supp-examples_attribute}\\
logical deduction               & to analyze information, recognize patterns, draw valid conclusions based on structured principles of logic and make reasoned decisions & Fig.~\ref{fig:supp-examples_logical}\\
scene understanding             & complex environment with objects, their attributes, spatial relationships, and activities, as well as surrounding information and circumstances within the scene & Fig.~\ref{fig:supp-examples_scene}\\
fine-grained recognition                & subtle differences and specific features within similar categories of objects & Fig.~\ref{fig:supp-examples_fine-grained}\\
language generation             & generate coherent and contextually appropriate text in various languages, styles, and formats based on instructions & Fig.~\ref{fig:supp-examples_lang}\\
in-context learning             & follow the demonstrations of the task within a given conversation & Fig.~\ref{fig:supp-examples_in-context}\\
optical character recognition              & the conversion between images of printed/handwritten text and machine-readable text & Fig.~\ref{fig:supp-examples_ocr}\\
\midrule
style & multi-choice, coordinate, yes/no, word/short-phrase, short description, detailed description, comparison, chain-of-thought (step-by-step), specified style & Fig.~\ref{fig:supp-examples_activity} - \ref{fig:supp-examples_ocr}\\
\bottomrule
\end{tabularx}
\caption{Criteria we recognize as the foundational pillars for developing vision perception and reasoning capabilities within MLLMs, and the interaction styles we identify from instructional multi-modal data.}
\label{tab:caps}
\end{table*}
}

\begin{tcolorbox}[float*,floatplacement=!tp,width=\textwidth,colback=black!5!white,colframe=black!75!black,fontupper=\scriptsize\ttfamily,fontlower=\scriptsize\ttfamily,title=Prompt to Derive the Capabilities of Interest]
You are an AI expert tasked with defining the essential capabilities for a next-generation multi-modal large language model. To achieve this, we will follow a structured, three-step process: Discovery, Expansion, and Refinement.\\ \\

\textbf{Stage 1: Discovery}

    First, I want you to act as a research analyst. Your task is to analyze the following data sources to identify a broad set of foundational capabilities and iterative tasks suggested by the data.\\
    
    Data Sources: [\par
    \hspace{2em}COCO Caption, \par
    \hspace{2em}Vision FLAN, \par
    \hspace{2em}...\\
    ]\footnote{\scriptsize\ttfamily See Tab.~\ref{tab:dist_source} for the full list.}\\

    Based on your analysis of these sources, generate a list of potential capabilities.\footnote{\scriptsize\ttfamily The initial list generated by the LLM in this round will be denoted as  \{initial capabilities\}.}\\ \\
 
\textbf{Step 2: Expansion}

    Now, I want you to think beyond this initial list. I will provide you with a list of initial capability candidates and random samples of visual question answer pairs. Your task is to think creatively and identify any new capabilities that are not adequately covered by the existing list but are necessary for answering these visual questions.\\

    Initial List:\\
    \{initial capabilities\}\\

    Random Samples:\\
    \{Random samples drawn from the datapool\}\\

    Your goal is to heuristically expand our list. Think creatively and do not worry about overlap at this stage; focus on generating a comprehensive set of potential new capabilities.\footnote{\scriptsize\ttfamily This step can be iterated multiple times to get a sufficient candidate list. The expanded list generated in this round will be denoted as \{expanded capabilities\}.}\\ \\

\textbf{Step 3: Refinement}

    Next, I want you to refine our expanded list of capabilities. I will provide you with a list of capabilities. Your task is to identify and merge criteria that are semantically similar or redundant. For each proposed merger, provide a brief justification. The goal is to create a comprehensive yet more concise and semantically coherent list.\\

    Expanded List:\\
    \{expanded capabilities\}\\

    Present the new, merged list.\footnote{\scriptsize\ttfamily The merged list suggested in this round will be denoted as  \{merged capabilities\}.}\\ \\

    Next, I want you to analyze the frequency and importance of each capability on the merged list. Your task is to analyze the merged list of capabilities below and identify capabilities that are the most critical for a wide variety of multi-modal questions. Also, identify the potentially long-tailed datapoints, which might be capabilities that are too niche, rarely required, or could be considered a sub-component of other capability we've already defined. \\

    Merged List:\\
    \{merged capabilities\}\\

    Your goal is to create a final, core set of capabilities for training while discarding less representative ones. Please provide the final list of essential capabilities and a separate list you chose to set aside.\footnote{\scriptsize\ttfamily This step can be iterated multiple times to get a refined list.}\\

\end{tcolorbox}

\subsection{Profiling Details}
\label{sec:sup-expset}

This subsection expands the profiling pipeline of Sec.~\ref{sec:method-profile}. Supp.~\ref{sec:app-cap-style} documents how the capability and interaction-style taxonomies were derived; Supp.~\ref{sec:app-expset-prompt} details the judge prompt that yields the rich per-sample profile consumed by selection.

\subsubsection{Capability and Style Taxonomies}
\label{sec:app-cap-style}

Tab.~\ref{tab:caps} lists the $|C|$ capabilities and $|\Sigma|$ interaction styles we identify from the current open-access multi-modal instruction-following datasets for building general-purpose large models. Our taxonomy is constructed through a three-stage, transferable procedure: (i) \emph{discovery} of an initial set of capabilities from analyzing the LLaVA-OVSI data sources; (ii) \emph{expansion} of this set by prompting GPT-4o with random samples to surface novel patterns; and (iii) two iterations of \emph{refinement} that merge semantically similar items and prune long-tail capabilities. The prompt used in this capability-definition process is shown immediately below in the boxed insert. A parallel methodology yields the taxonomy of interaction styles.

We emphasize that the efficacy of SSR is not rigidly tied to this specific capability composition. As demonstrated in Sec.~\ref{sec:exp-richness}, the success of SSR is predicated on the richness of the capability subspace, where even randomly sampled subsets of capabilities achieve superior performance. Furthermore, our analysis in Supp.~\ref{sec:sup-scale} and Fig.~\ref{fig:line-scale} shows that the framework seamlessly and effectively scales to incorporate new capabilities. This inherent extensibility ensures that as the data landscape evolves, either through new public datasets or private, specialized sources, our pipeline can be readily adapted to curate data for a broader and more sophisticated range of knowledge. This includes advanced skills such as creative image-based writing, complex chain-of-thought reasoning, and solving competition-level mathematical problems.

\subsubsection{Querying Rich Profiles}
\label{sec:app-expset-prompt}

\myparagraph{Prompt template.} Given that obtaining large-scale human annotations for both scores and styles is often impractical, prompting multi-modal large language models (MLLMs) serves as a practical and effective alternative. The boxed insert below shows our prompt template for querying rich scores and styles for each sample, where \textcolor{fgreen}{\{Input\}} and \textcolor{fgreen}{\{Response\}} are questions and answers. Multi-round user-assistant interactions are paired and concatenated for demonstration.

\begin{tcolorbox}[float*=!tp, width=\textwidth, colback=black!5!white, colframe=black!75!black, fontupper=\ttfamily, fontupper=\scriptsize\ttfamily, fontlower=\scriptsize\ttfamily,title=Prompt to Query GPT-4o for Rich Scores and Styles]

\textbf{System Prompt}:\\
    You are an AI expert rater designed to analyze the Visual Question Answering (VQA) instance in the user query to perform the following tasks step-by-step:
        
    Step 1: Classify the VQA instance into given conversation style.
    
    Step 2: Evaluate the helpfulness of the information provided in the VQA instance with respect to various model capabilities. Specifically, rate how well this information could enhance each capability of a multi-modal large language model through learning from it.
    
    Step 3: Output the results strictly follow the JSON format.\\

\textbf{User Prompt}:\\
\#\# Instruction \\
    You need to perform the following three steps to rate the User Query and output result in the dictionary format. 
    
    Step 1: Classify the instance in interaction style. 
        Determine the task style of the VQA instance and select styles from the list ``task\_styles" below. Sort the selected styles by frequency of occurrence.
        
    Step 2: Rate each capability from 0-5.
        For each capability listed and explained in ``task\_capabilities" below, analyze how effectively the VQA instance could enhance that capability of a Multimodal Large Language Model (MLLM) by learning from it. Rate each capability using the scores from the ``score\_scale" list below in refernce to the guidelines. Please ensure that the scores are well-distributed across the range.
        
    Finally, output the results strictly following the dictionary format defined in Output Format. Do not output any additional tokens outside it. \\
    
    \#\# User Query\\
    Question: \textcolor{fgreen}{\{Input\}} \\ 
    Answer: \textcolor{fgreen}{\{Response\}} \\

    \#\# Task Styling\\
    task\_styles = [\par
        \hspace{2em}multi-choice,\par
        \hspace{2em}coordinate,\par
        \hspace{2em}... \hspace{1em}\par
    ]\footnote{\scriptsize\ttfamily See Tab.~\ref{tab:caps} for the full list.}\\

    \#\# Task Capabilities\par
    task\_capabilities = [ \par
        \hspace{2em}optical character recognition, \hspace{1em}\# the conversion between images of printed/handwritten text and machine-readable text\par
        \hspace{2em}... \hspace{1em}\par
    ]\footnote{\scriptsize\ttfamily See Tab.~\ref{tab:caps} for the full list and explanation.}\\

    \#\# Rating Scale\\
    score\_scale = [\par
        \hspace{2em}0,  \hspace{1em}\# Not Relevant: The VQA instance does not present or relate to the capability in any meaningful way.\par
        \hspace{2em}1,  \hspace{1em}\# Minimal: The VQA instance offers very little information relevant to the capability, providing negligible value for enhancement.\par
        \hspace{2em}2,  \hspace{1em}\# Fair: The VQA instance contains some relevant information but lacks depth and clarity, contributing minimally to the model's learning in this capability.\par
        \hspace{2em}3,  \hspace{1em}\# Good: The VQA instance provides a fair amount of relevant information, which can moderately aid in the model's learning and enhancement of the capability.\par
        \hspace{2em}4,  \hspace{1em}\# Significant: The VQA instance offers substantial information that is highly relevant and beneficial, significantly aiding the model's learning and enhancement of the capability.\par
        \hspace{2em}5,  \hspace{1em}\# Excellent: The VQA instance is exceptionally rich in relevant information, providing comprehensive and clear insights that would greatly enhance the model's learning and mastery of the capability.\\
    ]\\

    \#\# Output Format
    \begin{verbatim}
    {
        "style": "<list of string>",
        "capability2score": "<dict of str:int>", 
        "capability2explanation": "<dict of str:str>", 
    }
    \end{verbatim}

\end{tcolorbox}

\myparagraph{Annotation protocol.} To enhance the stability of pointwise scores rated by the model, we define a scale from 0 to 5 and establish clear benchmarks. During the evaluation process, each data sample is assessed against all predefined capability dimensions, and all interaction styles present in the textual component are recalled. To enhance the reliability and self-consistency of these annotations, we instruct the model to provide a justification for each score it assigns.
A critical aspect of our methodology, particularly for multi-modal data, is ensuring a holistic evaluation. Our prompts specifically direct the model to assess the synergy and alignment between the image and text in the context of the task. This is crucial for preventing uni-modal bias, where the model's judgment might be disproportionately influenced by either the visual or the linguistic content alone.

\begin{figure*}[ht]
\begin{center}
\centerline{\includegraphics[width=0.93\textwidth]{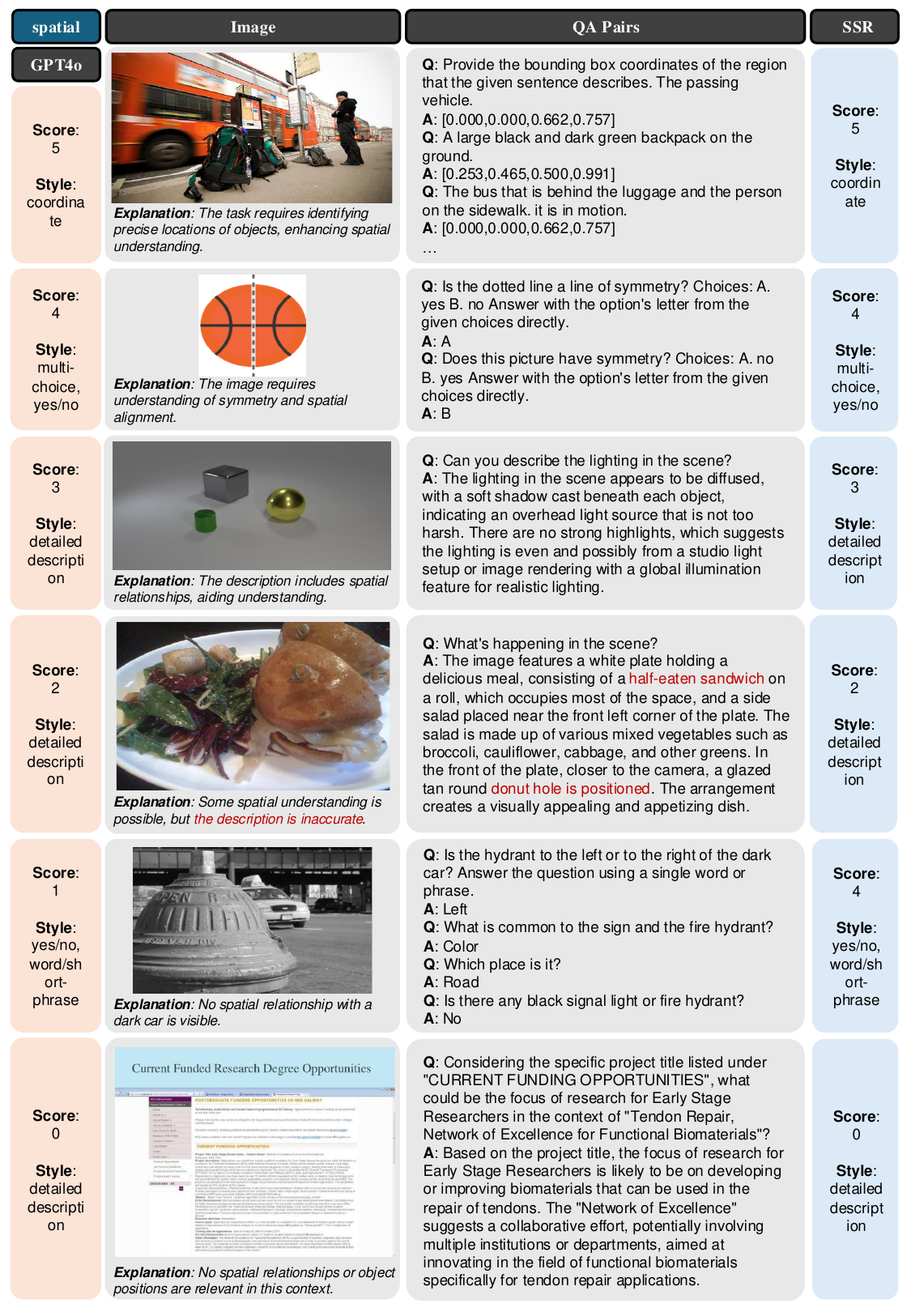}}
\caption{Randomly selected samples from the LLaVA-OVSI data pool within each score group for the \textbf{object spatial understanding} capability, ranging from 5 to 0, as evaluated by GPT-4o, along with corresponding score explanations and SSR predictions.}
\label{fig:supp-examples_spatial}
\end{center}
\end{figure*}

\begin{figure*}[ht]
\begin{center}
\centerline{\includegraphics[width=0.93\textwidth]{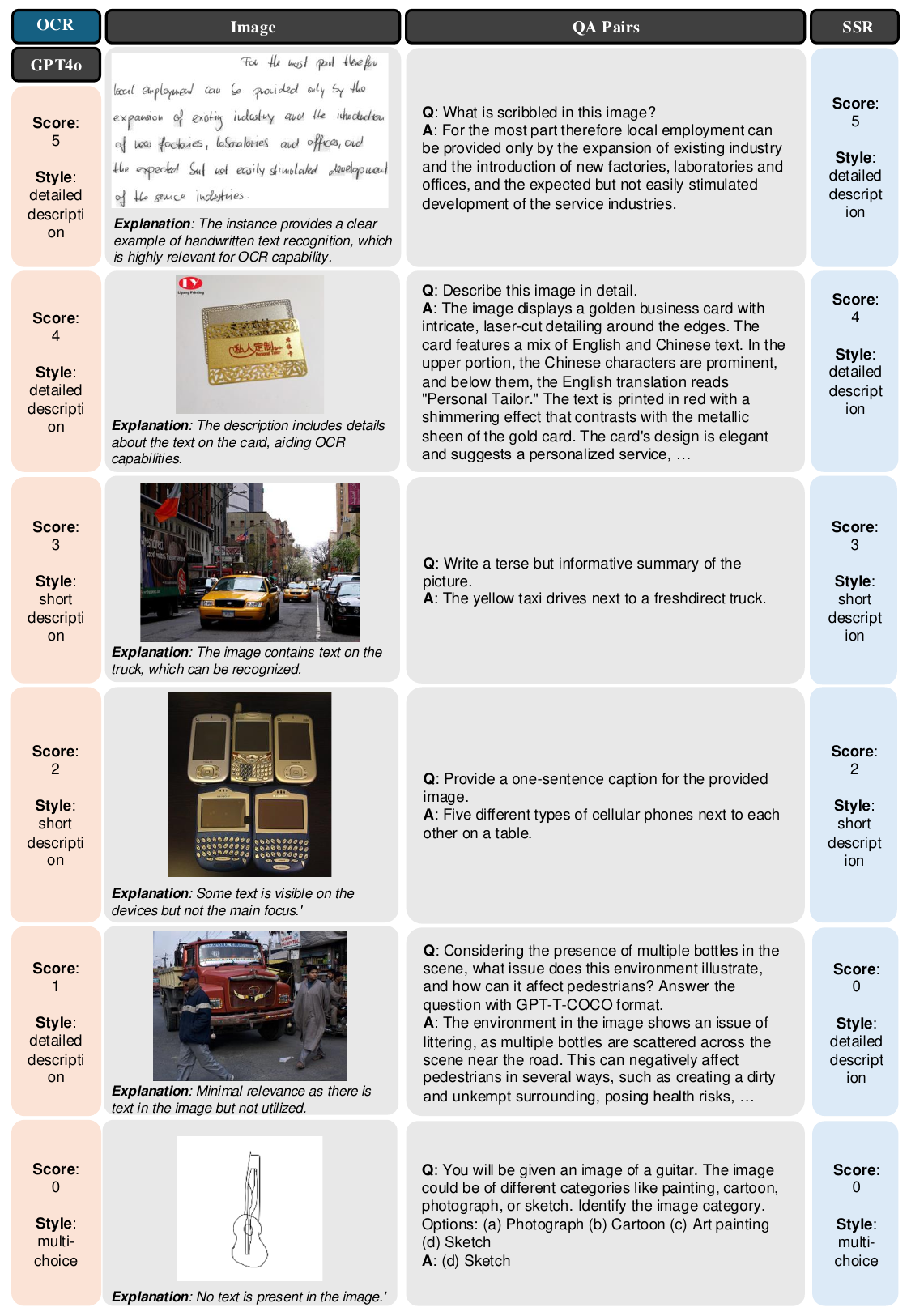}}
\caption{Randomly selected samples from the LLaVA-OVSI data pool within each score group for the \textbf{optical character recognition} capability, ranging from 5 to 0, as evaluated by GPT-4o, along with corresponding score explanations and SSR predictions.}
\label{fig:supp-examples_ocr}
\end{center}
\end{figure*}

\myparagraph{Annotator choice.} As established in Sec.~\ref{sec:exp-application} of our main paper, the SSR framework is robust to various annotation sources, demonstrating strong performance with both open-weight models and the proprietary one.
To rigorously evaluate the efficacy of our prompting strategy and the fidelity of the annotations it produces, we focus this analysis on our primary configuration, which leverages a leading proprietary model, \ie GPT-4o in our case, as the expert annotator. This choice achieves SOTA performance, while the strategy of annotating only a small data subset makes it highly cost-effective.
For qualitative illustration, we provide examples for each capability in Fig.~\ref{fig:supp-examples_activity}--\ref{fig:supp-examples_ocr}. These examples are organized by capability, showcasing the full 0--5 scoring scale, and are accompanied by detailed rationales for each assigned score.

\myparagraph{Case studies.} Our scoring mechanism demonstrates a strong sensitivity to task context. Consider, for instance, the 3rd and 5th samples in Fig.~\ref{fig:supp-examples_ocr}. Although the visual scenes are similar, the textual queries focus on distinct elements. When the task is to generate an ``informative summary" requiring text from a vehicle, the sample's contribution to the OCR capability is crucial, meriting a score of 3. Conversely, when the task shifts to global scene understanding where road details are paramount, the text becomes irrelevant, and the sample rightly receives an OCR score of 0.
Furthermore, our process effectively penalizes samples containing flawed or hallucinatory responses in the original dataset. For example, the hallucination present in the 4th example of Fig.~\ref{fig:supp-examples_spatial} is identified and results in a lower score, thereby preventing the propagation of erroneous information into the supervised fine-tuning (SFT) phase.
While model-generated annotations can occasionally include hallucinations (\eg 5th example in Fig.~\ref{fig:supp-examples_spatial}, 5th example in Fig.~\ref{fig:supp-examples_ocr}), we find that our finetuning process is remarkably robust to this low level of label noise, as validated by our accuracy analysis and empirical selection results.

\begin{figure*}[ht]
\centering
    \subfigure[SSR scorers trained with 15\% data]{\label{fig:sup-score-mae-15}\includegraphics[width=0.8\textwidth]{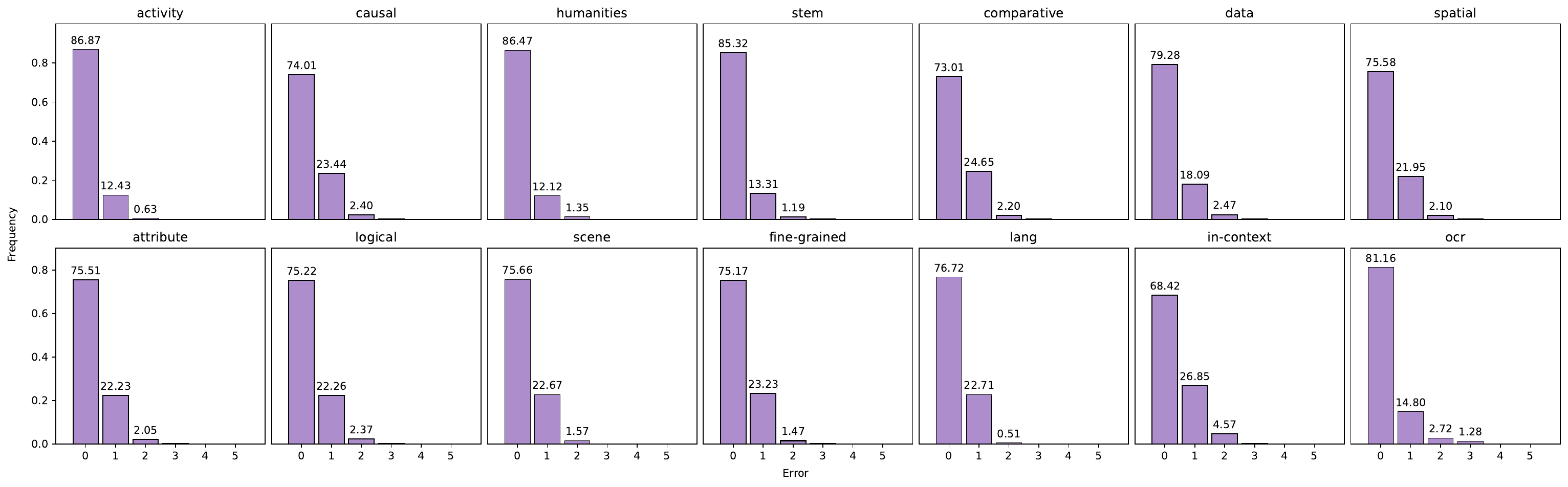}}\\
    \subfigure[SSR scorers trained with 30\% data]{\label{fig:sup-score-mae-30}\includegraphics[width=0.8\textwidth]{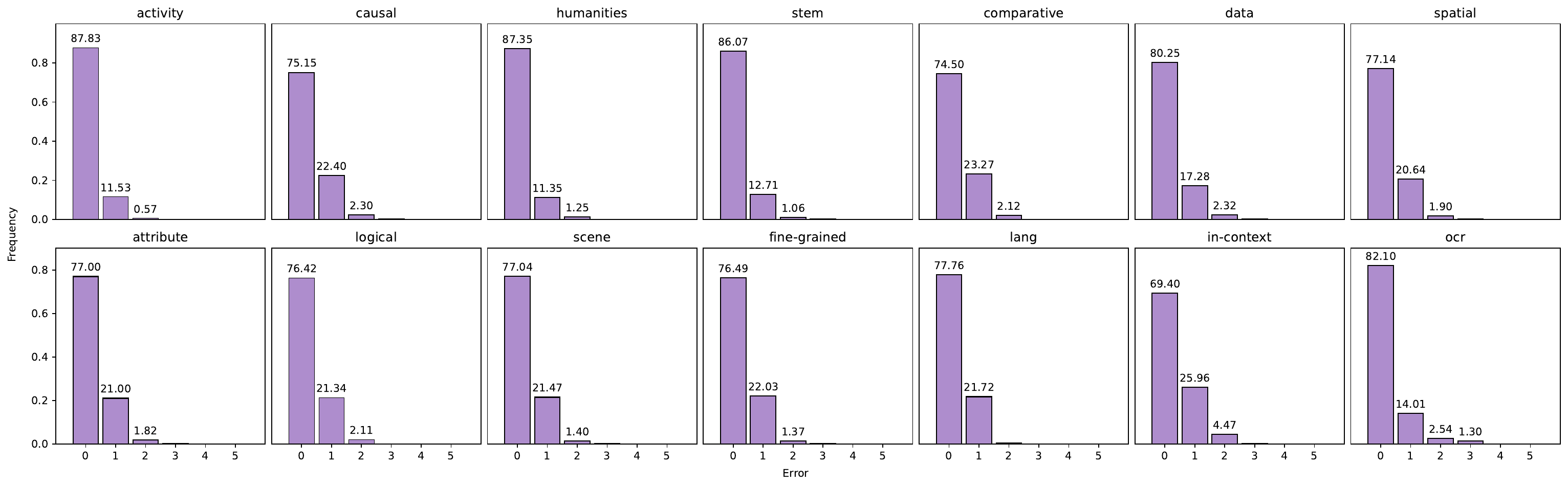}}
\caption{The mean absolute error of SSR scorer predictions against GPT-4o judgment over capabilities.}
\label{fig:sup-score-mae}
\end{figure*}
\begin{figure}[ht]
\centering
    \subfigure[Precisions of SSR styler]{\label{fig:sup-style-acc}\includegraphics[width=0.45\columnwidth]{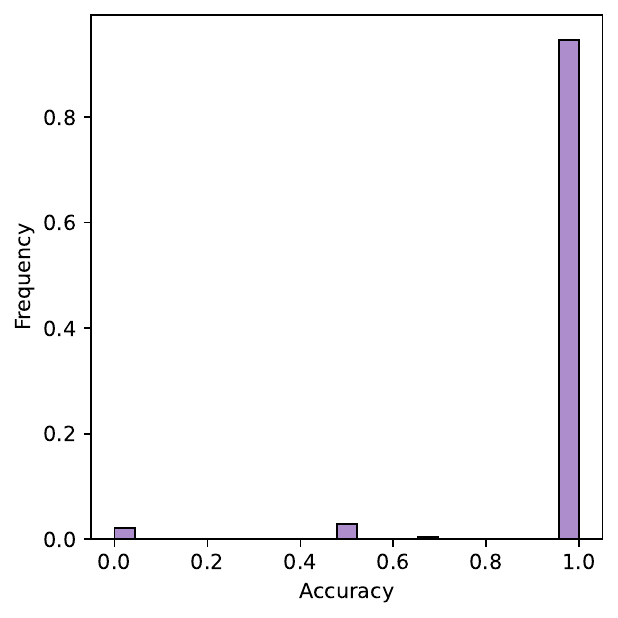}}
    \subfigure[Recalls of SSR styler]{\label{fig:sup-style-recall}\includegraphics[width=0.45\columnwidth]{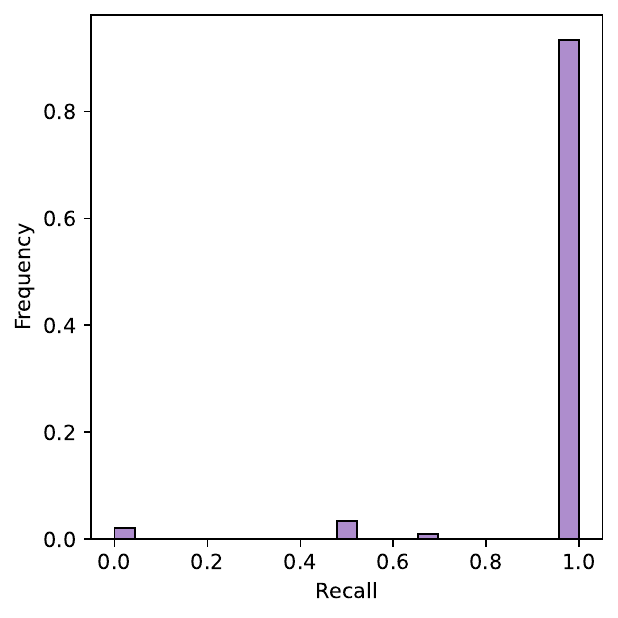}}
    \caption{The precision and recall of SSR styler predictions against GPT-4o judgment.}
\label{fig:sup-style-pr}
\end{figure}

Collectively, these cases validate the effectiveness of our task-aware prompting, demonstrating that our valuation criteria successfully balance the intricate interplay between image, query, and task context. Finally, this initial annotation phase, leveraging a small, high-quality set of judgments from GPT-4o, provides the foundation for fine-tuning the specialized SSR models, which are analyzed in detail in Supp.~\ref{sec:sup-score}.

\myparagraph{API cost of the 15\% GPT-4o annotation phase.} For practitioners reproducing the distilled pathway, we report the token accounting of the one-time GPT-4o annotation on 15\% of LLaVA-OVSI. The text portion of the prompts totals 428.33M input tokens. Images were submitted with \texttt{detail="auto"}; since the server-side routing rule is not disclosed, we bound the image cost between all-low-detail (33.76M tokens) and all-high-detail under the official tile formula (427.40M tokens), giving 462.09M--855.73M total input tokens. The structured JSON outputs total 218.43M tokens. At GPT-4o list prices (\$2.50/M input, \$10/M output), the standard-API cost is $\approx$\$3.34K--\$4.32K, and roughly half of that ($\approx$\$1.67K--\$2.16K) with the Batch API. This is a one-time expense: the distilled models profile the remaining 85\% locally, adapt zero-shot to new pools with no further API calls (Sec.~\ref{sec:exp-transfer}, Supp.~\ref{sec:sup-cambrian}), and can be bypassed altogether by prompting an open-weight judge (Tab.~\ref{tab:analysis}).

\section{Fidelity of Distilled SSR Models}
\label{sec:sup-score}

The distilled judge pathway introduced in Sec.~\ref{sec:method-sources} amortizes proprietary-API cost by fine-tuning small student MLLMs on a $15\%$ judge-labeled subset. This section validates the fidelity of those students against the GPT-4o judge on the remaining $85\%$, both quantitatively (Supp.~\ref{sec:sup-score-quant}) and qualitatively (Supp.~\ref{sec:sup-score-vis}).

\begin{figure*}[ht]
\begin{center}
\centerline{\includegraphics[width=1\textwidth]{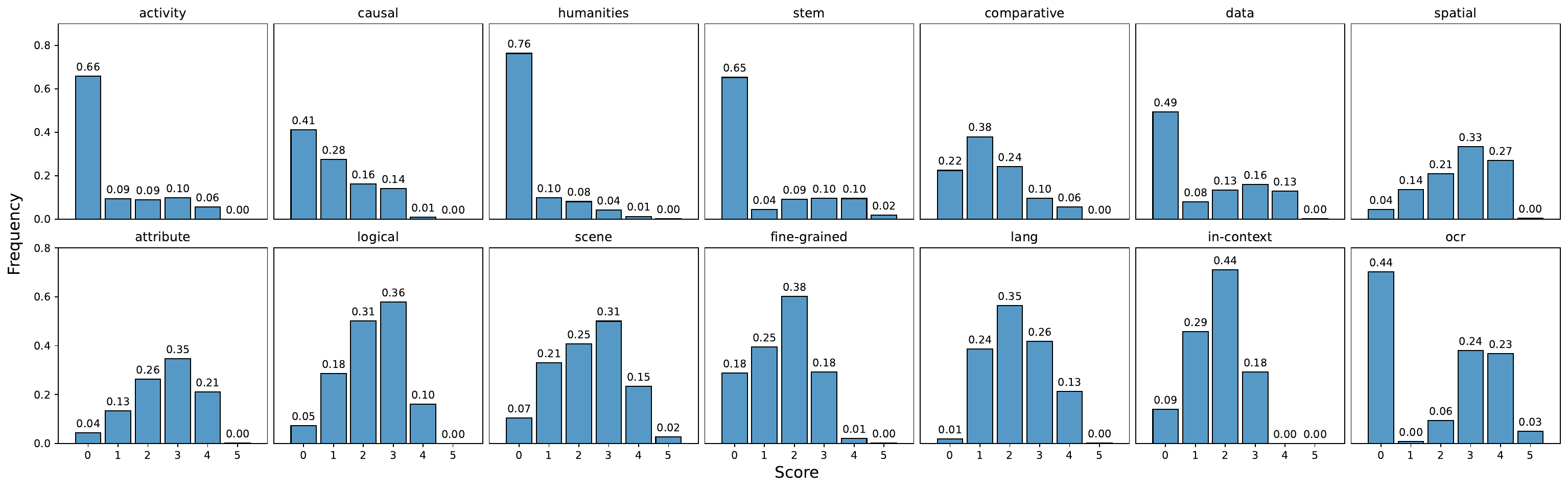}}
\caption{Score distribution of capabilities across the LLaVA-OVSI dataset inferred by SSR.}
\label{fig:supp-score-dist}
\end{center}
\end{figure*}

\subsection{Quantitative Validation}
\label{sec:sup-score-quant}
In our primary configuration, the fine-tuning of SSR models uses 15\% of the LLaVA-OVSI data, following the original instructional tuning strategy~\cite{li2024LLaVAOneVisionEasyVisual}\footnote{\urlstyle{same}\url{https://github.com/LLaVA-VL/LLaVA-NeXT/blob/main/scripts/train/finetune_si.sh}}.
To minimize the exploration cost of SSR in practical applications, no hyperparameter search was conducted in fine-tuning.
To validate the fidelity of our specialized model, we established a validation set by annotating the remaining 85\% of the LLaVA-OVSI data pool using GPT-4o.

\begin{figure*}[t]
    \begin{center}
    \centerline{\includegraphics[width=0.9\linewidth]{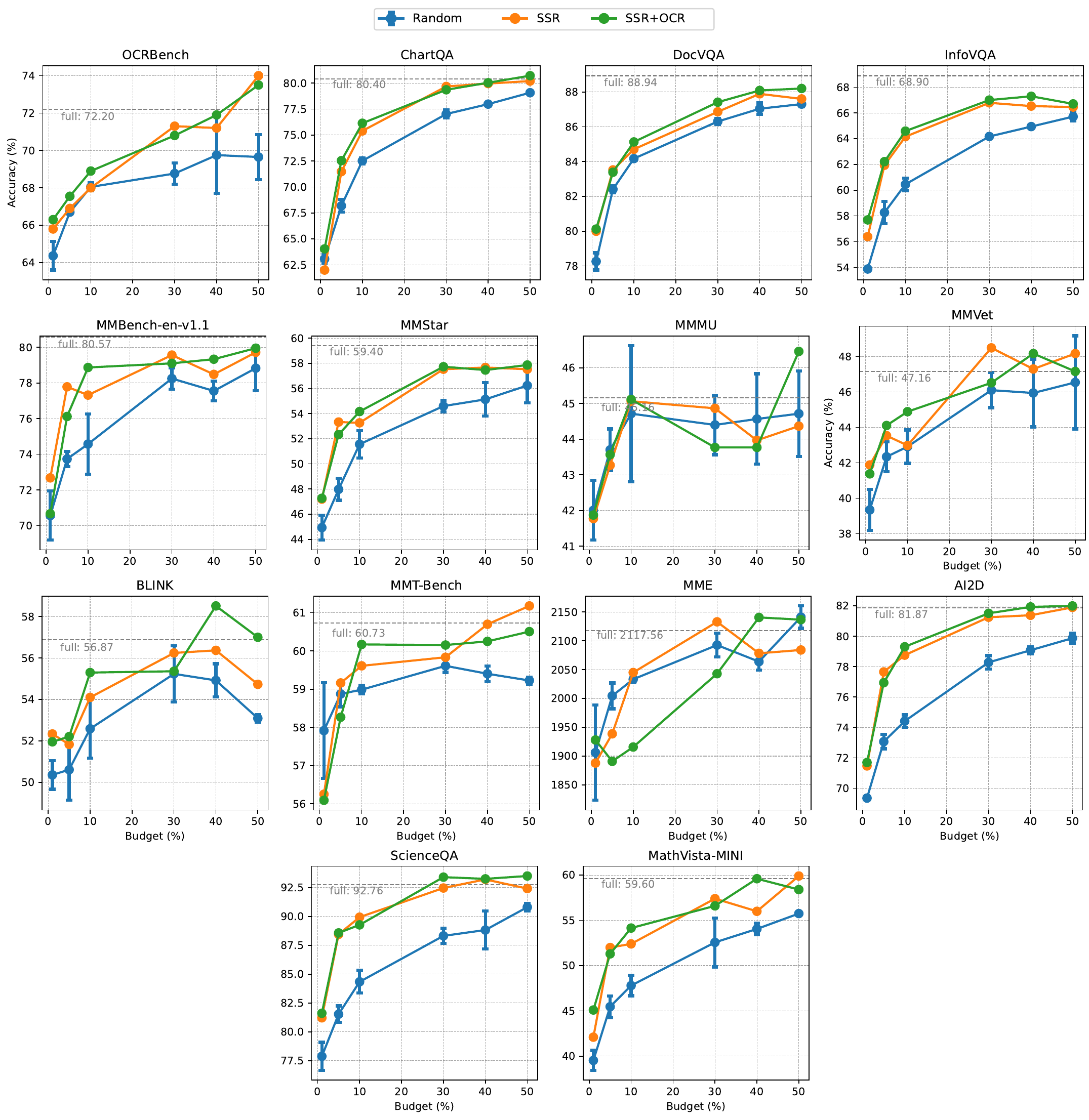}}
    \caption{\textbf{Scalability along both quantity and capability axes.} SSR consistently surpasses random across six data quantities (1\% $\rightarrow$ 5\% $\rightarrow$ 10\% $\rightarrow$ 30\% $\rightarrow$ 40\% $\rightarrow$ 50\%); adding OCR as a held-out capability (13 $\rightarrow$ 13 + OCR) extends the criteria set without degrading general performance, evidencing modular capability scalability. Legend: Random, SSR without OCR, SSR with OCR; 14 benchmarks.}
    \label{fig:line-scale}
    \end{center}
\end{figure*}

\myparagraph{Scorer Fidelity.}
The scorer's performance, quantified by Mean Absolute Error (MAE), is detailed in Fig.~\ref{fig:sup-score-mae-15}.
Our model demonstrates high alignment with the judge annotator. Averaged across all capabilities considered in the paper, spanning varying levels of granularity and difficulty, our scorer achieves an exact match with GPT-4o on 77.7\% of predictions. The agreement rate rises to an impressive 97.8\% when allowing for a tolerance of $\pm1$. This demonstrates that while pointwise judgments from the judge and our fine-tuned model exhibit inherent stochasticity~\cite{wettig2024QuRatingSelectingHighQuality}, much like human evaluation, SSR rarely makes unreasonable decisions. Consequently, it preserves the overall quality stratification essential for effective data selection.

\myparagraph{Styler Accuracy.}
Fig.~\ref{fig:sup-style-pr} reports the precision and recall distributions of the styler's predictions benchmarked against GPT-4o annotations. Compared to fine-grained capability scoring, the classification of interaction styles proves to be a more distinct and less ambiguous task. Consequently, the model achieves exceptional performance, registering an average precision of 96.35\% and a recall of 95.80\% across the entire data pool.

\myparagraph{Annotation Volume Analysis.}
We further explored the relationship between judge annotation volume and specialized model fine-tuning performance by scaling the fine-tuning set to 30\% of the data pool. The resulting MAE comparison in Fig.~\ref{fig:sup-score-mae-30} reveals that performance improvements over the 15\% baseline are negligible. This validates the data efficiency of our approach, demonstrating that the scorer achieves effective task alignment with an affordable API cost.

\myparagraph{Score Distribution Analysis.}
We visualize the capability-wise score distributions in Fig.~\ref{fig:supp-score-dist}.
For rare and specialized domains such as humanities and STEM, the model exhibits high precision, particularly in recognizing the presence or absence of a capability (\ie correctly assigning a score of 0). This sensitivity is crucial for our diversity-oriented round-robin sampling, ensuring that scarce, high-value data points are reliably preserved and not overlooked.

Collectively, these quantitative analyses confirm that our specialized SSR models deliver highly reliable judgments. This fidelity underpins the robust and superior performance of SSR observed in both in-domain inference and direct transfer to new data pools.

\subsection{Qualitative Analysis}
\label{sec:sup-score-vis}
\begin{figure*}[ht]
\begin{center}
\centerline{\includegraphics[width=0.93\textwidth]{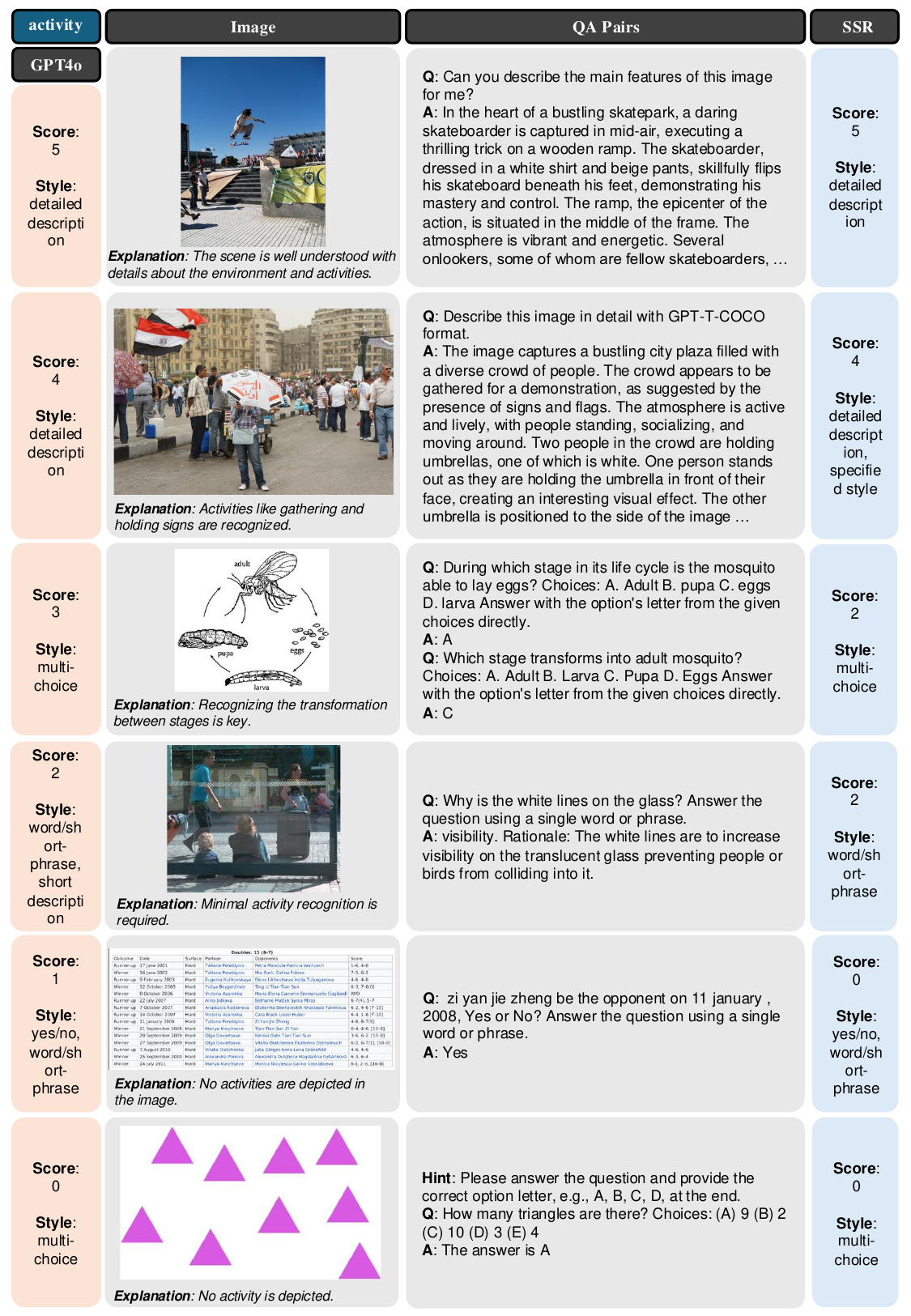}}
\caption{Randomly selected samples from the LLaVA-OVSI data pool within each score group for the \textbf{activity recognition} capability, ranging from 5 to 0, as evaluated by GPT-4o, along with corresponding score explanations and SSR predictions.}
\label{fig:supp-examples_activity}
\end{center}
\end{figure*}

\begin{figure*}[ht]
\begin{center}
\centerline{\includegraphics[width=0.93\textwidth]{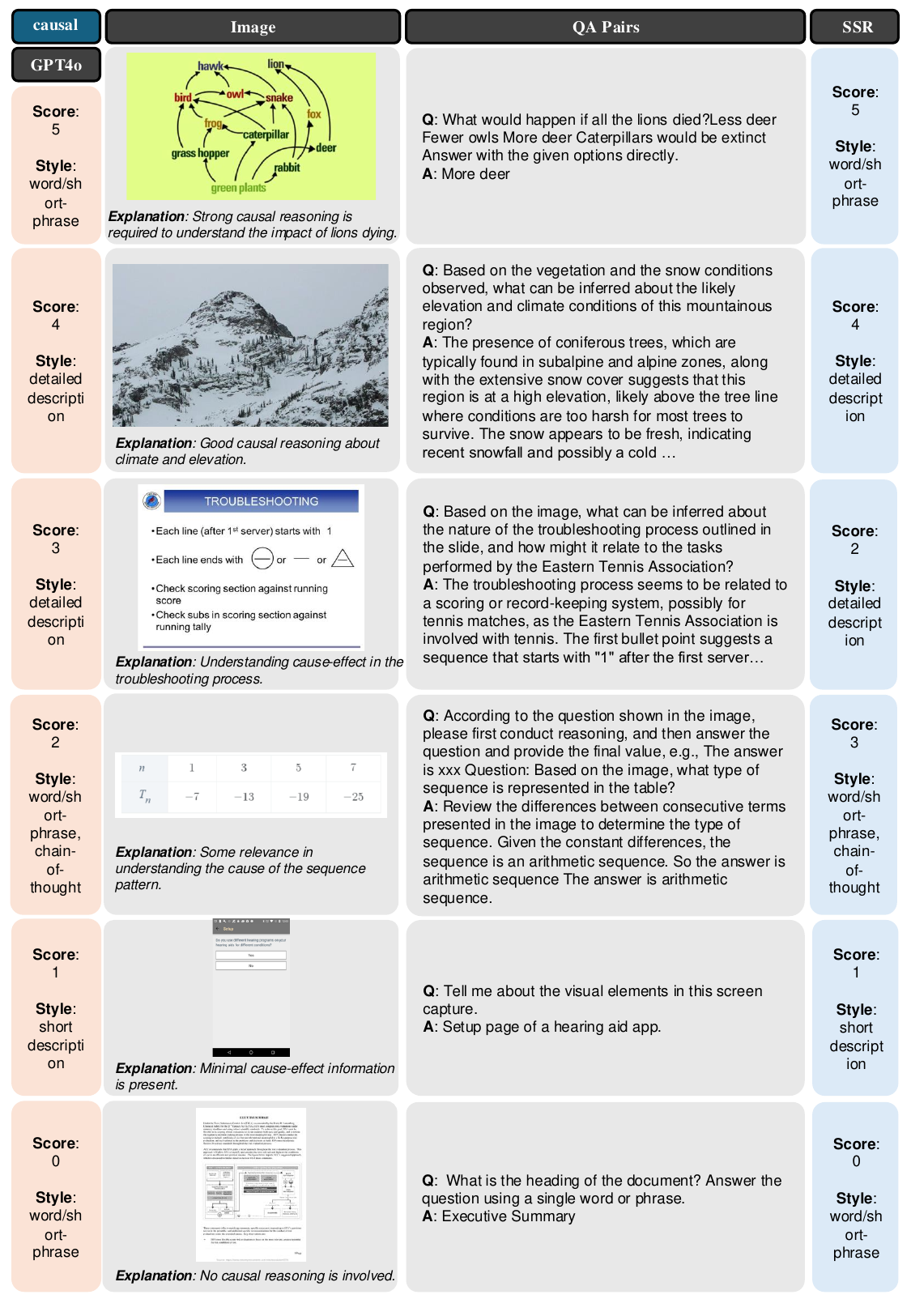}}
\caption{Randomly selected samples from the LLaVA-OVSI data pool within each score group for the \textbf{causal reasoning} capability, ranging from 5 to 0, as evaluated by GPT-4o, along with corresponding score explanations and SSR predictions.}
\label{fig:supp-examples_causal}
\end{center}
\end{figure*}

\begin{figure*}[ht]
\begin{center}
\centerline{\includegraphics[width=0.93\textwidth]{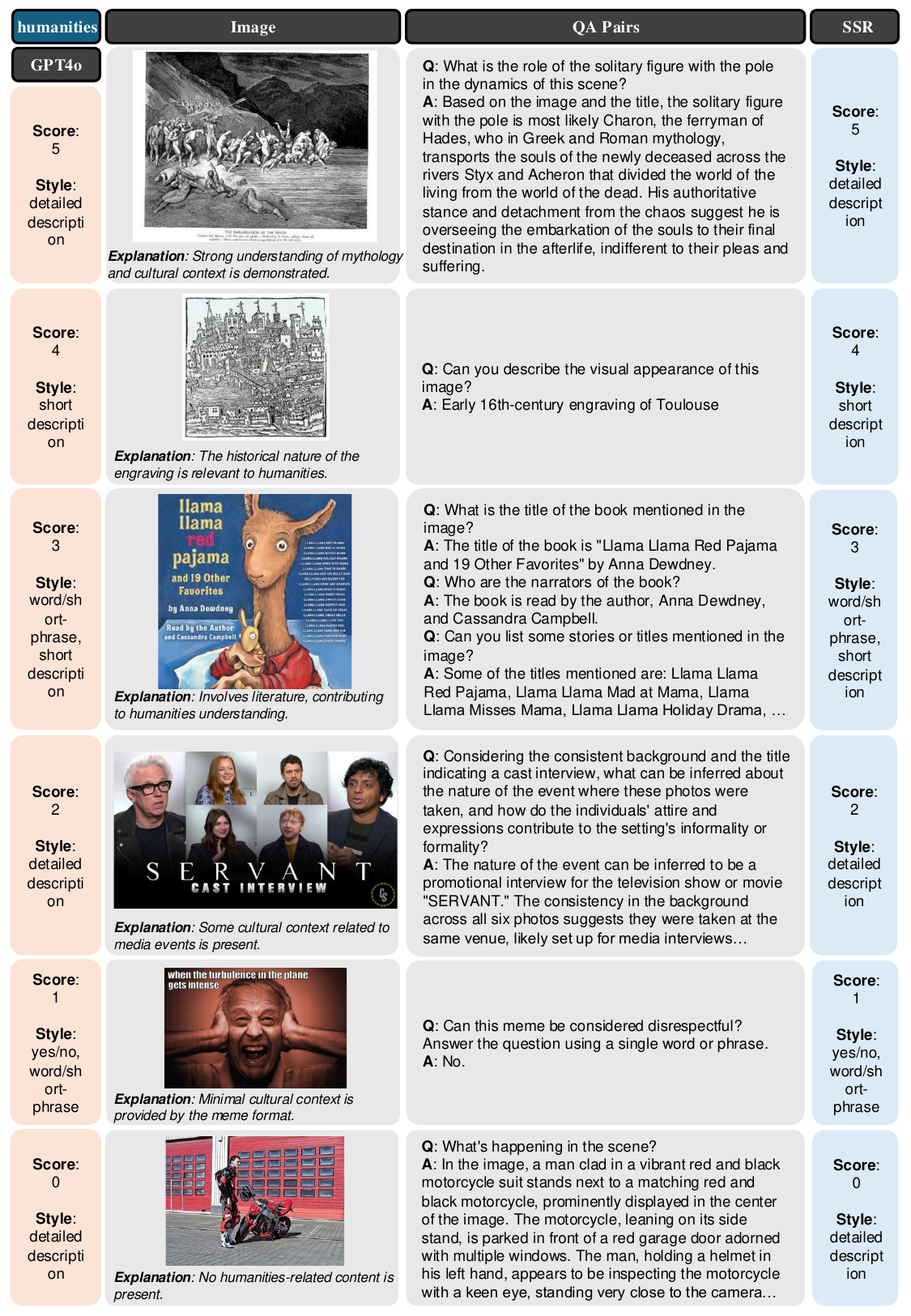}}
\caption{Randomly selected samples from the LLaVA-OVSI data pool within each score group for the \textbf{humanities} capability, ranging from 5 to 0, as evaluated by GPT-4o, along with corresponding score explanations and SSR predictions.}
\label{fig:supp-examples_humanities}
\end{center}
\end{figure*}

\begin{figure*}[ht]
\begin{center}
\centerline{\includegraphics[width=0.93\textwidth]{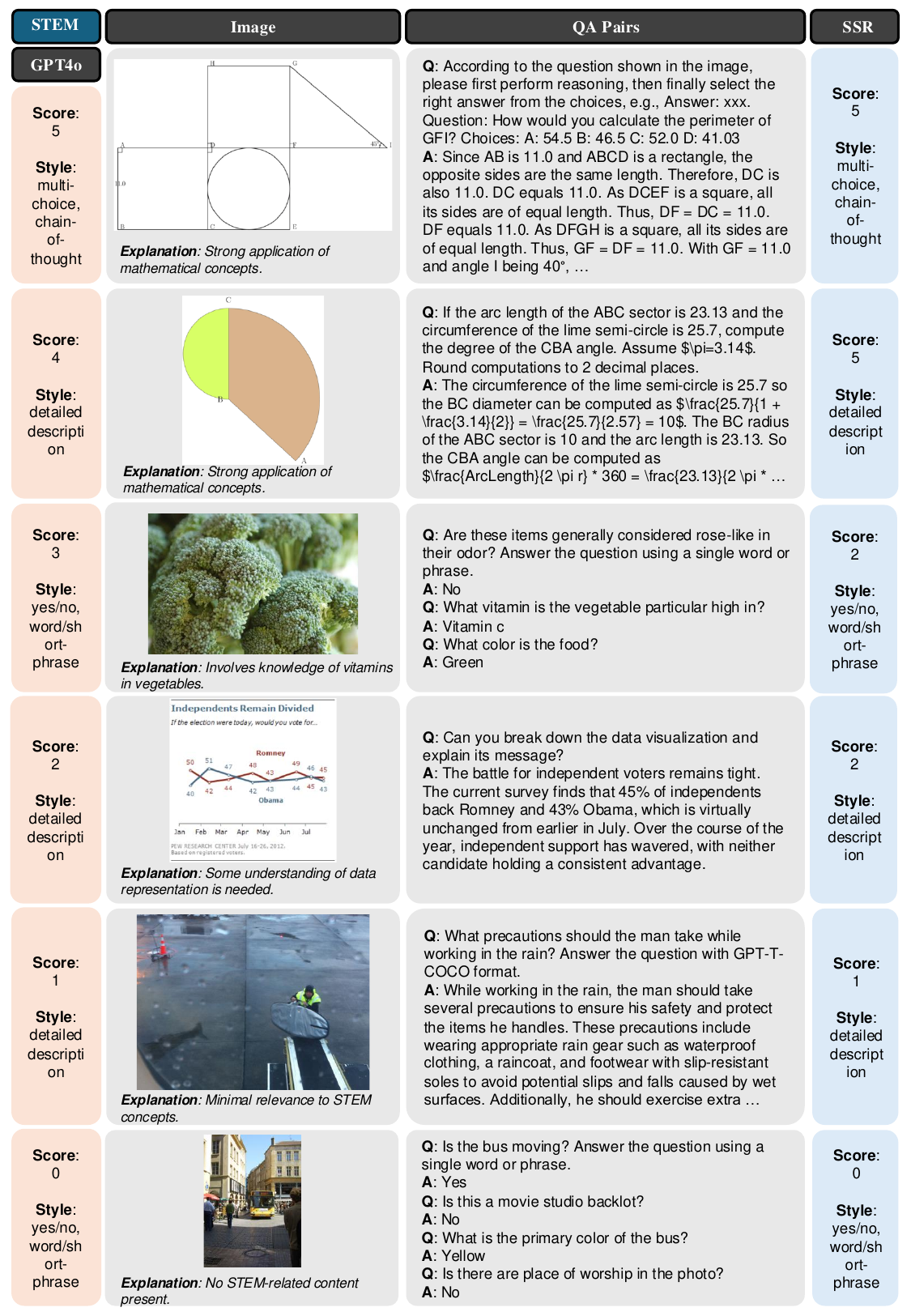}}
\caption{Randomly selected samples from the LLaVA-OVSI data pool within each score group for the \textbf{stem knowledge} capability, ranging from 5 to 0, as evaluated by GPT-4o, along with corresponding score explanations and SSR predictions.}
\label{fig:supp-examples_stem}
\end{center}
\end{figure*}

\begin{figure*}[ht]
\begin{center}
\centerline{\includegraphics[width=0.93\textwidth]{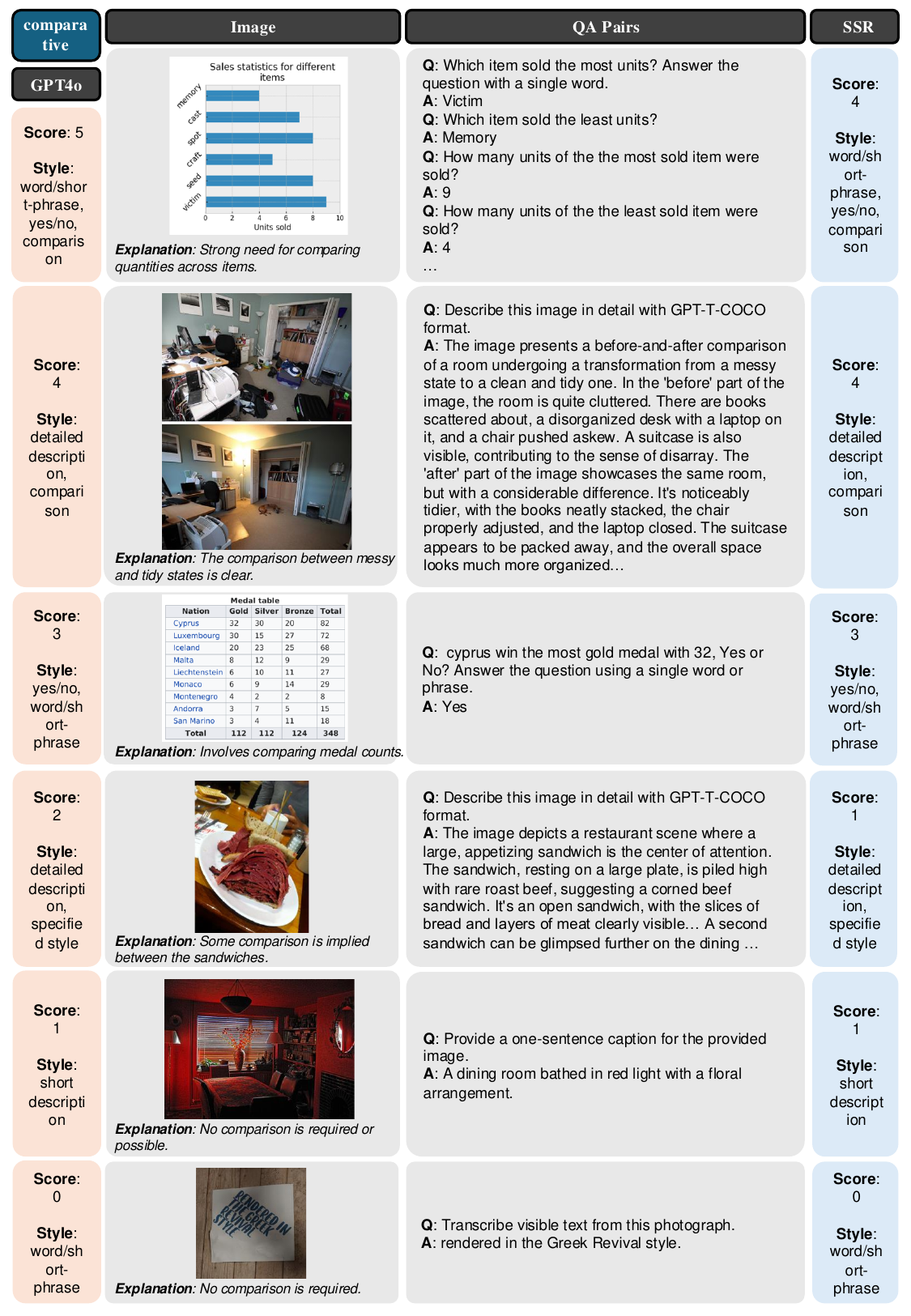}}
\caption{Randomly selected samples from the LLaVA-OVSI data pool within each score group for the \textbf{comparative analysis} capability, ranging from 5 to 0, as evaluated by GPT-4o, along with corresponding score explanations and SSR predictions.}
\label{fig:supp-examples_comparative}
\end{center}
\end{figure*}

\begin{figure*}[ht]
\begin{center}
\centerline{\includegraphics[width=0.93\textwidth]{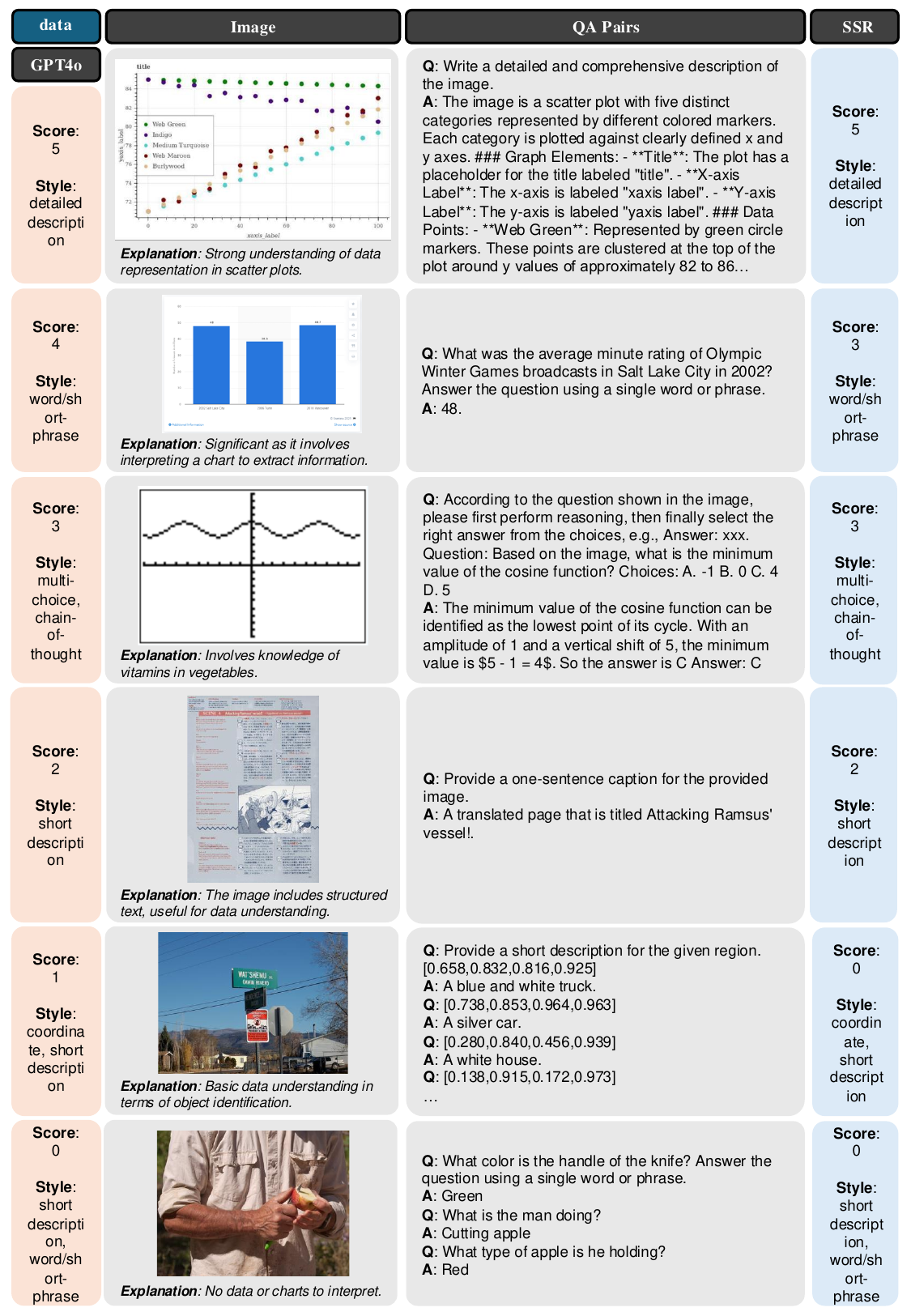}}
\caption{Randomly selected samples from the LLaVA-OVSI data pool within each score group for the \textbf{data understanding} capability, ranging from 5 to 0, as evaluated by GPT-4o, along with corresponding score explanations and SSR predictions.}
\label{fig:supp-examples_data}
\end{center}
\end{figure*}

\begin{figure*}[ht]
\begin{center}
\centerline{\includegraphics[width=0.93\textwidth]{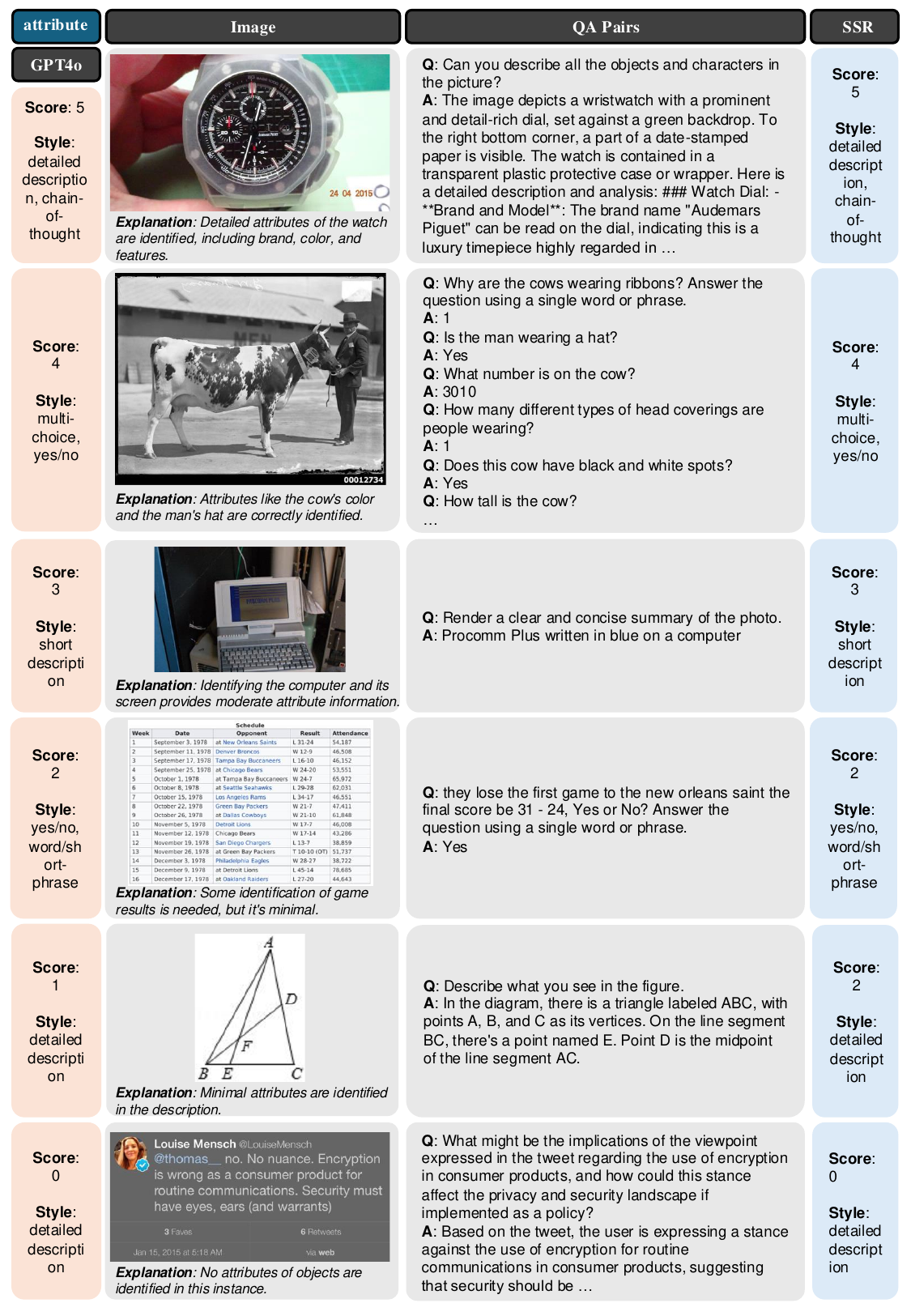}}
\caption{Randomly selected samples from the LLaVA-OVSI data pool within each score group for the \textbf{attribute identification} capability, ranging from 5 to 0, as evaluated by GPT-4o, along with corresponding score explanations and SSR predictions.}
\label{fig:supp-examples_attribute}
\end{center}
\end{figure*}

\begin{figure*}[ht]
\begin{center}
\centerline{\includegraphics[width=0.93\textwidth]{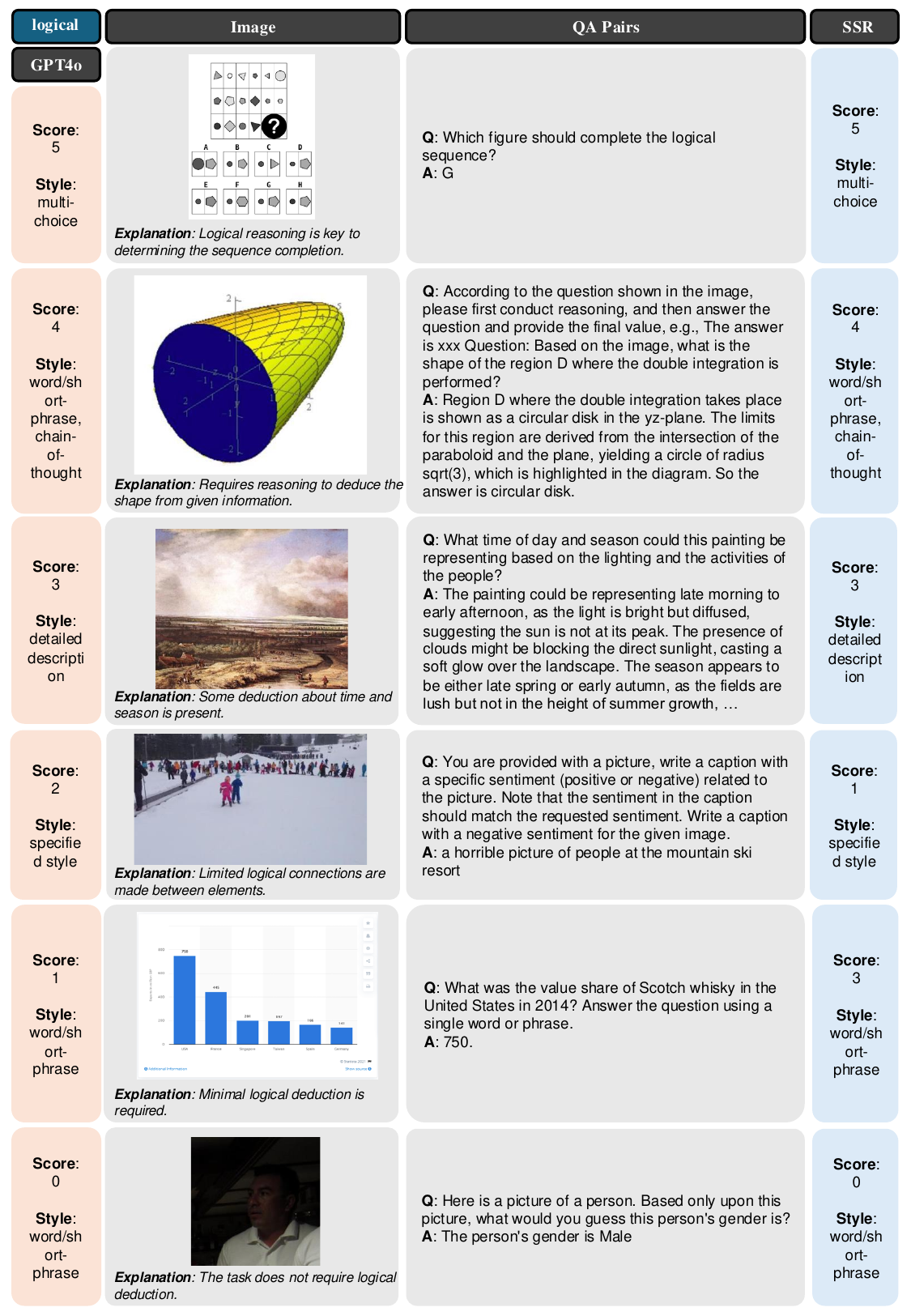}}
\caption{Randomly selected samples from the LLaVA-OVSI data pool within each score group for the \textbf{logical deduction} capability, ranging from 5 to 0, as evaluated by GPT-4o, along with corresponding score explanations and SSR predictions.}
\label{fig:supp-examples_logical}
\end{center}
\end{figure*}

\begin{figure*}[ht]
\begin{center}
\centerline{\includegraphics[width=0.93\textwidth]{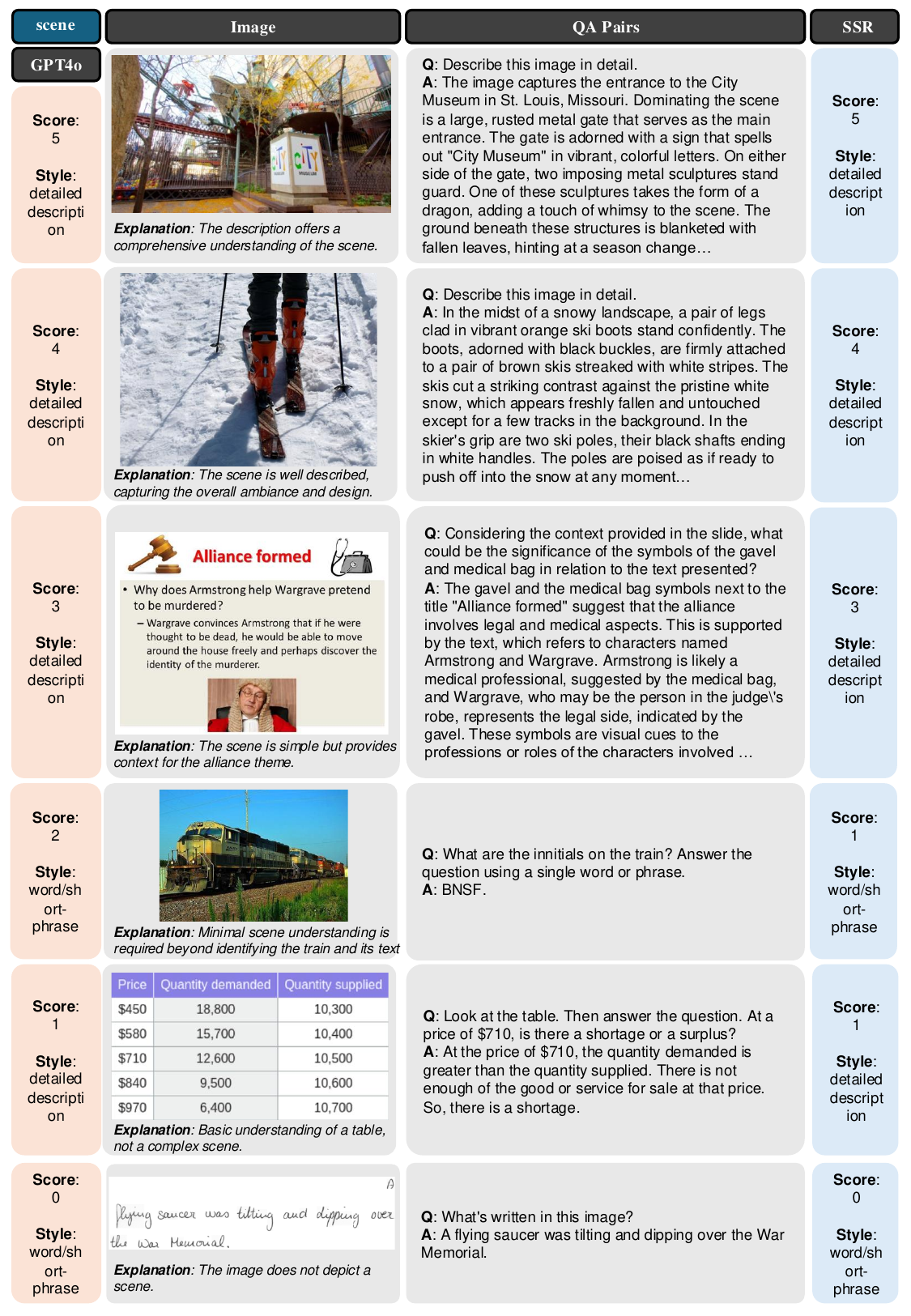}}
\caption{Randomly selected samples from the LLaVA-OVSI data pool within each score group for the \textbf{scene understanding} capability, ranging from 5 to 0, as evaluated by GPT-4o, along with corresponding score explanations and SSR predictions.}
\label{fig:supp-examples_scene}
\end{center}
\end{figure*}

\begin{figure*}[ht]
\begin{center}
\centerline{\includegraphics[width=0.93\textwidth]{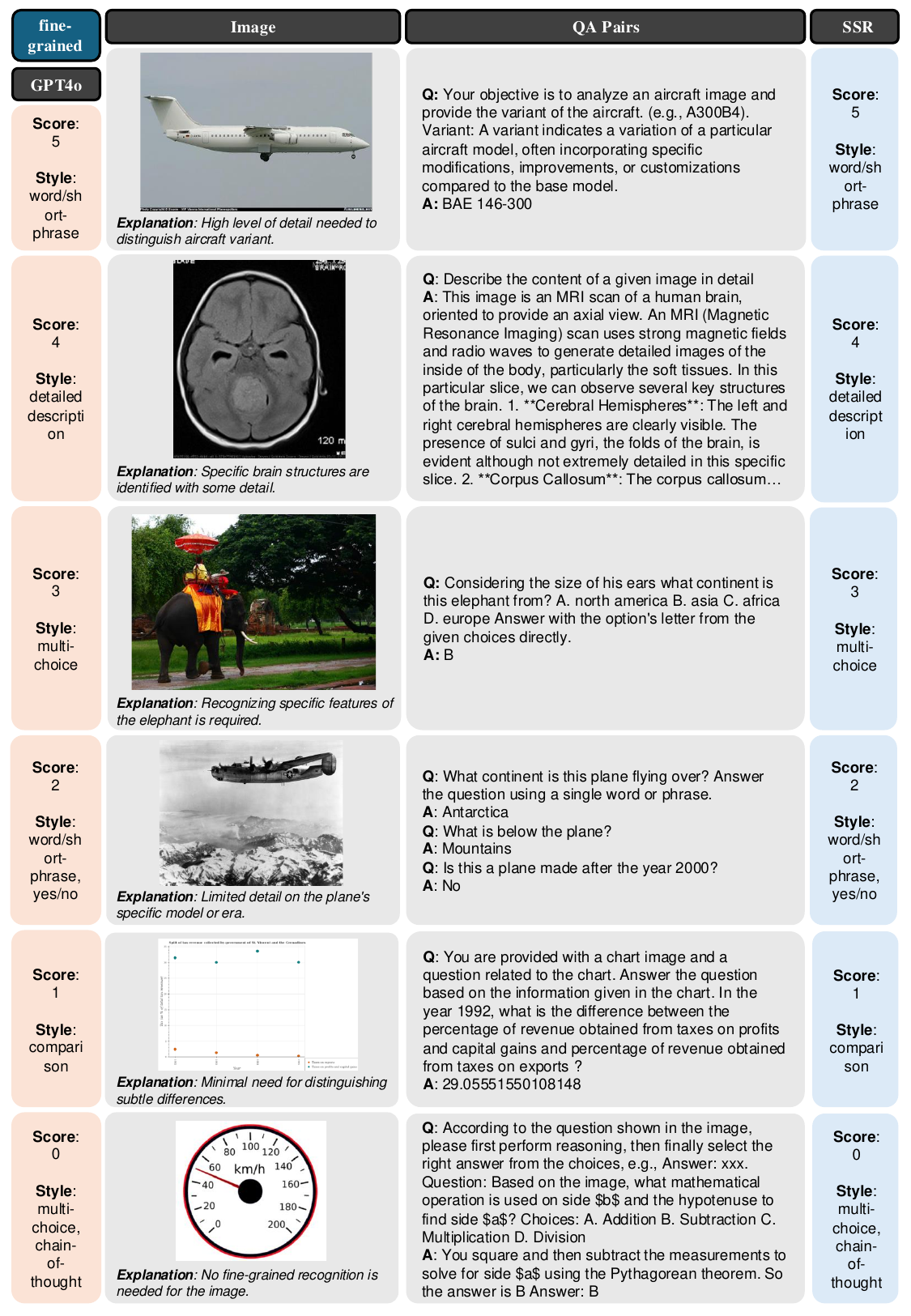}}
\caption{Randomly selected samples from the LLaVA-OVSI data pool within each score group for the \textbf{fine-grained recognition} capability, ranging from 5 to 0, as evaluated by GPT-4o, along with corresponding score explanations and SSR predictions.}
\label{fig:supp-examples_fine-grained}
\end{center}
\end{figure*}

\begin{figure*}[ht]
\begin{center}
\centerline{\includegraphics[width=0.93\textwidth]{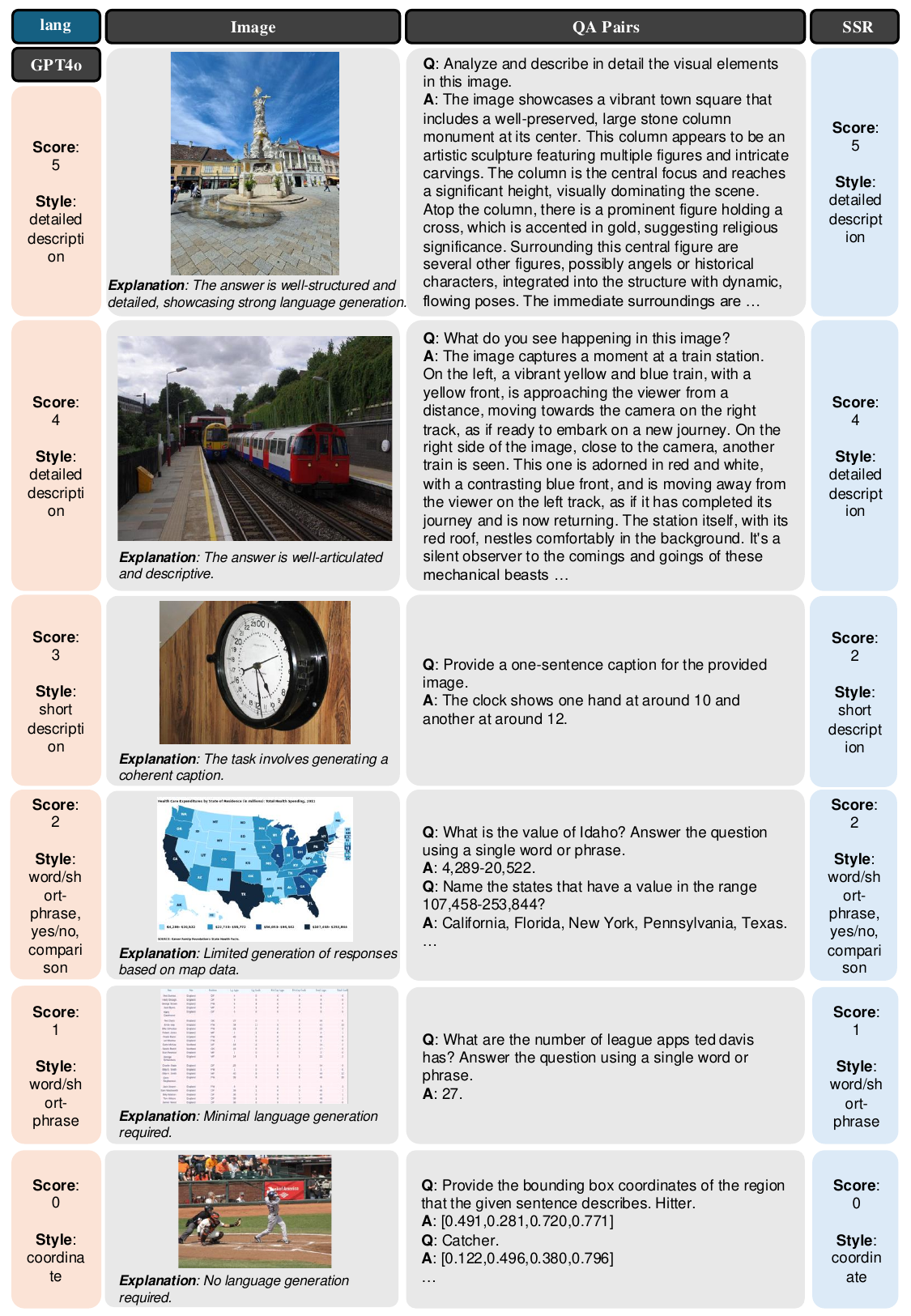}}
\caption{Randomly selected samples from the LLaVA-OVSI data pool within each score group for the \textbf{language generation} capability, ranging from 5 to 0, as evaluated by GPT-4o, along with corresponding score explanations and SSR predictions.}
\label{fig:supp-examples_lang}
\end{center}
\end{figure*}

\begin{figure*}[ht]
\begin{center}
\centerline{\includegraphics[width=0.93\textwidth]{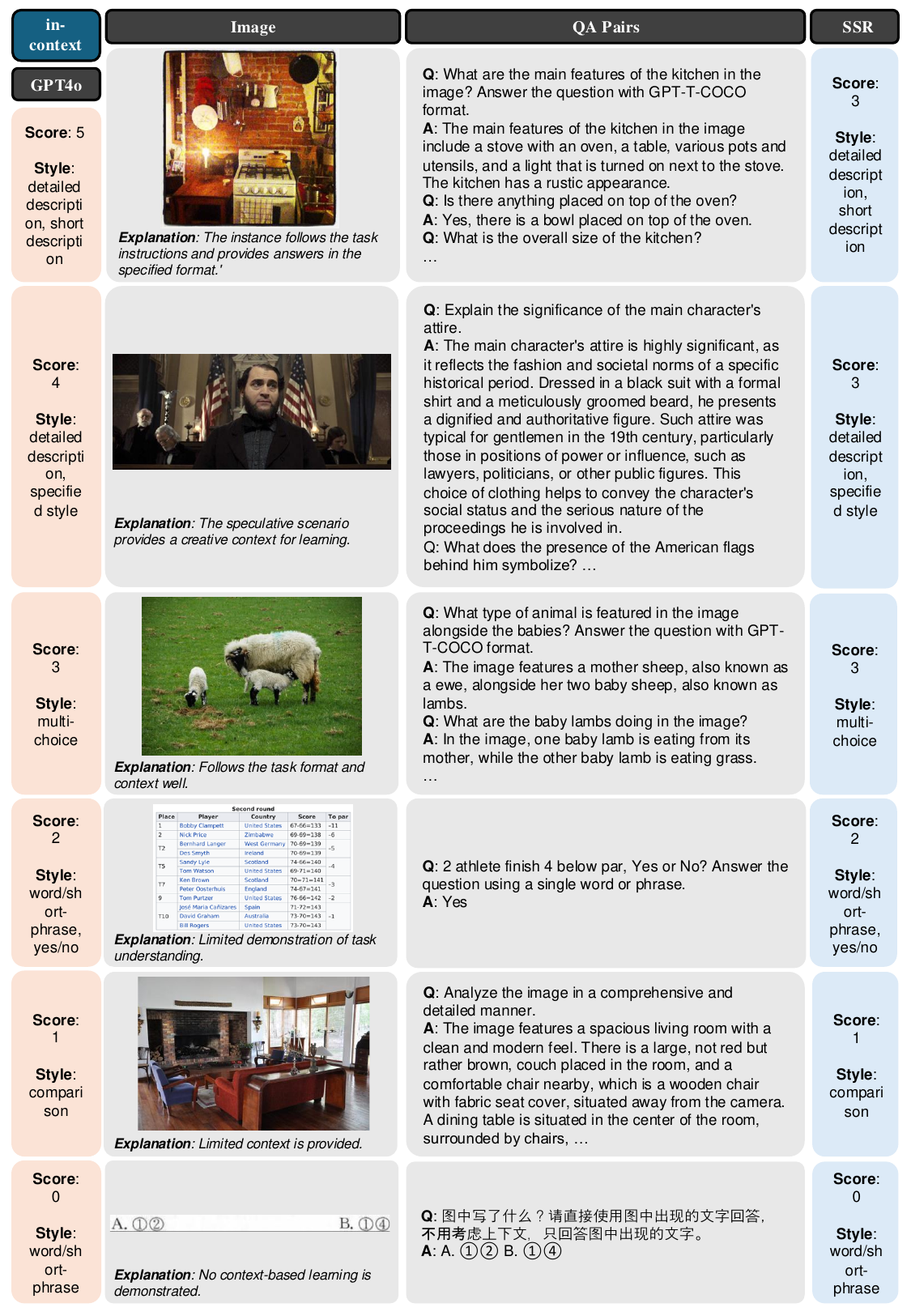}}
\caption{Randomly selected samples from the LLaVA-OVSI data pool within each score group for the \textbf{in-context learning} capability, ranging from 5 to 0, as evaluated by GPT-4o, along with corresponding score explanations and SSR predictions.}
\label{fig:supp-examples_in-context}
\end{center}
\end{figure*}

For each capability, we employed stratified random sampling based on the GPT-4o scoring scale (0--5) to present examples. The resulting visualizations, comprising image-text pairs, GPT-4o scores, identified styles, explanations, and our SSR judgments, are presented in Fig.~\ref{fig:supp-examples_activity}--\ref{fig:supp-examples_ocr}.

The consistent quality observed across these randomly sampled instances underscores the robustness and fidelity of our SSR judgments. Qualitative inspection further confirms that, for both the proprietary judge and our fine-tuned scorers, discrepancies between adjacent scores often reflect negligible semantic differences that are indistinguishable even to human evaluators. For instance, the distinction between logical reasoning scores of 1 and 2 in the 4th example of Fig.~\ref{fig:supp-examples_logical}, or between comparative analysis scores of 4 and 5 in the 1st example of Fig.~\ref{fig:supp-examples_comparative}, is subtle. Thus, while the model may occasionally diverge on the exact integer, it accurately captures the semantic caliber of the data, ensuring robust quality stratification.

\begin{figure}[ht]
\begin{center}
\centerline{\includegraphics[width=1\columnwidth]{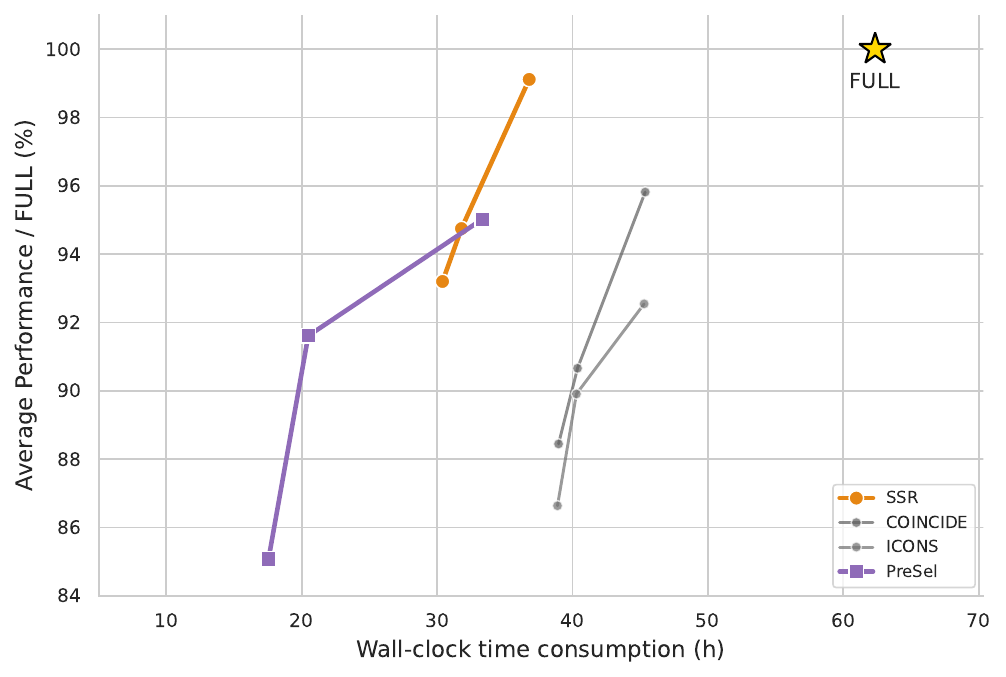}}
\caption{Comparison of selection methods on average performance and total time consumption.}
\label{fig:supp-time}
\end{center}
\end{figure}

\section{Capability Scaling and Customization}
\label{sec:sup-scale}

Beyond the data-\textit{quantity} scalability already shown on the 2.6M pool in Sec.~\ref{sec:exp-main}, real-world curation often demands \textit{adding} new capabilities into an existing pool. We use OCR as a held-out case study; a comprehensive sweep across all 14 capabilities would require substantially more compute and is outside the scope of this work, and we do not claim that all capabilities customize identically.

\myparagraph{Quantity axis.} Along the quantity axis (Fig.~\ref{fig:line-scale}), SSR delivers consistently superior MLLM performance from extreme cold-start (1\%) to resource-abundant (40\%, 50\%) budgets, extending the million-scale scalability of Sec.~\ref{sec:exp-main} to a wider budget range.

\myparagraph{Capability axis.} Along the capability axis, we elicit OCR-specific judgments and let SSR prioritize high-scoring OCR samples: the added samples yield steady improvements on OCR-related benchmarks while maintaining (or further extending) gains on general benchmarks. SSR's modular profile therefore supports targeted capability expansion without sacrificing prior coverage.

\section{Efficiency Comparison}
\label{sec:sup-efficiency}

Fig.~\ref{fig:supp-time} presents a time-performance analysis, comparing multi-modal data selection methods against the full fine-tuning baseline on LLaVA-OVSI. For each method, the total wall-clock time encompasses both the data selection phase and the subsequent training of the target model on the selected budget.
The results demonstrate that SSR significantly outperforms all baselines in efficiency and effectiveness. The gains are particularly striking when compared to full fine-tuning, which requires 62.35 hours on 64 H100 GPUs. Even when executing the entire pipeline from scratch, SSR achieves 93.20\%, 94.75\%, and 99.11\% of the full performance within 30.4, 31.8 and 36.8 hours respectively. This demonstrates that our method can recover nearly all of the performance of the full dataset while reducing the end-to-end training time by over 40\%.

A key architectural advantage of explicit data valuation is its efficiency. While each data point of SSR in Fig.~\ref{fig:supp-time} accounts for the end-to-end time, the two primary computational costs, model fine-tuning and data valuation, are one-time, amortizable costs.
In practice, the fine-tuning of the scorer/styler can be entirely obviated by directly prompting an open-weight MLLM or by directly applying our pre-trained models to a new domain, validated in Sec.~\ref{sec:exp-application} of the main paper. Data valuation is also a one-time computation, allowing near-instantaneous adaptation to different budget settings through a simple re-ranking of the pre-computed results.
This stands in sharp contrast to prior work. COINCIDE~\cite{coincide}, for example, must re-run its time-consuming clustering and selection process (over an hour on million-scale datasets) for each new selection task. The inefficiency of ICONS~\cite{wu2024icons} is even more pronounced, as it necessitates computing gradients on validation data for each target benchmark, followed by a large-scale similarity estimation against the entire candidate pool. This makes its approach not only computationally prohibitive but also fundamentally tied to specific validation sets.

\section{Validation on Benchmark Test Splits}
\label{sec:sup-test-res}

\begin{table*}[htbp]
\fontsize{8}{9.6}\selectfont
    \centering
    \begin{tabular}{l|cccaa}
    \toprule
    &MMBench\textsubscript{en-v1.1-test} & MMMU\textsubscript{test} & MMT-Bench\textsubscript{all}& $>$Rand & /FULL\\

    \midrule
     \multicolumn{6}{c}{\textbf{Budget: 5\%}} \\
    \midrule

Random  & 73.45 & 40.37 & 59.98 & - & 96.32\% \\
Deita  & \underline{73.97} & 36.30 & 56.57 & 1/3 & 91.40\% \\
COINCIDE  & 73.09 & 40.30 & 57.47 & 0/3 & 94.74\% \\
ICONS  & 69.04 & 37.30 & 57.49 & 0/3 & 90.63\% \\
SSR  & \underline{\textbf{75.84}} & \underline{\textbf{41.30}} & \underline{\textbf{60.10}} & \textbf{3/3} & \underline{\textbf{98.15}}\% \\

    \midrule
     \multicolumn{6}{c}{\textbf{Budget: 10\%}} \\
    \midrule
    
Random  & 74.55 & 40.40 & 60.54 & - & 97.12\% \\
Deita  & \underline{75.17} & 37.00 & 57.40 & 1/3 & 92.92\% \\
COINCIDE  & 74.44 & \underline{40.40} & 58.68 & 1/3 & 96.05\% \\
ICONS  & 71.90 & 38.80 & 58.66 &1/3 &  93.68\% \\
SSR  & \underline{\textbf{76.05}} & \underline{\textbf{40.90}} & \underline{\textbf{60.68}} & \textbf{3/3} & \underline{\textbf{98.23}}\% \\

    \midrule
     \multicolumn{6}{c}{\textbf{Budget: 30\%}} \\
    \midrule
Random  & 77.33 & \textbf{41.13} & 59.59 & - & 98.36\% \\
Deita  & 76.88 & 40.00 & 59.56 &0/3 &  97.24\% \\
COINCIDE  & \underline{\textbf{78.18}} & 41.00 & \underline{59.77} &\textbf{2/3}& \underline{98.71}\% \\
ICONS  & 72.21 & 40.40 & \underline{59.70} &1/3& 95.67\% \\
SSR  & \underline{78.13} & 41.10 & \underline{\textbf{59.80}} &\textbf{2/3}& \underline{\textbf{98.78}}\% \\

    \midrule
     \multicolumn{6}{c}{\textbf{FULL}} \\
    \midrule
    LLaVA-OVSI  & 79.27 & 41.40 & 60.70 & - & 100\% \\
    \bottomrule
    \end{tabular}
    \caption{Performance comparison on the benchmarks with online test splits conducted across varying budgets of 5\%, 10\% and 30\% of LLaVA-OVSI. We highlight the best result in \textbf{boldface} and \underline{underline} the result if it beats the random baseline.
    The column {$>$Rand} presents the number of benchmarks where the method exceeds random sampling, and {/FULL} compares the performance of sampled data with that of the FULL dataset.}
    \label{tab:test_split}
\end{table*}

The comparative results presented in Tab.~\ref{tab:main} of the main paper were conducted within a unified framework under identical data budgets across our method and all baselines, ensuring a fair comparison. Inside this batch, the relative ordering between methods is already sufficient to validate the effectiveness of each sampling strategy.
To further substantiate the effectiveness of SSR, we additionally evaluate the top competitors on the available test splits that require online assessment, specifically MMBench$_{en-v1.1}$~\cite{liu2023mmbench}, MMMU~\cite{yue2023mmmu}, and MMT-Bench~\cite{mmtbench}.

Consistent with the trends from Sec.~\ref{sec:exp-main}, Tab.~\ref{tab:test_split} shows that SSR not only consistently outperforms or matches the competitors across all data budgets, but its advantage is particularly pronounced in the challenging low-data regime.
Unlike methods that rely on model-specific features or pre-heated finetuning, our approach is grounded in explicit capabilities and interactional styles, which translates into superior efficiency and robustness on the held-out test splits.
It is worth noting that the trends on MMMU suggest a performance ceiling for the current data pool. Achieving further breakthroughs on this benchmark will likely necessitate augmenting the training data with external sources rich in college-level, multidisciplinary knowledge, an extension that the capability-extensible SSR framework readily supports (Supp.~\ref{sec:sup-scale}).

\section{Analysis of Selected Subsets}
\label{sec:sup-data-analysis}

Having validated SSR's transferability (Sec.~\ref{sec:exp-transfer}), scalability (Supp.~\ref{sec:sup-scale}), efficiency (Supp.~\ref{sec:sup-efficiency}), and effectiveness (Supp.~\ref{sec:sup-test-res}), we now look \emph{inside} the selected subsets to understand \emph{why} the selection works. We examine the distribution of the resulting subsets along three complementary axes: capability scores (Supp.~\ref{sec:sup-data-score}), interaction styles (Supp.~\ref{sec:sup-data-style}), and data sources (Supp.~\ref{sec:sup-data-source}).

\begin{figure*}[htb]
\centering
    \subfigure[Distribution of total scores across all capabilities for SSR-10\%]{\label{fig:sup-score-SSR-select10-sum}\includegraphics[width=0.24\linewidth]{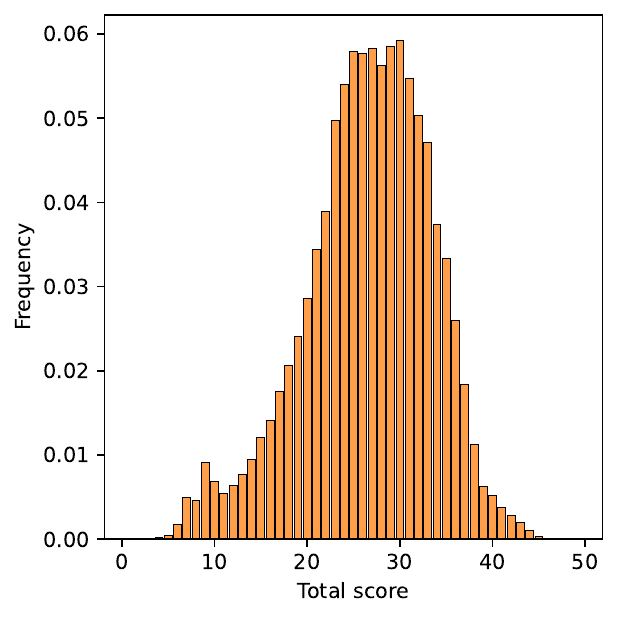}}
    \subfigure[Distribution of total scores across all capabilities for SSR-30\%]{\label{fig:sup-score-SSR-select30-sum}\includegraphics[width=0.24\linewidth]{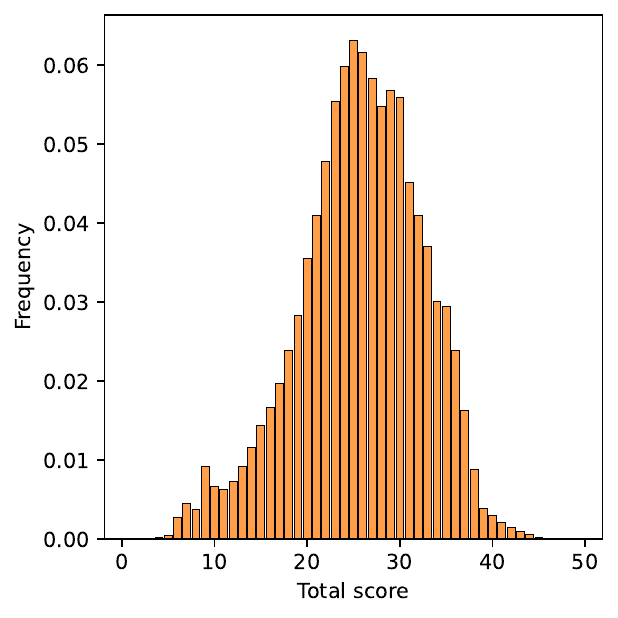}}
    \subfigure[Distribution of the max score across all capabilities for SSR-10\%]{\label{fig:sup-score-SSR-select10-max}\includegraphics[width=0.24\linewidth]{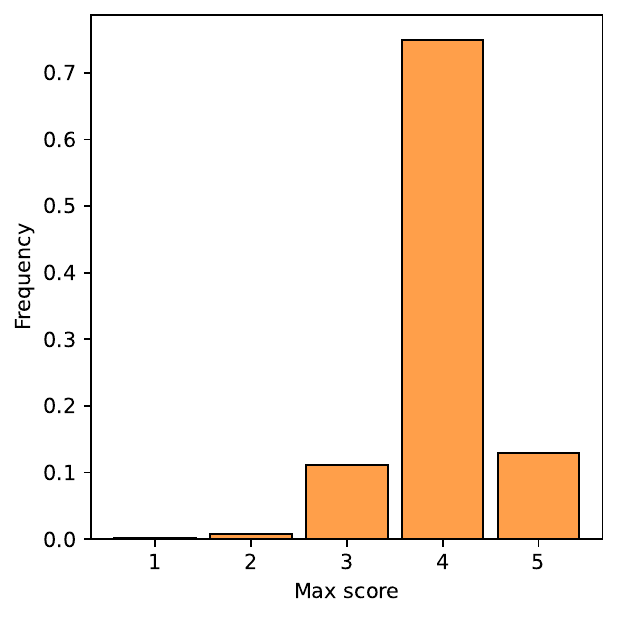}}
    \subfigure[Distribution of the max score across all capabilities for SSR-30\%]{\label{fig:sup-score-SSR-select30-max}\includegraphics[width=0.24\linewidth]{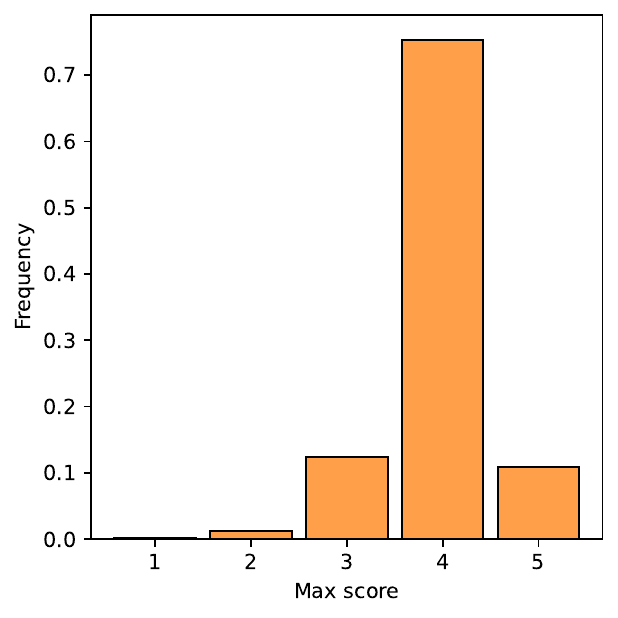}}
\caption{Sample-level score distribution of the selected SSR-10\% and SSR-30\%.}
\label{fig:sup-score-SSR-select}
\end{figure*}

\subsection{Score Distribution}
\label{sec:sup-data-score}
To illustrate the information distribution of the data obtained by our SSR sampler, Fig.~\ref{fig:sup-score-SSR-select} presents the score distributions for subsets at 10\% and 30\% sampling ratios. These statistics are derived from the predictions of the SSR model (trained on 15\% annotation data) following our main configuration.

As depicted in Fig.~\ref{fig:sup-score-SSR-select10-sum} and Fig.~\ref{fig:sup-score-SSR-select30-sum}, the sample-aggregated capability scores exhibit a distinct bell-shaped curve. This distribution pattern arises from the natural scarcity of samples with maximum scores. After exhausting the top tier, the sampler satisfies the remaining budget with mid-range samples, converging toward a normal distribution centered on the high-quality midrange.
Notably, our sampler selects data points even if their aggregate contribution is low, provided they excel in specific dimensions. This behavior is critical because specialized capabilities (\eg STEM) often have less correlation with other skills, resulting in lower total scores for these ``specialist'' samples. By coupling an explicit capability subspace with a round-robin selection mechanism, we specifically target these minority dimensions, preventing them from being overshadowed by generalist samples. By enforcing coverage across all dimensions, our selected subsets can address the niche challenges found in complex benchmarks, thereby broadening the model's operational spectrum.
Fig.~\ref{fig:sup-score-SSR-select10-max} and Fig.~\ref{fig:sup-score-SSR-select30-max} provide further insight into our sampler's strategy by illustrating the distribution of sample-wise maximum capability scores. The pronounced concentration of scores around 4, coupled with the deliberate inclusion of lower-scoring samples, underscores SSR's capacity to strike a critical balance between information efficiency and capability richness.

By integrating these robust specialized capabilities alongside peak-value samples, SSR ensures a rich and efficient capability representation, ultimately enhancing the final model's overall performance and adaptability.

\subsection{Style Distribution}
\label{sec:sup-data-style}
\begin{figure*}[ht]
\begin{center}
\centerline{\includegraphics[width=0.8\textwidth]{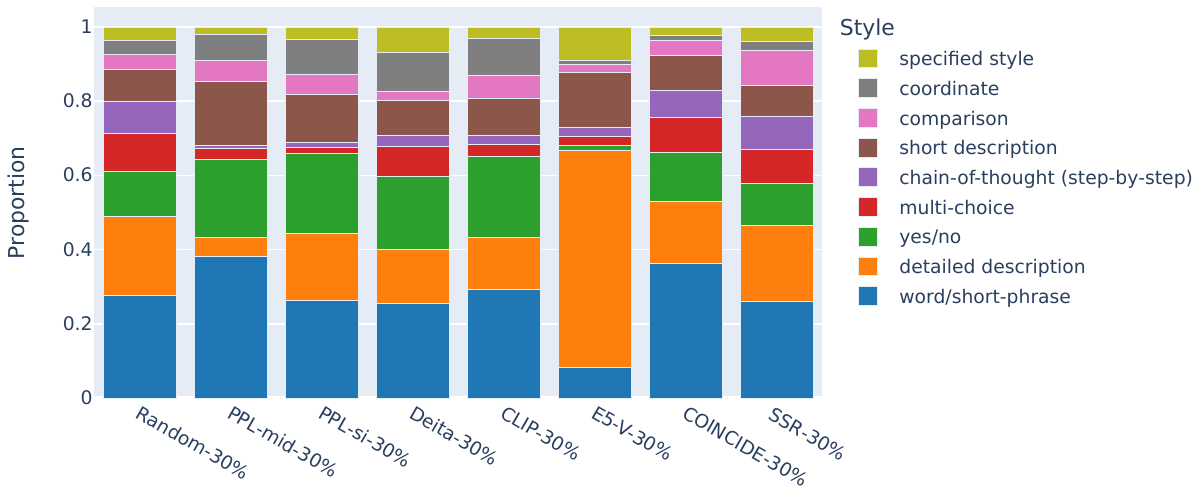}}
\caption{Comparison of data style statistics between our selection and those of competitors.}
\label{fig:supp-dist-bin-style-comp}
\end{center}
\end{figure*}

Fig.~\ref{fig:supp-dist-bin-style-comp} presents a comparative analysis of interaction style distributions. As illustrated, SSR exhibits a distinct pattern characterized by a balanced representation across diverse styles. This comprehensive coverage is critical for the objective of the SFT stage, directly contributing to the robustness of the final model.
In contrast, other sampling methods display skewed distributions, disproportionately favoring specific styles such as \textit{detailed description} (which inflates token counts) or \textit{yes/no} and \textit{short-phrase} answers (ubiquitous in evaluation benchmarks). Such imbalances can limit the versatility and generalization of the resulting models. Notably, SSR achieves a more equitable distribution, successfully preserving complex styles like \textit{comparison} and \textit{chain-of-thought} that are vital for reasoning tasks. This balance highlights our method's capacity to support a broad spectrum of downstream applications, establishing SSR as a general-purpose tool for effective dataset curation.

\subsection{Data Source Distribution}
\label{sec:sup-data-source}
\begin{figure*}[ht]
\begin{center}
\centerline{\includegraphics[width=0.8\textwidth]{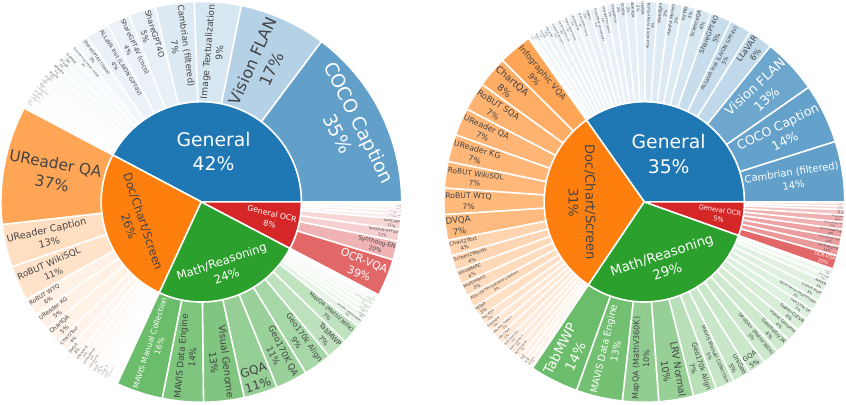}}
\caption{Data source statistics of the original LLaVA-OVSI data pool (L) and our SSR-10\% (R).}
\label{fig:supp-dist-pie-src}
\end{center}
\end{figure*}

Given the inherent heterogeneity of multi-modal data sources and the rigorous demands of comprehensive evaluation benchmarks, maintaining diversity during instruction fine-tuning is paramount. Adhering to the original data hierarchy defined in~\cite{li2024LLaVAOneVisionEasyVisual}, we present a comparative statistical breakdown of the full data pool versus our 10\% sampled subset in Tab.~\ref{tab:dist_source}, with corresponding visualizations in Fig.~\ref{fig:supp-dist-pie-src}.

The selected subset reveals a distinct shift toward a more balanced distribution under SSR. As shown in the original LLaVA-OVSI (left), the dataset is dominated by the General category (42\%), with COCO Caption alone accounting for 35\%. In contrast, the SSR-sampled subset (right) achieves a significantly more equitable distribution. The General category is modulated to 35\%, with COCO Caption reduced to 14\%, while high-value categories see substantial growth: Math/Reasoning expands from 24\% to 29\%, and Doc/Chart/Screen increases from 26\% to 31\%.

\begin{figure*}[ht]
\begin{center}
\hspace*{-0.55in}
\centerline{\includegraphics[width=0.72\textwidth]{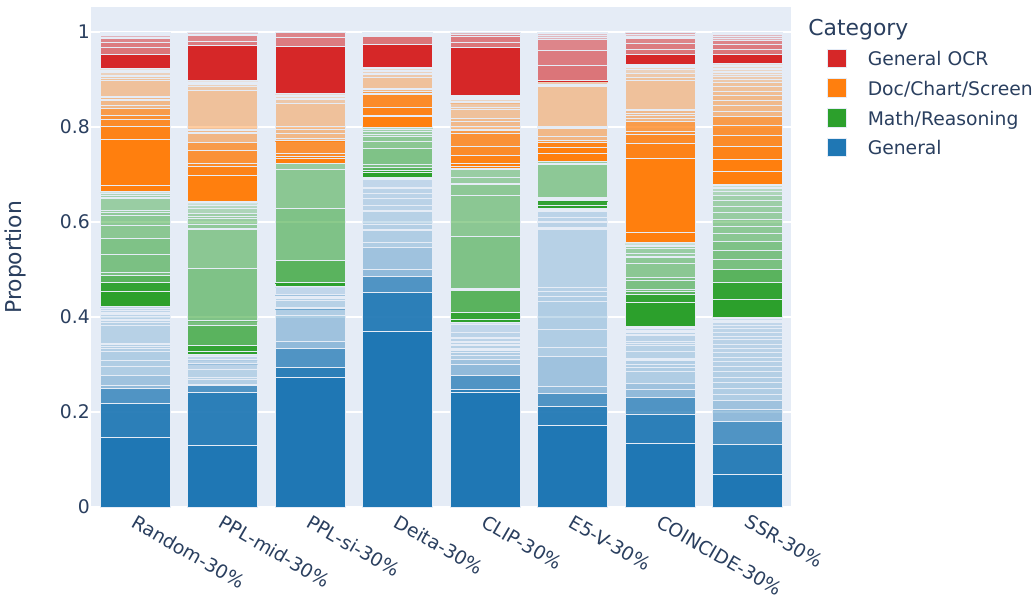}}
\caption{Comparison of data source statistics between our selection and those of competitors. For brevity, the figure displays only a small subset of the legends. Data sources of the same category are represented by shared color schemes, in accordance with Fig.~\ref{fig:supp-dist-pie-src}.}
\label{fig:supp-dist-bin-comp}
\end{center}
\end{figure*}

Fig.~\ref{fig:supp-dist-bin-comp} further contrasts our approach with comparative methods. While SSR maintains balance across diverse sources, baseline methods exhibit clear biases: Deita and E5-V remain heavily concentrated in dominant general data; PPL and CLIP disproportionately favor math/reasoning (especially Visual Genome); and COINCIDE is skewed toward document/chart data. The effective reallocation of the data budget by SSR underscores its ability to achieve a diverse and representative training set. This diversification directly enhances the target model's robustness on general instruction-following while boosting adaptability for challenging tasks such as mathematical problem-solving and infographic reasoning.

In conclusion, the distribution analysis presented in this section confirms that SSR effectively provides highly informative subsets spanning rich capabilities and diverse instruction styles. This approach significantly advances both the data efficiency and curation of MLLM instruction tuning.

\onecolumn %
{\small
\begin{center} %

\bottomcaption{Number of samples and proportions of sources for LLaVA-OVSI and SSR selected subsets. \label{tab:dist_source}}

\tablefirsthead{
    \toprule
    \multicolumn{1}{c}{} & \multicolumn{2}{c}{\textbf{LLaVA-OVSI}} & \multicolumn{2}{c}{\textbf{SSR-10\%}} & \multicolumn{2}{c}{\textbf{SSR-30\%}} \\
    \cmidrule(lr){2-3} \cmidrule(lr){4-5} \cmidrule(lr){6-7}
    Source & \# Samples & Prop. & \# Samples & Prop. & \# Samples & Prop. \\
    \midrule
}

\tablehead{
    \multicolumn{7}{l}{{Continued from previous page}} \\
    \toprule
    \multicolumn{1}{c}{} & \multicolumn{2}{c}{LLaVA-OVSI} & \multicolumn{2}{c}{SSR-10\%} & \multicolumn{2}{c}{SSR-30\%} \\
    \cmidrule(lr){2-3} \cmidrule(lr){4-5} \cmidrule(lr){6-7}
    Source & \# Samples & Prop. & \# Samples & Prop. & \# Samples & Prop. \\
    \midrule
}

\tabletail{
    \midrule
    \multicolumn{7}{r}{{Continued on next page}} \\
}

\tablelasttail{
    \bottomrule
}

\begin{xtabular}{lcc|cc|cc}
\shrinkheight{-5cm}
\multicolumn{7}{c}{General} \\ \midrule
    COCO Caption & 391219 & 34.97\% & 12828 & 13.83\% & 54707 & 17.21\% \\
    Vision FLAN & 186060 & 16.63\% & 11922 & 12.85\% & 50068 & 15.75\% \\
    Image Textualization & 99573 & 8.90\% & 1523 & 1.64\% & 7760 & 2.44\% \\
    Cambrian (filtered) & 83125 & 7.43\% & 12997 & 14.01\% & 39517 & 12.43\% \\
    ShareGPT4O & 57284 & 5.12\% & 4533 & 4.89\% & 15400 & 4.84\% \\
    ShareGPT4V (coco) & 50017 & 4.47\% & 1510 & 1.63\% & 9755 & 3.07\% \\
    ALLaVA Inst (LAION GPT4V) & 49990 & 4.47\% & 4696 & 5.06\% & 9976 & 3.14\% \\
    ShareGPT4V (llava) & 29990 & 2.68\% & 1613 & 1.74\% & 9768 & 3.07\% \\
    ALLaVA Inst (Vision FLAN) & 19990 & 1.79\% & 2577 & 2.78\% & 9220 & 2.90\% \\
    LLaVAR & 19790 & 1.77\% & 5231 & 5.64\% & 19509 & 6.14\% \\
    ST-VQA & 17242 & 1.54\% & 2096 & 2.26\% & 7013 & 2.21\% \\
    AOKVQA & 16534 & 1.48\% & 2188 & 2.36\% & 7486 & 2.35\% \\
    Visual7W & 14361 & 1.28\% & 1478 & 1.59\% & 4802 & 1.51\% \\
    WebSight & 9995 & 0.89\% & 2632 & 2.84\% & 8742 & 2.75\% \\
    VisText & 9964 & 0.89\% & 1769 & 1.91\% & 6363 & 2.00\% \\
    TallyQA & 9868 & 0.88\% & 1126 & 1.21\% & 7309 & 2.30\% \\
    ShareGPT4V (sam) & 8990 & 0.80\% & 1862 & 2.01\% & 8451 & 2.66\% \\
    Hateful Memes & 8495 & 0.76\% & 2765 & 2.98\% & 8495 & 2.67\% \\
    LAION GPT4V & 8048 & 0.72\% & 1525 & 1.64\% & 7139 & 2.25\% \\
    LLaVA Pretrain LCS & 6989 & 0.62\% & 1512 & 1.63\% & 6580 & 2.07\% \\
    VizWiz & 6604 & 0.59\% & 2809 & 3.03\% & 5220 & 1.64\% \\
    ScienceQA & 5932 & 0.53\% & 3388 & 3.65\% & 5930 & 1.87\% \\
    IconQA & 2496 & 0.22\% & 2214 & 2.39\% & 2496 & 0.79\% \\
    ShareGPT4V (knowledge) & 1988 & 0.18\% & 1770 & 1.91\% & 1988 & 0.63\% \\
    ShareGPT4V & 1926 & 0.17\% & 1911 & 2.06\% & 1926 & 0.61\% \\
    InterGPS & 1275 & 0.11\% & 1275 & 1.37\% & 1275 & 0.40\% \\
    CLEVR & 700 & 0.06\% & 700 & 0.75\% & 700 & 0.22\% \\
    VQARAD & 308 & 0.03\% & 308 & 0.33\% & 308 & 0.10\% \\
    \midrule\multicolumn{7}{c}{Doc/Chart/Screen} \\ \midrule 
    UReader QA & 252954 & 36.96\% & 5962 & 7.27\% & 21233 & 10.51\% \\
    UReader Caption & 91434 & 13.36\% & 1784 & 2.18\% & 6861 & 3.40\% \\
    RoBUT WikiSQL & 74984 & 10.95\% & 5688 & 6.94\% & 20290 & 10.05\% \\
    RoBUT WTQ & 38241 & 5.59\% & 5622 & 6.86\% & 15094 & 7.47\% \\
    UReader KG & 37550 & 5.49\% & 5872 & 7.16\% & 21335 & 10.56\% \\
    ChartQA & 36577 & 5.34\% & 6669 & 8.14\% & 18550 & 9.18\% \\
    Chart2Text & 26956 & 3.94\% & 3403 & 4.15\% & 8751 & 4.33\% \\
    DVQA & 22000 & 3.21\% & 5489 & 6.70\% & 17219 & 8.52\% \\
    UReader IE & 17322 & 2.53\% & 1060 & 1.29\% & 3307 & 1.64\% \\
    Screen2Words & 15725 & 2.30\% & 3256 & 3.97\% & 9225 & 4.57\% \\
    AI2D (InternVL) & 12403 & 1.81\% & 1530 & 1.87\% & 6715 & 3.32\% \\
    DocVQA & 10194 & 1.49\% & 1999 & 2.44\% & 5272 & 2.61\% \\
    RoBUT SQA & 8509 & 1.24\% & 6148 & 7.50\% & 8509 & 4.21\% \\
    Infographic VQA & 8489 & 1.24\% & 7233 & 8.82\% & 8489 & 4.20\% \\
    MultiHiertt & 7614 & 1.11\% & 2855 & 3.48\% & 7614 & 3.77\% \\
    AI2D (GPT4V Detailed Caption) & 4864 & 0.71\% & 2746 & 3.35\% & 4864 & 2.41\% \\
    AI2D (Original) & 3247 & 0.47\% & 1457 & 1.78\% & 3247 & 1.61\% \\
    VisualMRC & 3022 & 0.44\% & 3021 & 3.69\% & 3022 & 1.50\% \\
    HiTab & 2495 & 0.36\% & 2495 & 3.04\% & 2495 & 1.24\% \\
    AI2D (cauldron) & 2429 & 0.35\% & 1499 & 1.83\% & 2429 & 1.20\% \\
    VSR & 2152 & 0.31\% & 1062 & 1.30\% & 2152 & 1.07\% \\
    FigureQA & 1880 & 0.27\% & 1880 & 2.29\% & 1880 & 0.93\% \\
    LRV Chart & 1776 & 0.26\% & 1776 & 2.17\% & 1776 & 0.88\% \\
    TQA & 1366 & 0.20\% & 1177 & 1.44\% & 1366 & 0.68\% \\
    Diagram Image2Text & 295 & 0.04\% & 295 & 0.36\% & 295 & 0.15\% \\
    \midrule\multicolumn{7}{c}{Math/Reasoning} \\ \midrule 
    MAVIS Manual Collection & 99990 & 15.60\% & 4033 & 5.22\% & 15763 & 7.08\% \\
    MAVIS Data Engine & 87348 & 13.63\% & 9964 & 12.90\% & 30124 & 13.52\% \\
    Visual Genome & 86417 & 13.49\% & 2895 & 3.75\% & 14992 & 6.73\% \\
    GQA & 72140 & 11.26\% & 3771 & 4.88\% & 13056 & 5.86\% \\
    Geo170K QA & 67823 & 10.58\% & 2330 & 3.02\% & 10150 & 4.56\% \\
    Geo170k Align & 60242 & 9.40\% & 5067 & 6.56\% & 12398 & 5.56\% \\
    TabMWP & 45169 & 7.05\% & 10516 & 13.61\% & 28677 & 12.87\% \\
    MapQA (MathV360K) & 42637 & 6.65\% & 7735 & 10.01\% & 21827 & 9.80\% \\
    GeoQA+ (MathV360K) & 17162 & 2.68\% & 3578 & 4.63\% & 16106 & 7.23\% \\
    UniGeo & 11949 & 1.86\% & 3855 & 4.99\% & 11947 & 5.36\% \\
    LRV Normal & 10489 & 1.64\% & 7481 & 9.68\% & 10489 & 4.71\% \\
    Geometry3K & 9724 & 1.52\% & 3415 & 4.42\% & 9724 & 4.36\% \\
    GEOmVerse (MathV360K) & 9298 & 1.45\% & 2326 & 3.01\% & 9029 & 4.05\% \\
    Super-CLEVR & 8642 & 1.35\% & 2883 & 3.73\% & 6774 & 3.04\% \\
    CLEVR Math & 5280 & 0.82\% & 2248 & 2.91\% & 5248 & 2.36\% \\
    RAVEN & 2100 & 0.33\% & 2100 & 2.72\% & 2100 & 0.94\% \\
    Geo3k & 2091 & 0.33\% & 1165 & 1.51\% & 2091 & 0.94\% \\
    PMC-VQA & 1798 & 0.28\% & 1387 & 1.80\% & 1797 & 0.81\% \\
    GEOS & 498 & 0.08\% & 498 & 0.64\% & 498 & 0.22\% \\
    \midrule\multicolumn{7}{c}{General OCR} \\ \midrule 
    OCR-VQA & 80000 & 39.29\% & 2942 & 20.29\% & 15556 & 29.56\% \\
    Synthdog-EN & 40093 & 19.69\% & 2006 & 13.84\% & 8730 & 16.59\% \\
    TextOCR-GPT4V & 25104 & 12.33\% & 1528 & 10.54\% & 8013 & 15.23\% \\
    TextCaps & 21942 & 10.78\% & 1520 & 10.48\% & 7262 & 13.80\% \\
    Rendered Text & 9995 & 4.91\% & 1881 & 12.97\% & 2796 & 5.31\% \\
    ChromeWriting & 8825 & 4.33\% & 1813 & 12.50\% & 3277 & 6.23\% \\
    HME100K & 7450 & 3.66\% & 761 & 5.25\% & 3364 & 6.39\% \\
    IAM & 5658 & 2.78\% & 1167 & 8.05\% & 2151 & 4.09\% \\
    K12 Printing & 2567 & 1.26\% & 710 & 4.90\% & 982 & 1.87\% \\
    IIIT5K & 1990 & 0.98\% & 171 & 1.18\% & 486 & 0.92\% \\
\end{xtabular}
\end{center}
}
\twocolumn

\clearpage
\pagestyle{empty}
\clearpage

\clearpage
\pagestyle{empty}
\clearpage

\clearpage
\pagestyle{empty}
\clearpage

\clearpage
\pagestyle{empty}
\clearpage

\clearpage
\pagestyle{empty}
\clearpage

\clearpage
\pagestyle{empty}
\clearpage

\clearpage
\pagestyle{empty}
\clearpage

\clearpage
\pagestyle{empty}
\clearpage

\clearpage
\pagestyle{empty}
\clearpage

\clearpage
\pagestyle{empty}
\clearpage

\clearpage
\pagestyle{empty}
\clearpage

\clearpage
\pagestyle{empty}
\clearpage

\clearpage
\pagestyle{empty}
\clearpage

\clearpage
\pagestyle{empty}
\clearpage

\end{document}